\documentclass[11pt,twoside]{article}
\usepackage{fullpage}
\pdfoutput=1 

\usepackage[utf8]{inputenc} 
\usepackage{hyperref}       
\usepackage{url}            
\usepackage{booktabs}       
\usepackage{amsfonts}       
\usepackage{nicefrac}       
\usepackage{microtype}      

\usepackage{graphicx}  
\graphicspath{{figures/}}
\usepackage{subcaption}

\usepackage[square,sort,comma,numbers]{natbib}
\usepackage{algorithm}
\usepackage{algorithmic}
\usepackage{amssymb,amsfonts,amsmath}
\usepackage{amsthm}
\usepackage{enumerate}
\usepackage{wrapfig}

\usepackage[flushleft]{threeparttable}
\usepackage{makecell,booktabs}

\newtheorem{theorem}{Theorem}[section]
\newtheorem{lemma}[theorem]{Lemma}

\newtheorem{definition}[theorem]{Definition}
\let\olddefinition\definition
\renewcommand{\definition}{\olddefinition\normalfont}
\theoremstyle{remark}
\newtheorem{remark}[theorem]{Remark}

\usepackage{stmaryrd}

\usepackage[usenames, dvipsnames]{color}

\usepackage{makecell}
\providecommand{\keywords}[1]{\textbf{\textit{Keywords:}} #1}

\newcommand{\R}[0] {{ \mathbb{R} }}

\newcommand{\N}{{\mathcal{N}}}
\newcommand{\Expect}[0] {{ \mathbb{E} }}

\newcommand{\Jcal}{{\mathcal{J}}}
\newcommand{\Gcal}{{\mathcal{G}}}

\newcommand{\dotprod}[2]{{\langle #1, #2 \rangle}}
\newcommand{\y}[0]{{y}}
\newcommand{\x}[0]{{x}}
\newcommand{\z}[0]{{z}}
\newcommand{\h}[0]{{h}}
\newcommand{\sparse}[0]{{s}}

\newcommand{\pad}[0]{{B}}
\newcommand{\template}[0]{{\Gamma}}
\newcommand{\render}[0]{{\Lambda}}
\newcommand{\ttrans}[0]{{T}}
\newcommand{\trans}[0]{{t}}
\newcommand{\p}[0]{{p}}
\newcommand{\pp}[0]{{\mathcal{P}}}
\newcommand{\rpn}[0]{{\eta}}
\newcommand{\bias}[0]{{b}}
\newcommand{\acts}[0]{{\psi}}
\newcommand{\w}[0]{{W}}
\newcommand{\muy}[0]{{\mu}}
\newcommand{\sratio}[0]{{\tau}}
\DeclareMathOperator{\relu}{ReLu}
\DeclareMathOperator{\maxpool}{MaxPool}
\DeclareMathOperator{\convo}{Conv}

\DeclareMathOperator{\amax}{argmax\,}

\newcommand{\tss}{\hspace*{0.66mm}}
\newcommand{\ze}{\hspace*{\mzerolen}}
\newlength{\mzerolen}

\begin{document}

\begin{center}

  {\bf{\LARGE{A Bayesian Perspective of Convolutional Neural Networks through a Deconvolutional Generative Model}}}

\vspace*{.2in}

{\large{
\begin{tabular}{cccc}
 Tan Nguyen$^{\star, \diamond}$, Nhat Ho$^{\dagger, \diamond}$, Ankit Patel$^{\star, \ddagger}$, \\
 Anima Anandkumar$^{\dagger \dagger, \ddagger \ddagger, \circ}$, Michael I. Jordan$^{\dagger, \circ}$, Richard G. Baraniuk$^{\star, \circ}$
\end{tabular}
}}

\vspace*{.2in}

\begin{tabular}{c}
Rice University, Houston, USA $^\star$ \\
University of California at Berkeley, Berkeley, USA $^\dagger$ \\
Baylor College of Medicine, Houston, USA $^{\ddagger}$ \\
California Institute of Technology, Pasadena, USA $^{\dagger \dagger}$ \\
NVIDIA, Santa Clara, USA $^{\ddagger \ddagger}$
\end{tabular}

\vspace*{.2in}

\today

\begin{abstract}
Inspired by the success of Convolutional Neural Networks (CNNs) for supervised prediction in images, we design the Deconvolutional Generative Model (DGM), a new probabilistic generative model whose inference calculations correspond to those in a given CNN architecture. 
The DGM uses a CNN to design the prior distribution in the probabilistic model. 
Furthermore, the DGM generates images from coarse to finer scales. It introduces a small set of latent variables at each scale, and enforces dependencies among all the latent variables via a conjugate prior distribution. This conjugate prior yields a new regularizer based on paths rendered in the generative model for training CNNs--the Rendering Path Normalization (RPN). We demonstrate that this regularizer improves generalization, both in theory and in practice. In addition, likelihood estimation in the DGM yields training losses for CNNs, and inspired by this, we design a new loss  termed as the Max-Min cross entropy which outperforms the traditional cross-entropy loss for object classification. The Max-Min cross entropy suggests a new deep network architecture, namely the Max-Min network, which can learn from less labeled data while maintaining good prediction performance.
Our experiments demonstrate that the DGM with the RPN and the Max-Min architecture exceeds or matches the-state-of-art on benchmarks including SVHN, CIFAR10, and CIFAR100 for semi-supervised and supervised learning tasks.

\noindent
\keywords{
neural nets, generative models, semi-supervised learning, cross-entropy, statistical guarantee}
\end{abstract}
\let\thefootnote\relax\footnotetext{$\diamond$ Tan Nguyen and Nhat Ho contributed equally to this work.

\hspace{ 0.5 em} $\circ$ Anima Anandkumar, Michael I. Jordan, and Richard G. Baraniuk contributed equally to this work.}
\end{center}

\section{Introduction}
\label{sec:intro}




Unsupervised and semi-supervised learning have still lagged behind despite performance leaps we have seen in supervised learning over the last five years. This is partly due to a lack of good generative models that can capture  all latent variations in complex domains such as natural images and provide useful structures that help learning. When it comes to probabilistic generative models, it is hard to design good priors for the latent variables that drive the generation.

Instead, recent approaches  avoid the explicit design of image priors. For instance, Generative Adversarial Networks (GANs) use implicit feedback from an additional discriminator that distinguishes real from fake images~\cite{goodfellow2014generative}. Using  such feedback helps GANs to generate visually realistic images, but it  is not clear if this is the most effective form of feedback for predictive tasks. Moreover, due to separation of generation and discrimination in GANs, there are typically more parameters to train, and this might make it harder to obtain gains for semi-supervised learning in the low (labeled) sample setting.



We propose an alternative approach to GANs by designing a class of probabilistic generative models, such that inference in those models also has good performance on  predictive  tasks. This approach is well-suited for semi-supervised learning since it eliminates the need for a separate prediction network.  Specifically, we answer the following question:  what generative processes yield Convolutional Neural Networks (CNNs) when inference is carried out?  This is natural to ask since CNNs are state-of-the-art (SOTA) predictive models for images, and intuitively, such powerful predictive models should capture some essence of image generation. However, standard CNNs are not directly reversible and likely do not have all the information for generation since they are trained for predictive tasks such as image classification. We can instead invert the irreversible operations in CNNs, such as the rectified linear units (ReLUs) and spatial pooling, by assigning auxiliary latent variables to account for uncertainty in the CNN's inversion process due to the information loss.

\begin{figure}[t!]
\begin{subfigure}{0.6\textwidth}
\centering
    \includegraphics[width=1\textwidth]{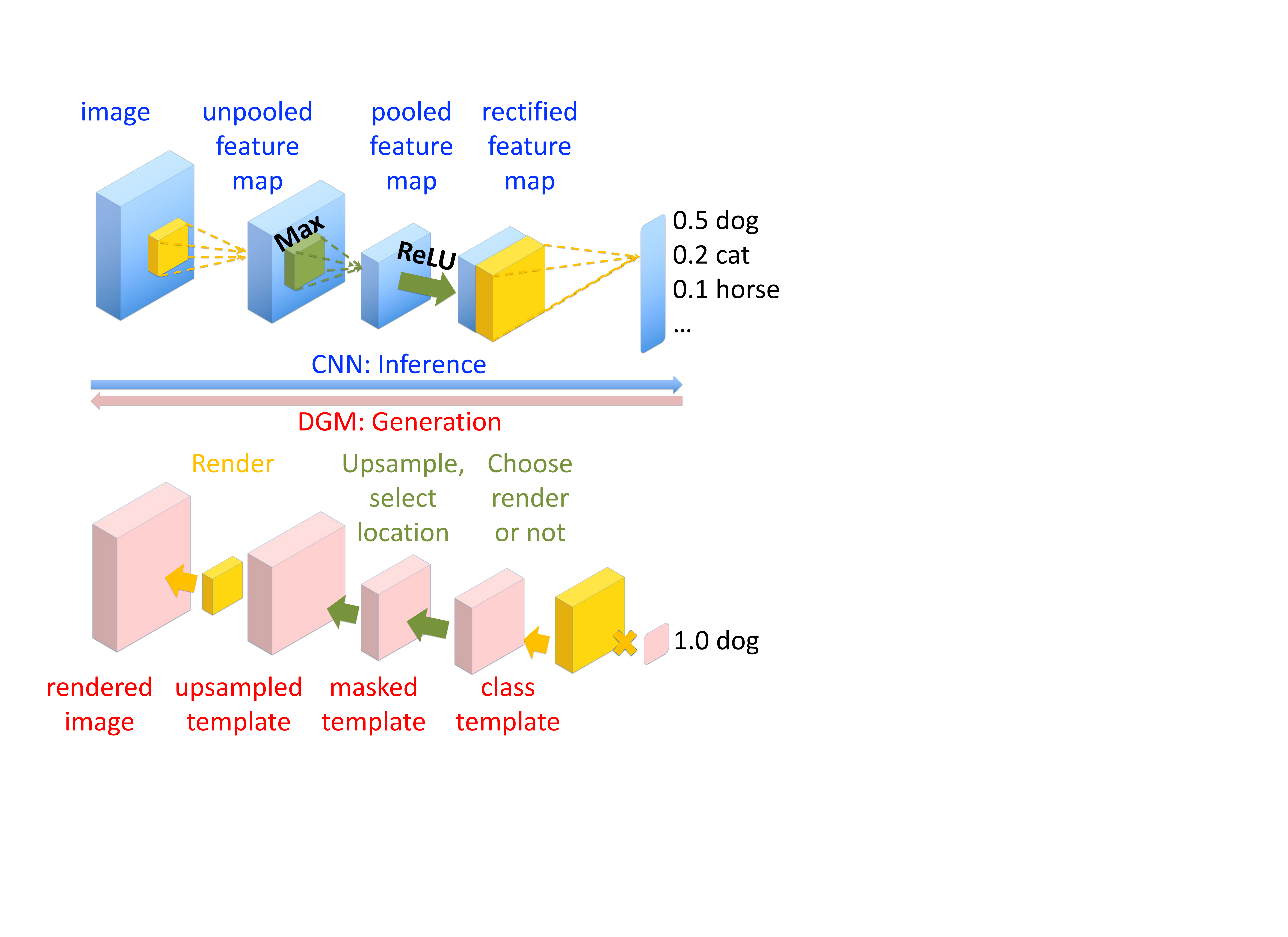}
\label{fig:informed_DGM}
\caption{}
\end{subfigure}
\begin{subfigure}{0.4\textwidth}
\centering
    \vspace{10.25mm}
    \includegraphics[width=1\textwidth]{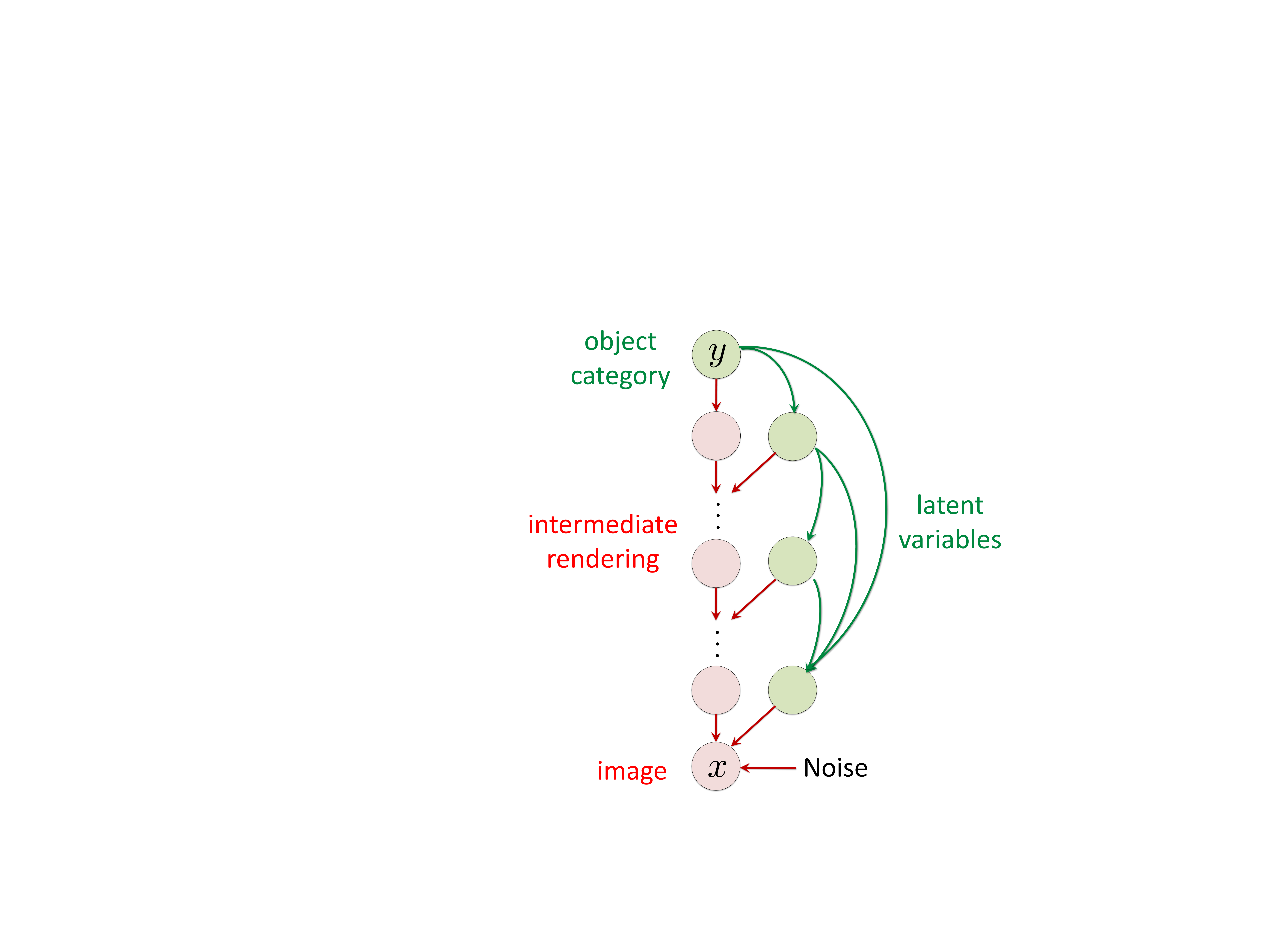} 
    \vspace{10.25mm}
\label{fig:graph_details}
\caption{}
\end{subfigure}
\caption{(a) {\it Deconvolutional Generative Model (DGM)} is a probabilistic generative model that captures latent variations in the data and reduces CNNs as its inference algorithm for supervised learning. In particular, CNN architectures and training losses can be derived from a joint maximum a posteriori (JMAP) inference and likelihood estimation in DGM, respectively. DGM can learn from unlabeled data, overcoming the weakness of CNNs, and achieve good performance on various semi-supervised learning tasks. (b) {\it Graphical model depiction of DGM}. Latent variables in DGM depend on each other. Starting 
    from the top of the model with the object category $\y$, new latent 
    variables are incorporated into the model at each layer, and intermediate images are rendered with finer details. 
    At the bottom of DGM, pixel noise is added to render the final image $x$.} 
\label{fig:lddrm2cnn-high-level-fig}
\end{figure}

\textbf{Contribution 1 -- Deconvolutional Generative Model:} We develop the Deconvolutional Generative Model (DGM) whose  bottom-up inference corresponds to a CNN architecture of choice (see Figure \ref{fig:lddrm2cnn-high-level-fig}a). The ``reverse''  top-down process of image generation is through   coarse-to-fine rendering, which progressively increases the resolution of the rendered image (see Figure \ref{fig:lddrm2cnn-high-level-fig}b). This is intuitive since the reverse process of bottom-up inference   reduces the resolution (and dimension)  through operations such as spatial pooling.  We also introduce structured stochasticity in the rendering process through a small set of discrete latent variables, which capture the uncertainty in reversing the CNN feed-forward process. The rendering in DGM follows a product of linear transformations, which can be considered as the transpose of the inference process in CNNs. In particular, the rendering weights in DGM are proportional to the transpose of the filters in CNNs. Furthermore, the bias terms in the ReLU units at each layer (after the convolution operator) make the latent variables in different network layers dependent (when the bias terms are non-zero). 
This design of image prior has an interesting interpretation from a predictive-coding perspective in neuroscience: the dependency between latent variables can be considered as a form of backward connections that captures prior knowledge from coarser levels in DGM and helps adjust the estimation at the finer levels~\cite{rao1999predictive, friston2018predictive}. The correspondence between DGM and CNN is given in Figure \ref{fig:lddrm2cnn-details} and Table \ref{tbl:correspondence-lddrm-cnn} below.
\begin{figure}[t!]
    \centering
    \includegraphics[width=0.8\textwidth]{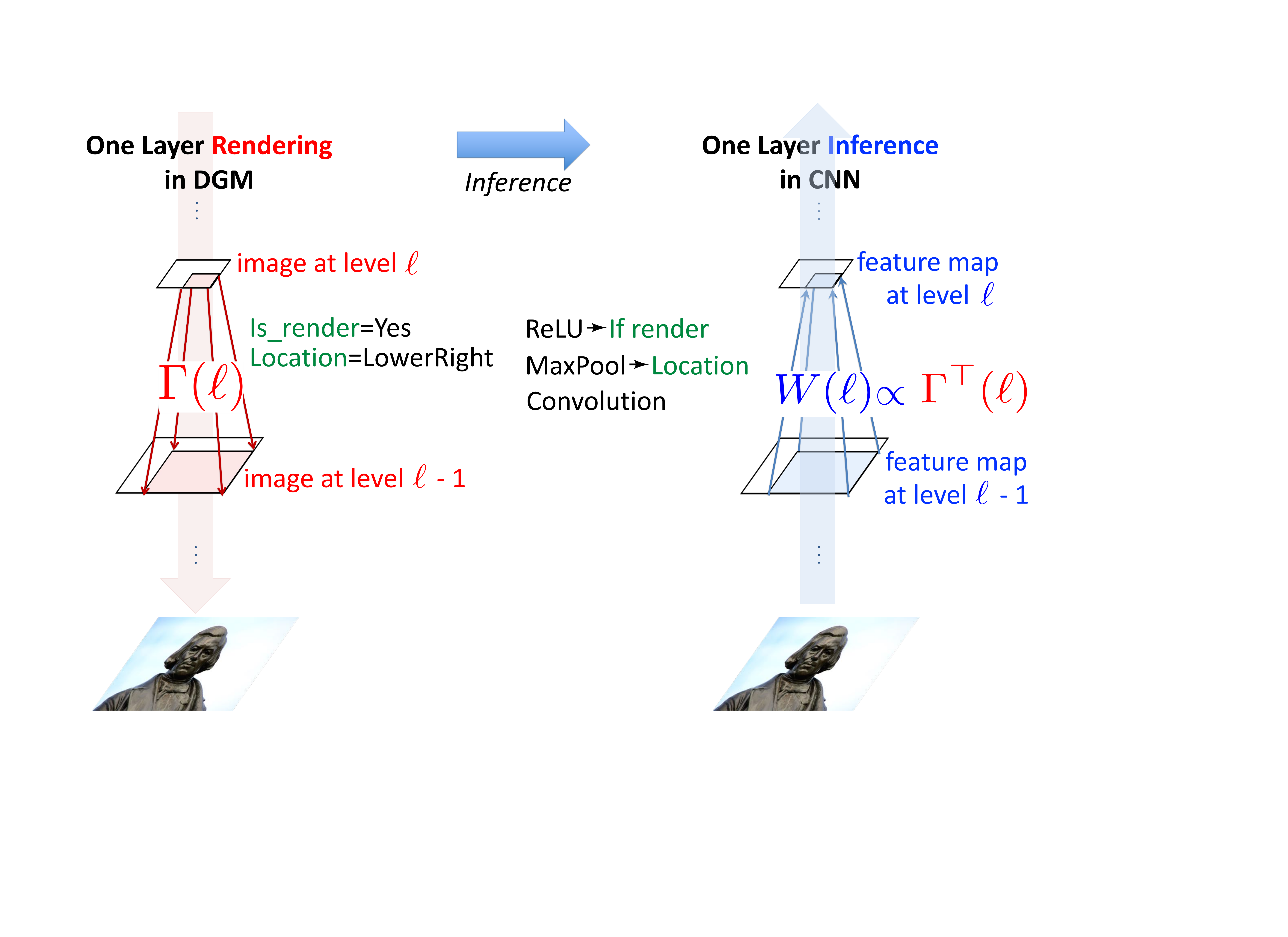} 
    \caption{CNN is the E-step JMAP inference of the optimal latent 
    variable $z(\ell)$ in DGM. In particular, inferring the 
    template selecting latent variables $\sparse(\ell)$ in DGM 
    results in the ReLU non-linearities in CNN. Similarly, inferring 
    the local translation latent variables $\trans(\ell)$ and 
    inverting the zero-padding in DGM yield the MaxPool operators 
    in CNN. In addition, during inference, the rendering step using 
    the templates $\template(\ell)$ in DGM becomes the convolutions 
    with weights $W(\ell)=\template^{\top}(\ell)$ in the CNN.} 
    \label{fig:lddrm2cnn-details}
\end{figure}

DGM is a likelihood-based framework, where unsupervised learning can be derived by maximizing the expected complete-data log-likelihood of the model while supervised learning is done through optimizing the class-conditional log-likelihood. Semi-supervised learning unifies both log-likelihoods into an objective cost for learning from both labeled and unlabeled data. The DGM prior has the desirable property of being a conjugate prior, which makes learning in DGM computationally efficient. 

Interestingly, we derive the popular \emph{cross-entropy loss} used to train CNNs for supervised learning \emph{as an upper bound of the DGM's negative class-conditional log-likelihood}. A broadly-accepted interpretation of cross-entropy loss in training CNNs is from logistic regression perspective. Given features extracted from the data by the network, logistic regression is applied to classify those features into different classes, which yields the cross-entropy. In this interpretation, there is a gap between feature extraction and classification. On the contrary, our derivation ties feature extraction and learning for classification in CNN into an end-to-end optimization problem that estimates the conditional log-likelihood of DGM. This new interpretation of cross-entropy allow us to develop better losses for training CNNs. An example is the \emph{Max-Min cross-entropy} discussed in Contribution 2 and Section \ref{sec:max_min}.
\begin{table}[t!]
  \caption{Correspondence between DGM and CNN}
  \centering
  \begin{threeparttable}
 
    \begin{tabular}{c@{\qquad}c}
      DGM & CNN  \\ \midrule\midrule 
        \makecell{Rendering templates $\template(\ell)$} & \makecell{Transpose of filter weights $W(\ell)$} \\
     \cmidrule(l r){1-2}
     \makecell{Class templates $\muy(\y)$} & \makecell{Softmax weights} \\
     \cmidrule(l r){1-2}
      \makecell{Parameters $\bias(\trans;\ell)$ \\ of the conjugate prior $\pi_{\y|\x}$} & \makecell{Bias terms $\bias(\ell)$ in the ReLUs \\ after each convolution} \\
     \cmidrule(l r){1-2}
    \makecell{Max-marginalize over $\sparse(\ell)$} & ReLU \\ \cmidrule(l r){1-2}
    \makecell{Max-marginalize over $\trans(\ell)$} & MaxPool \\ \cmidrule(l r){1-2}
     \makecell{Conditional log-likelihood \\ $\dfrac{1}{n} \sum_{i=1}^{n} \log p(\y_{i}|\x_{i}, \z_{i};\theta)$ } & Cross-entropy loss \\ \cmidrule(l r){1-2}
     \makecell{Expected complete-data log-likelihood \\ $\dfrac{1}{n} \sum_{i=1}^{n} \Expect\left[\log p(\x_{i}, \z_{i}|\y_{i})\right]$ } & Reconstruction loss \\ \cmidrule(l r){1-2}
     \makecell{Normalize the intermediate image $h(\ell)$} & \makecell{Batch Normalization} \\
     \cmidrule(l r){1-2}
    \end{tabular}
  
  \end{threeparttable}
  \label{tbl:correspondence-lddrm-cnn}
 \end{table}

\textbf{Contribution 2 -- New regularization, loss function, architecture and generalization bounds:} 
The joint nature of generation, inference, and learning in DGM allows us to develop new training procedures for semi-supervised and supervised learning, as well as new theoretical (statistical) guarantees for learning. In particular, for training, we derive a new form of regularization termed as the \textit{Rendering Path Normalization} (RPN) from the DGM's conjugate prior. A rendering path is a set of latent variable values in DGM. Unlike the path-wise regularizer in ~\cite{neyshabur2015path}, RPN uses information from a generative model to penalizes the number of the possible rendering paths and, therefore, encourages the network to be compact in terms of representing the image. It also helps to enforce the dependency among different 
layers in DGM during training and improves classification performance.

We provide new theoretical bounds based on DGM. In particular, we prove that DGM is statistically consistent and derive a generalization bound of DGM for (semi-)supervised learning tasks. Our generalization bound is proportional to the 
number of active rendering paths that generate close-to-real images.
This suggests that RPN regularization may help in 
generalization since RPN enforces the dependencies among latent variables in DGM and, therefore, reduces the number of active 
rendering paths. We observe that RPN helps improve generalization in our experiments.
 
\underline{\emph{Max-Min cross-entropy and network}}: We propose the new \emph{Max-Min cross-entropy} loss function for learning, based on negative class-conditional log-likelihood in DGM. It combines the traditional cross-entropy with another loss, which we term as the Min cross-entropy.  While the traditional (Max) cross-entropy maximizes the probability of the correct labels, the Min cross-entropy minimizes the probability of the incorrect labels. We show that the Max-Min cross-entropy is also an upper bound to the negative conditional log-likehood of DGM, just like the cross-entropy loss. The Max-Min cross-entropy is realized through a new CNN architecture, namely the \emph{Max-Min network}, which is a CNN with an additional branch sharing weights with the original CNN but containing minimum pooling (MinPool) operator and negative rectified linear units (NReLUs), i.e., $\min(\cdot,0)$ (see Figure~\ref{fig:max-min}). Although the Max-Min network is derived from DGM, it is a meta-architecture that can be applied independently on any  CNN architecture. We show empirically that Max-Min networks and cross-entropy help improve the SOTA on object classification for supervised and semi-supervised learning.

\textbf{Contribution 3 -- State-of-the-art empirical results for semi-supervised and supervised learning:} We show strong results for semi-supervised learning over
CIFAR10, CIFAR100 and SVHN benchmarks in comparison with SOTA methods that use and do not use consistency regularization. Consistency regularization, such as those used in Temporal Ensembling ~\cite{laine2016temporal} and Mean Teacher ~\cite{tarvainen2017mean}, enforces the 
networks to learn representation invariant to realistic perturbations of the data. DGM alone outperforms most SOTA methods which do not use consistency regularization~\cite{salimans2016improved, dumoulin2016adversarially} in most settings. Max-Min cross-entropy then helps improves DGM's semi-supervised learning results significantly. When combining DGM, Max-Min cross-entropy, and Mean Teacher, we achieve SOTA results or very close to those on CIFAR10, CIFAR100, and SVHN (see Table \ref{tbl:cifar10}, \ref{tbl:cifar100}, and \ref{tbl:svhn}). 
Interestingly, compared to the other competitors, our method is consistently good, achieving either best or second best results in all experiments. Furthermore, Max-Min cross-entropy also helps supervised learning. Using the Max-Min cross-entropy, we achieve SOTA result for supervised learning on CIFAR10 (2.30\% test error). Similarly, Max-Min cross-entropy helps improve supervised training on ImageNet.

Despite good classification results, there is a caveat that DGM may not generate good looking images since that objective is not ``baked'' into its training. DGM is primarily aimed at improving semi-supervised and supervised learning through better regularization. Potentially, an adversarial loss can be added to DGM to improve visual characteristics of the image, but that is beyond the scope of this paper.

{\bf Notation:} To facilitate the presentation, the DGM's notations are explained in Table \ref{tbl:notations-lddrm}. 
Throughout this paper, we denote $\|x\|$ and $x^{\top}
$ the Euclidean norm and transpose of $x$ respectively for any $x \in 
\mathbb{R}^{d}$. Additionally, for any matrix $A$ and $B$ of the same 
dimension, $A \odot B$ denotes the Hadamard product between $A$ and $B
$. For any vector $a$ and $b$ of the same dimension, $\dotprod{a}{b}$ 
denotes the dot product between $a$ and $b$. 

\begin{table}[htbp]\caption{Table of notations for DGM}
\begin{center}
\begin{tabular}{r c p{10cm} }
\toprule
\multicolumn{3}{c}{\underline{Variables}}\\
\multicolumn{3}{c}{}\\
$\x$ & $\triangleq$ & input image of size $D(0)$\\
$\y$ & $\triangleq$ & object category\\
$\z(\ell) = \{s(\ell), t(\ell)\}$ & $\triangleq$ & all latent variables of size $D(\ell)$ in layer $\ell$\\
$\sparse(\ell,p)$ & $\triangleq$ & switching latent variable at pixel location $p$ in layer $\ell$\\  
$\trans(\ell,p)$ & $\triangleq$ & local translation latent variable at pixel location $p$ in layer $\ell$\\  
$\h(\y,\z;\ell)=\h(\ell)$ & $\triangleq$ & intermediate rendered image of size $D(\ell)$ in layer $\ell$\\ 
$\h(\y,\z;0)=\h(0)$ & $\triangleq$ & rendered image of size $D(0)$ from DGM before adding noise\\ 
$\psi(\ell)$ & $\triangleq$ & corresponding feature maps in layer $\ell$ in CNNs.\\ 
\multicolumn{3}{c}{}\\
\multicolumn{3}{c}{\underline{Parameters}}\\
\multicolumn{3}{c}{}\\
$\muy(\y)=\h(\y;L)=\h(L)$ & $\triangleq$ & the template of class $\y$, as well as the coarsest image of size $D(L)$ determined by the category $\y$ at the top of DGM before adding any fine detail. $\muy(\y)$ is learned from the data.\\ 
$\render(\ell)$ & $\triangleq$ & rendering matrix of size $D(\ell-1) \times D(\ell)$ at layer $\ell$.\\  
$\template(\ell)$ & $\triangleq$ & dictionary of $D(\ell)$ rendering template $\template(\ell, p)$ of size $F(\ell) \times 1$ at layer $\ell$. $\template(\ell)$ is learned from the data.\\  
$W(\ell)=\template^{\top}(\ell, p)$ & $\triangleq$ & corresponding weight at the layer $\ell$ in CNNs\\  
$\pad(\ell)$ & $\triangleq$ & set of zero-padding matrices $\pad(\ell,p) \in \R^{D(\ell-1) \times F(\ell)})$  at layer $\ell$\\  
$\ttrans(\ell)$ & $\triangleq$ & set of local translation matrices $\ttrans(\ell,p) \in \R^{D(\ell-1) \times D(\ell-1)})$  at layer $\ell$. $\ttrans(\ell,p)$ is chosen according to value of $t(\ell,p)$\\
$\bias(\trans;\ell) = \bias(\ell)$ & $\triangleq$ & parameter of the conjugate 
prior $p(\z|\x,\y)$ at layer $\ell$. This term is of size $D(\ell)$ 
and becomes the bias term after convolutions in CNNs. It can be made 
independent of $\trans$, which is equivalent to using the same bias in 
each feature map in CNNs. Here, $\bias(\trans;\ell)$ is learned from 
data.\\
$\pi_{\y}$ & $\triangleq$ & probability of object category $\y$.\\
$\sigma^{2}$ & $\triangleq$ & pixel noise variance\\
\multicolumn{3}{c}{}\\
\multicolumn{3}{c}{\underline{Other Notations}}\\
\multicolumn{3}{c}{}\\
$\rpn(\y,\z)$ & $\triangleq$ & $\sum_{\ell=1}^{L}
     \frac{1}{\sigma^{2}}\dotprod{\bias(\trans;\ell)}{\sparse(\ell)\odot \h(\ell)}$.\\
$\text{Softmax}\left(\rpn\right)$ & $\triangleq$ & $\frac{\exp (\rpn)}{\sum \limits_{\rpn^{'}} \exp (\rpn^{'})}$.\\
RPN & $\triangleq$ & $- \frac{1}{n}\sum \limits_{i=1}^{n} \log p(\z^{*}_{i}|\y_{i}) = -\frac{1}{n}\sum \limits_{i=1}^{n}\text{Softmax}\left(\rpn(\y_{i},\z_{i}^{*}) \right)$.\\
(\y,\z(L), \dots, \z(1)) & $\triangleq$ & rendering configuration.\\
\bottomrule
\end{tabular}
\end{center}
\label{tbl:notations-lddrm}
\end{table}
\section{Related Work}
\label{sec:related_works}
\paragraph{Deep Generative Models:}
In addition to GANs, other recently developed deep generative models include the Variational Autoencoders (VAE)~\cite{kingma2013auto} and the Deep Generative Networks~\cite{kingma2014semi}. Unlike these models, which replace complicated or intractable inference by CNNs, DGM derives CNNs as its inference. This advantage allows us to develop better learning algorithms for CNNs with statistical guarantees, as being discussed in Section~\ref{sec:learning-in-lddrm} and \ref{sec:stats}. Recent works including the Bidirectional GANs~\cite{donahue2016bigan} and the Adversarially Learned Inference model~\cite{dumoulin2016adversarially} try to make the discriminators and generators in GANs reversible of each other, thereby providing an alternative way to invert CNNs. These approaches, nevertheless, still employ a separate network to bypass the irreversible operators in CNNs. Furthermore, the flow-based generative models such as NICE~\cite{dinh2014nice}, Real NVP~\cite{dinh2016density}, and Glow~\cite{kingma2018glow} are invertible. However, the inference algorithms of these models, although being exact, do not match the CNN architecture. DGM is also close in spirit to the Deep Rendering Model (DRM)~\cite{patel2016probabilistic} but markedly different. Compared to DGM, DRM has several limitations.
In particular, all the latent variables in DRM are 
assumed to be independent, which is rather unrealistic. This lack of dependency causes the missing of the bias terms in the ReLUs of the CNN derived from DRM. Furthermore, the 
cross-entropy loss used in training CNNs for supervised learning tasks is not captured naturally by DRM. Due to these limitations, model consistency and generalization bounds are not derived for DRM.

\paragraph{Semi-Supervised Learning:}
In addition to deep generative model approach, consistency regularization methods, such as Temporal Ensembling \cite{laine2016temporal} and Mean Teacher \cite{tarvainen2017mean}, have been recently developed for semi-supervised learning and achieved state-of-the-art results. These methods enforce that the baseline network learns invariant representations of the data under different realistic perturbations. Consistency regularization approaches are complimentary to and can be applied on most deep generative models, including DGM, to further increase the baseline model's performance on semi-supervised learning tasks. Experiments in Section~\ref{sec:experiments} demonstrate that DGM achieves better test accuracy on CIFAR10 and CIFAR100 when combined with Mean Teacher.

\paragraph{Explaining Architecture of CNNs:}
The architectures and training losses of CNNs have been studied from other perspectives. 
\cite{achille2017emergence, achille2018infodropout} employ principles from information theory 
such as the Information Bottleneck Lagrangian 
introduced by \cite{Tishby99theinformation} to show that stacking layers encourages CNNs to learn representations invariant to latent variations. They also study the cross-entropy loss to understand possible causes of over-fitting in CNNs and suggest a new regularization term for training that helps the trained CNNs generalize better. This regularization term relates to the amount of information about the labels memorized in the weights. 
Additionally, \cite{papyan2016convolutional} suggests a connection between CNNs and the convolutional sparse coding (CSC) \cite{bristow2013fast, wohlberg2014efficient, heide2015fast, papyan2017working}. They propose a multi-layer CSC (ML-CSC) model and prove that CNNs are the thresholding pursuit serving the ML-CSC model.  This thresholding pursuit framework implies alternatives to CNNs, which is related to the deconvolutional and recurrent networks. The architecture of CNNs is also investigated using the wavelet scattering transform. 
In particular, scattering operators help elucidate different properties of CNNs, including  how the image sparsity and geometry captured by the networks \cite{bruna2013invariant, mallat2016understanding}. In addition, CNNs are also studied from the optimization perspective \cite{arora2018optimization, freeman2016topology, choromanska2015loss, kawaguchi2016deep}, statistical learning theory perspective \cite{kawaguchi2017generalization, neyshabur2017exploring}, approximation theory perspective~\cite{balestriero2017arxiv}, and other approaches \cite{vidal2017mathematics, gal2016dropout}.

Like these works, DGM helps explain different components in CNNs. It employs tools and methods in probabilistic inference to interpret CNN from a probabilistic perspective. That says, DGM can potentially be combined with aforementioned approaches to gain a better understanding of CNNs.
\section{The Deconvolutional Generative Model} 
\label{sec:latent_drm}
We first define the Deconvolutional Generative Model (DGM). Then we discuss the inference in DGM. Finally, we derive different learning losses from DGM, including the cross-entropy loss, the reconstruction loss, and the RPN regularization, for supervised learning, unsupervised learning, and semi-supervised learning.

\subsection{Generative Model}
\label{sec:model}
DGM attempts to invert CNNs as its inference so that the information in the posterior $p(y|x)$ can be used to inform the generation process of the model. DGM realizes this inversion by employing the structure of its latent variables. Furthermore, the joint prior distribution of latent variables in the model is parametrized such that it is the conjugate prior to the likelihood of the model. This conjugate prior is the function of intermediate rendered images in DGM and implicitly captures the dependencies among latent variables in the model. More precisely, DGM can be defined as follows:

\begin{definition}[\textbf{Deconvolutional Generative Model (DGM)}] \label{defn:drmm}
DGM is a deep generative model in which the latent variables $\z(\ell)= \{\trans(\ell), \sparse(\ell)\}_{\ell=1}^{L}$ at different layers $\ell$ are dependent. Let $x$ be the input image and $y \in \left\{1,\ldots,K\right\}$ be the target variable, e.g. object category. Generation in DGM takes the form:
\begin{align}
     \pi_{\z|\y}  &\triangleq \text{Softmax}\left(\frac{1}{\sigma^{2}}\rpn(\y,\z)\right) \label{eqn:joint-prior} \\
      \z|\y &\sim \textrm{Cat}(\pi_{\z|\y}) \nonumber \\
     \h(\y,\z;0) &\triangleq  \render(\z;1)\render(\z;2)\cdots \render(\z;L)\muy(\y) \label{eqn:render-eqn}\\
    \x|\z,\y &\sim  \N(\h(0),\sigma^{2}\textbf{1}_{D(0)}) \nonumber, 
\end{align}
\end{definition}
\noindent where:
\begin{align}
\rpn(\y,\z) &\triangleq \sum_{\ell=1}^{L}
     \dotprod{\bias(\trans;\ell)}{\sparse(\ell)\odot \h(\ell)} \\
    \text{Softmax}\left(\rpn\right) &\triangleq \frac{\exp (\rpn)}{\sum \limits_{\rpn} \exp (\rpn)}.
\end{align}

The generation process in DGM can be summarized in the following steps:
\begin{enumerate}
    \item Given the class label $\y$, DGM first samples the latent variables $\z$ from a categorical distribution whose prior is $\pi_{\z|\y}$. 
    \item Starting from the class label $\y$ at the top layer $L$ of the model, DGM renders its coarsest image, $\h(L)=\muy(\y)$, which is also the object template of class $\y$.
    \item At layer $L-1$, a set of of latent variations $\z(L)$ is incorporated into $\h(\y;L)$ via a linear transformation $\Lambda(\z; L)$ to render the finer image $\h(\y, \z;L-1)$. The same process is repeated at each subsequent layer $\ell$ to finally render the finest image $\h(\y, \z;0)$ at the bottom of DGM.
    \item Gaussian pixel noise is added into $\h(\y, \z;0)$ to render the final image $x$.
\end{enumerate}

\begin{figure}[t!]
    \centering
    \includegraphics[width=1.0\textwidth]{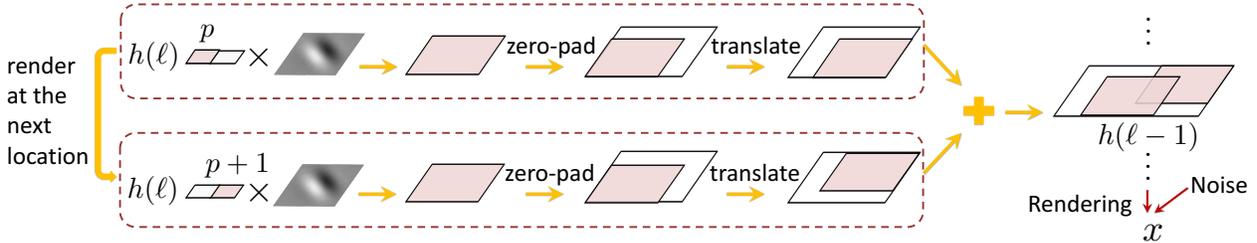} 
    \caption{Rendering process from layer $\ell$ to layer $\ell - 1$ 
    in DGM. At each pixel $\p$ in the intermediate image $\h(\ell)
    $, DGM decides to render if the template selecting latent 
    variable $\sparse(\ell, \p) = 1$. Otherwise, it does not render. 
    If rendering, the template $\template(\ell, \p)$ is multiplied by 
    the pixel value $\h(\ell, \p)$. Then the matrix $\pad(\ell,\p)$ 
    zero-pads the result to the size of the intermediate image at 
    layer $\ell -1 $, which is $D(\ell-1) \times D(\ell -1)$. After 
    that, the translation matrix $\ttrans(\ell, \p)$ locally 
    translates the rendered image to location specified by the latent 
    variable $\trans(\ell,\p)$. The same rendering process repeats at 
    pixel $\p + 1$, as well as at other pixels of $\h(\ell)$. DGM 
    adds all rendered images to achieve the intermediate image $
    \h(\ell - 1)$ at layer $\ell -1$.} 
    \label{fig:main-fig}
\end{figure}

In the generation process above, $\Lambda(\z;\ell)$ can be any linear 
transformation, and the latent variables $\z(\ell)$, in their most 
generic form, can capture any latent variation in the data. While such 
generative model can represent any possible imagery data, it cannot be 
learned to capture the properties of natural images in reasonable time 
due to its huge number of degrees of freedom. 
Therefore, it is necessary to have 
more structures in DGM given our prior knowledge of natural 
images. One such prior knowledge is from classification models, $p(\y|
\x)$. In particular, since classification models like CNNs have 
achieved excellent performance on a wide range of object 
classification tasks, we hypothesize that a generative model whose 
inference yields CNNs will also be a good model for natural images. As a 
result, we would like to introduce new structures into DGM so that 
CNNs can be derived as DGM's inference. In other word, we use the 
posterior $p(\y|\x)$, i.e., CNNs, to inform the likelihood $p(\x|\y)$ 
in designing DGM.

In our attempt to invert CNNs, we constrain the latent variables $
\z(\ell)$ at layer $\ell$ in DGM to a set of template selecting 
latent variables $\sparse(\ell)$ and local translation latent 
variables $\trans(\ell)$. As been shown later in Section 
\ref{sec:jmap-inference}, during inference of DGM, the ReLU non-linearity at layer $\ell$ ``inverts" $\sparse(\ell)$ to find if 
particular features are in the image or not. Similarly, the MaxPool 
operator ``inverts" $\trans(\ell)$ to locate where particular features, 
if exist, are in the image. Both $\sparse(\ell)$ and $\trans(\ell)$ 
are vectors indexed by $\p(\ell) \in \pp(\ell) \equiv \{\textrm{pixels 
in layer } \ell\}$. 

The rendering matrix $\Lambda(\ell)$ is now a function of $
\sparse(\ell)$ and $\trans(\ell)$, and the rendering process from 
layer $\ell$ to layer $\ell - 1$ using $\Lambda(\sparse, \trans;\ell)$ 
is described in the following equation:
\begin{align}
    \h(\ell - 1) & \triangleq \render(\ell)\h(\ell) = \sum_{\p\in \pp(\ell)}  \sparse(\ell, \p)\ttrans(t;\ell,\p) \pad(\ell, \p)\template(\ell, \p) \h(\ell,\p). \label{eqn:render-1layer}
\end{align}
Even though the rendering equation above seems complicated at first, 
it is quite intuitive as illustrated in Figure \ref{fig:main-fig}. At 
each pixel $\p$ in the intermediate image $\h(\ell)$ at layer $\ell$, 
DGM decides to use that pixel to render or not according to the 
value of the template selecting latent variable $\sparse(\p;\ell)$ at 
that pixel location. If $\sparse(\p;\ell)=1$, then DGM renders. 
Otherwise, it does not. If rendering, then the pixel value $\h(\ell,
\p)$ is used to scale the rendering template $\Gamma(\ell,\p)$. This 
rendering template is local. It has the same number of feature maps as 
the next rendered image $\h(\ell - 1)$, but is of smaller size, e.g. 
$3 \times 3$ or $5 \times 5$. As a result, the rendered image 
corresponds to a local patch in $\h(\ell-1)$. Next, the padding matrix 
$\pad(\ell,\p)$ pads the resultant patch to the size of the image $
\h(\ell-1)$ with zeros, and the translation matrix $\ttrans(t;\ell,p)$ 
translates the result to a local location. DGM then keeps rendering 
at other pixel location of $
\h(\ell)$ following the same process. All 
rendered images are added to form the final rendered image $\h(\ell-1)
$ at layer $\ell - 1$.

Note that in DGM, there is one rendering template at each pixel 
location $\p$ in the image $\h(\ell)$. For example, if $\h(\ell)$ is 
of size $128 \times 6 \times 6$, then the DGM uses $4608$ templates 
to render at layer $\ell$. This is too many rendering templates and 
would require a very large amount of data to learn, considering all 
layers in DGM. Therefore, we further constrain DGM by enforcing 
all pixels in the same feature maps of $\h(\ell)$ share the same 
rendering template. In other word, $\Gamma(\ell,\p)$ are the same if $
\p$ are in the same feature map. This constrain helps yield 
convolutions in CNNs during the inference of DGM, and the rendering 
templates in DGM now correspond to the convolution filters in CNNs.

While $\sparse(\ell)$ and $\trans(\ell)$ can be let independent, we further constrain the model by \emph{enforcing the dependency 
among $\sparse(\ell)$ and $\trans(\ell)$ at different layers in DGM}. This constraint is motivated from realistic rendering of natural 
objects: \emph{different parts of a natural object are dependent on 
each other}. For example, in an image of a person, the locations of 
the eyes in the image are restricted by the location of the head or if 
the face is not painted, then it is likely that we cannot find the 
eyes in the image either. Thus, DGM tries to capture such 
dependency in natural objects by imposing more structures into the 
joint prior of latent variables  $\sparse(\ell)$ and $\trans(\ell)$ at 
all layer $\ell$, $\ell = 1, 2, \dots, L$ in the model. In particular, 
the joint prior $\pi_{\z|\y}=p(\{\z(\ell)\}_{\ell=1}^{L}|\y)$ is given 
by Eqn.~\ref{eqn:joint-prior}. The form of the joint prior $\pi_{\z|\y}$ might look mysterious at first, but DGM parametrizes $\pi_{\z|\y}$ in this particular way so that $\pi_{\z|\y}$ is the conjugate prior of the model likelihood as proven in Appendix~\ref{sec:conjugate-prior-proof}. Specifically, in order to derive conjugate prior, we would like the log conditional distribution $\log p(\z(\ell)|\z(\ell + 1),\dots,\z(L))$ to have the linear piece-wise form as the  CNNs which compute the posterior $p(\y|\x)$. This design criterion results in each term $\dotprod{\bias(\trans;\ell)}{\sparse(\ell)\odot \h(\ell)}$ in the joint prior $\pi_{\z|\y}$ in Eqn.~\ref{eqn:joint-prior}. The conjugate form of $\pi_{\z|\y}$ allows efficient inference in the DGM.
 Note that $\bias(\trans; \ell)$ are the parameters of the
conjugate prior $\pi_{z|y}$. Due to the form  of $\pi_{\z|\y}$, during 
inference, \emph{$\bias(\trans;\ell)$ will become the bias terms after 
convolutions in CNNs} as will be shown in Theorem~\ref{theorem:jmap_inference_theorem}. 
Furthermore, when training 
in an unsupervised setup, the conjugate prior results in the RPN 
regularization as shown in Theorem~\ref{theorem:learning_lddrm}(b). 
This RPN regularization helps enforce the dependencies among latent 
variables in the model and increases the likelihood of latent configuration presents in the data during training. For the sake of clarity, we summarize all the notations used in DGM in Table \ref{tbl:notations-lddrm}. 

\begin{figure}
\centering
\begin{tabular}{c@{\hskip 0.5mm}c@{\hskip 0.5mm}c@{\hskip 0.5mm}c@{\hskip 0.5mm}c@{\hskip 2mm}c}
Layer 4 & Layer 3 & Layer 2 & Layer 1 & Layer 0 & Original \\[1pt]
  \includegraphics[width=18mm]{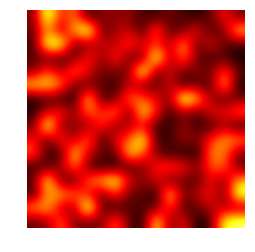} & \includegraphics[width=18mm]{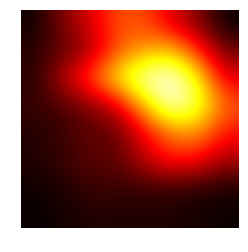} & \includegraphics[width=18mm]{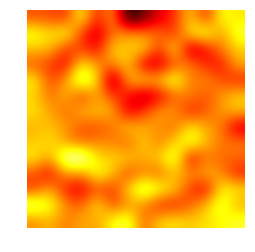} & \includegraphics[width=18mm]{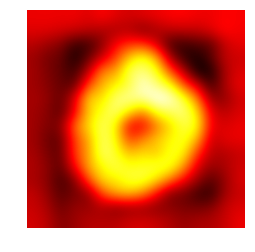} & \includegraphics[width=18mm]{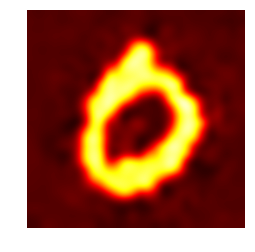} &
  \includegraphics[width=18mm]{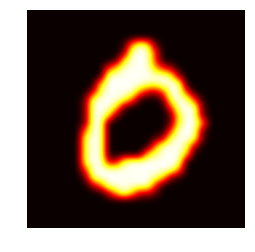} \\
  \includegraphics[width=18mm]{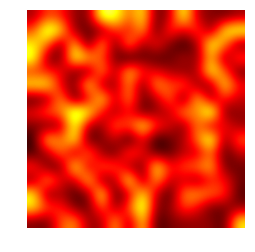} & \includegraphics[width=18mm]{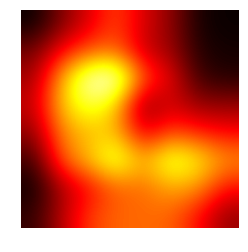} & \includegraphics[width=18mm]{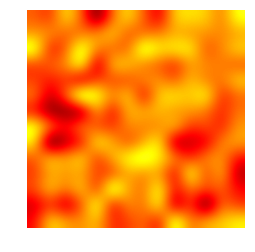} & \includegraphics[width=18mm]{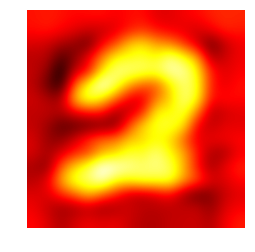} & \includegraphics[width=18mm]{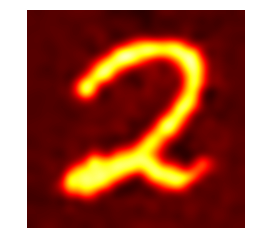} &
  \includegraphics[width=18mm]{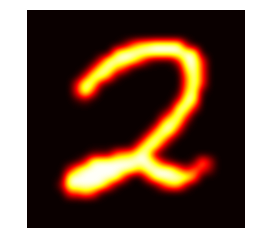}\\
  \includegraphics[width=18mm]{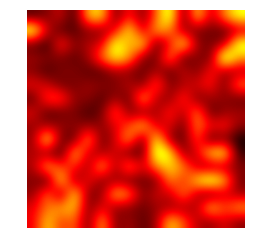} & \includegraphics[width=18mm]{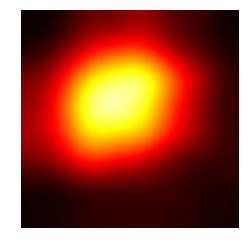} & \includegraphics[width=18mm]{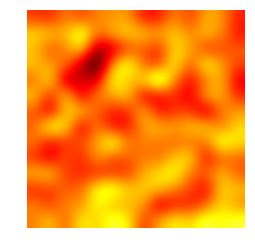} & \includegraphics[width=18mm]{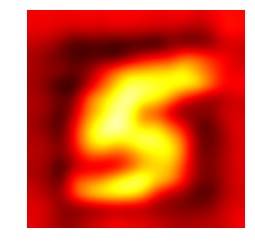} & \includegraphics[width=18mm]{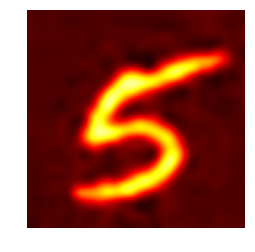} &
  \includegraphics[width=18mm]{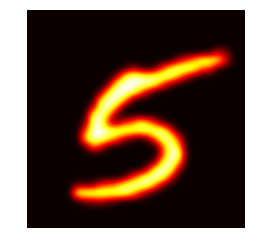}
\end{tabular}
\caption{Reconstructed images at each layer in a 5-layer DGM trained on MNIST with 50K labeled data. Original images are in the rightmost column. Early layers in the rendering such as layer 4 and 3 capture coarse-scale features of the image while later layers such as layer 2 and 1 capture finer-scale features. From layer 2, we begin to see the gist of the rendered digits.
}
\label{fig:reconst-imgs}
\end{figure}

We summarize 
the DGM's rendering process in Algorithm~\ref{alg:rendering-lddrm} 
below. Reconstructed images at each layer of a 5-layered DGM trained on MNIST are visualized in Figure \ref{fig:reconst-imgs}. DGM reconstructs the images in two steps. First, the bottom-up E-step inference in DGM, which has a CNN form, keeps track of the optimal latent variables $\sparse^{*}(\ell)$ and $\trans^{*}(\ell)$ from the input image. Second, in the top-down E-step reconstruction, DGM uses $\sparse^{*}(\ell)$ and $\trans^{*}(\ell)$ to render the reconstructed image according to Eqn.~\ref{eqn:render-eqn} and \ref{eqn:render-1layer}. The network is trained using the semi-supervised learning framework discussed in Section \ref{sec:learning-in-lddrm}. The reconstructed images show that DGM renders images from coarse to fine. Early layers in the model such as layer 4 and 3 capture coarse-scale features of the image while later layers such as layer 2 and 1 capture finer-scale features. Starting from layer 2, we begin to see the gist of the rendered digits which become clearer at layer 1. Note that images at layer 4 represent the class template $\muy_{\y}$ in DGM, which is also the softmax weights in CNN.

In this paper, in order to simplify the notations, we only model $p(z|y)$.
More precisely, in the subsequent sections, $y_{i}$ is set to $
\mathop{\arg \max}\limits_{\y'} \biggr(\dotprod{\h(\y',\z^{*}_{i};0)}
{\x_{i}} + \rpn(\y',\z^{*}_{i}) + \ln \pi_{y'}\biggr)$ for unlabeled 
data. An extension to model $p(z,y)$ can be achieved by adding the 
term $\ln \pi_y$ inside the $\text{Softmax}$ operator in Eqn.~\ref{eqn:joint-prior}. All theorems and proofs can also be 
easily extended to the case when $p(z,y)$ is modeled. Furthermore, to 
facilitate the discussion, we will utilize $\template(\ell)$, $\pad(\ell)$, 
and $\ttrans(t;\ell)$ to denote the set of all rendering templates $
\template(\ell, \p)$, zero-padding matrices $\pad(\ell, \p)$, and local 
translation matrices $\ttrans(t;\ell,\p)$ at layer $\ell$, respectively. 
We will also use $\render(\ell)$, $\render(\z;\ell)$, and $
\render(\trans,\sparse;\ell)$ interchangeably. Similarly, the notations 
$\h(\y, \z;\ell)$ and $\h(\ell)$, as well as $\bias(\trans;\ell)$ and 
$\bias(\ell)$ are used interchangeably.
\begin{algorithm}[h!]
   \caption{Rendering Process in DGM}
   \label{alg:rendering-lddrm}
\begin{algorithmic}
   \STATE {\bfseries Input:} Object category $\y$. 
   \STATE {\bfseries Output:} Rendered image $\x$ given the object category $\y$.
   \STATE {\bfseries Parameters:} $\theta=\biggr(\left\{\muy(\y)\right\}_{\y=1}^{K},\left\{\template(\ell)\right\}_{\ell=1}^{L},\left\{\bias(\ell)\right\}_{\ell=1}^{L}\biggr)$ where $\muy(\y)$ is the class template, $\template(\ell)$ is the rendering template at layer $\ell$, and $\bias(\ell)$ is the parameters of the conjugate prior $p(\z|\y,\x)$ at layer $\ell$, which turn out to be the bias terms in the ReLU after convolutions at each layer in CNNs.
   \\
   \STATE
   \STATE 1. Use Markov chain Monte Carlo method to sample the latent variables $\z(\ell)$ in DGM, $\ell=1,2,\dots,L$ from $\pi_{\z|\y}=\text{Softmax}\left(\frac{1}{\sigma^{2}}\rpn(\y,\z)\right)$, where $\rpn(\y,\z) \triangleq \sum_{\ell=1}^{L}
     \dotprod{\bias(\trans;\ell)}{\sparse(\ell)\odot \h(\ell)}$.
   \STATE
   \STATE 2. Render $\h(\ell)$, $\ell=0,1,\dots,L-1$ using the recursion $\h(\ell - 1) = \sum_{\p\in \pp(\ell)}  \sparse(\ell, \p)\ttrans(t;\ell,\p) \pad(\ell, \p)\template(\ell, \p) \h(\ell,\p)$ in Eqn. \ref{eqn:render-1layer}, in which $\h(L)=\muy(\y)$ and $\ttrans(t;\ell,\p)$ and $\pad(\ell, \p)$ are the local translation matrix and the zero-padding matrix at pixel location $\p$ in layer $\ell$ as described above. 
   \STATE
   \STATE 3. Add Gaussian pixel noise $\N(0,\sigma^{2}\textbf{1}_{D(0)})$ into $\h(0)$ to achieve the final rendered image $\x$, where $D(0)$ is the dimension, i.e. the number of pixels, of $\h(0)$ and $\x$.
\end{algorithmic}
\end{algorithm}

\vspace{2mm}
\noindent {\bf DGM with skip connections:} In Section \ref{sec:jmap-inference} 
below, we will show that CNNs can be derived as a joint maximum a 
posteriori (JMAP) inference in DGM. By introducing skip connections 
into the structure of the rendering matrices $\Lambda(\ell)$, we can 
also derive the convolutional neural networks with skip connections 
including ResNet and DenseNet. The detail derivation of ResNet and DenseNet can be found in Appendix~\ref{sec:unifying-framework}.


\subsection{Inference}
\label{sec:jmap-inference}
We would like to show that inference in 
DGM has the CNN form (see Figure~\ref{fig:lddrm2cnn-details}) and, 
therefore, is still tractable and efficient. This correspondence 
between DGM and CNN helps us achieve two important goals when 
developing a new generative model. First, the desired generative model 
is complex enough with rich needed structures to capture the great 
diversity of forms and
appearances of the objects surrounding us. Second, the inference in the 
model is fast and efficient so that the model can be learned in 
reasonable time. Such advantages justify our modeling choice of using 
classification models like CNNs to inform our design of generative 
models like DGM. 

The following theorem establishes the aforementioned correspondence, 
showing that the JMAP inference of the optimal latent variables $z$ in DGM is indeed a CNN:
\begin{theorem} \label{theorem:jmap_inference_theorem}
Denote $\theta := \biggr(\left\{\muy(\y)\right\}_{\y=1}^{K},\left\{\template(l)\right\}_{l=1}^{L},\left\{\pi_{\y}\right\}_{\y=1}^{K},
\left\{\bias(\ell)\right\}_{l=1}^{L}\biggr)$ the set of all parameters 
in DGM. The JMAP inference of latent variable z in DGM 
is the feedforward step in CNNs. Particularly, we have:
\begin{align}
\max \limits_{\z}\left\{p(z|x,y) \right\} &= \max \limits_{\z}\frac{1}{\sigma^{2}}\left\{\dotprod{\h(\y,\z;0)}{\x} + \rpn(\y,\z) \right\} + \text{const} \nonumber \\
&\ge \frac{1}{\sigma^{2}}\dotprod{\muy(\y)}{\acts(L)} + \text{const} \label{eqn:max-sum-product}
\end{align}
where $\acts(L)$ is computed recursively. In particular, $\acts(0)=\x$ and: \begin{align}
\acts(\ell) = \max \limits_{\z(\ell)} \left\{\Lambda^{\top}(\z;\ell)\acts(\ell - 1) + \bias(\ell)\right\} &=\maxpool\left(\relu\left(\convo\left(\Gamma^{\top}(\ell), \acts(\ell -1)\right)\right) + \bias(\ell) \right) \nonumber \\
&\overset{d}{=}  \maxpool\left(\relu\left(\convo\left(\w(\ell), \acts(\ell -1)\right)\right) + \bias(\ell)\right) \label{eqn:dis-relax}
\end{align}
The equality holds in Eqn.~\ref{eqn:max-sum-product} when the parameters $\theta$ in DGM satisfy the non-negativity assumption that the intermediate rendered image $h(\ell) \ge 0, \ \forall{\ell=1,2,\dots, L}$.
\end{theorem}

There are four takeaways from the results in Theorem~\ref{theorem:jmap_inference_theorem}:
\begin{enumerate}
    \item ReLU non-linearities in CNNs find the optimal value for the 
    template selecting latent variables $\sparse(\ell)$ at each layer 
    $\ell$ in DGM, detecting if particular features exist in the 
    image or not.
    \item MaxPool operators in CNNs find the optimal value for the 
    local translation latent variables $\trans(\ell)$ at each layer $
    \ell$ in DGM, locating where particular features are rendered in the image.
    \item Bias terms after each convolution in CNNs are from the prior distribution of latent variables in the model. Those bias terms 
    update the posterior estimation of latent variables from data 
    using the knowledge encoded in the prior distribution of those 
    latent variables.
    \item Convolutions in CNNs result from reversing the local 
    rendering operators, which use the rendering template $\template(\ell)$, in DGM. 
    Instead of rendering as in DGM, convolutions in CNNs perform 
    template matching. Particularly, it can be shown that convolution 
    weights $W(\ell)$ in CNNs are the transposes of the rendering 
    templates $\template(\ell)$ in DGM.
\end{enumerate}

Table \ref{tbl:correspondence-lddrm-cnn} summarizes the correspondences 
between DGM and CNNs. The proofs for these correspondences are 
postponed to Appendix~\ref{Sec:appendix_C}. The non-negativity assumption
that the intermediate rendered image $\h(\ell) \ge 0, 
\ \forall{\ell=1,2,\dots, L}$ allows us to apply the max-product message 
passing and send the $\max$ over latent variables $\z(\ell)$ operator through the 
product of rendering matrices $\render(\z;\ell)$. Thus, given this assumption, the equality holds in Eqn.~\ref{eqn:max-sum-product}. In Eqn.~\ref{eqn:dis-relax}, 
we have removed the generative constraints inherited from DGM to 
derive the weights $\w(\ell)$ in CNNs, which are free parameters. As a 
result, when faced with training data that violates DGM’s 
underlying assumptions, CNNs will have more freedom to compensate. We 
refer to this process as a discriminative relaxation of a generative 
classifier \cite{ng2002discriminative, bishop2007generative}. Finally, 
the dot product with the object template $\muy(\y)$ in Eqn.~\ref{eqn:max-sum-product} corresponds to the fully connected layer before the 
Softmax non-linearity is applied in CNNs.

Given Theorem~\ref{theorem:jmap_inference_theorem} and the four takeaways above, DGM has successfully reverse-engineered CNNs. However, the impact of Theorem~\ref{theorem:jmap_inference_theorem} goes beyond a reverse-engineering 
effort. First, it provides probabilistic semantics for components in 
CNNs, justifying their usage, and providing an opportunity to employ 
probabilistic inference methods in the context of CNNs. In particular, 
convolution operators in CNNs can be seen as factor nodes in the 
factor graph associated with DGM. Similarly, activations from the 
convolutions in CNNs correspond to bottom-up messages in that factor 
graph. The bias terms added to the activations in CNNs, which are from 
the joint prior distribution of latent variables, are equivalent to 
the top-down messages from the top layers of DGM. These top-down 
messages have receptive fields of the whole image and are used to 
update the bottom-up messages, which are estimated from local information with smaller 
receptive fields. Finally, ReLU non-linearities and MaxPool operators 
in CNNs are max-marginalization operators over the template selecting and local 
translation latent variables $\sparse(\ell)$ and $\trans(\ell)$ in DGM, respectively. These max-marginalization operators are from max-product message passing used to infer the latent variables in DGM.

Second, Theorem~\ref{theorem:jmap_inference_theorem} provides a 
flexible framework to design CNNs. Instead of directly engineering 
CNNs for new tasks and datasets, we can modify DGM to incorporate 
our knowledge of the tasks and datasets into the model and then perform 
JMAP inference to achieve a new CNN architecture. For example, in 
Theorem~\ref{theorem:jmap_inference_theorem}, we show how ReLU can be 
derived from max-marginalization of $\sparse(\ell)$. By changing the 
distribution of $\sparse(\ell)$, we can derive Leaky ReLU. Furthermore, batch normalization in CNNs can be derived from 
DGM by normalizing intermediate rendered images at each layer in DGM. Also, as mentioned above, by introducing skip connections into 
the rendering matrices $\Lambda(\ell)$, we can derive ResNet and 
DenseNet. Details of those derivations can be found in Appendix~\ref{Sec:appendix_B} and \ref{Sec:appendix_C}. 

\subsection{Learning}
\label{sec:learning-in-lddrm}
DGM learns from both labeled and unlabeled data. Learning in DGM 
can be posed as likelihood estimation problems which optimize the 
conditional log-likelihood $\log p(\y|\x,\z)$ and the expected 
complete-data log-likelihood $\Expect_{\z}[\log p(\x, \z|\y)]$ for 
supervised and unsupervised learning respectively. Interestingly, 
\emph{the cross-entropy loss used in training CNNs with labeled data 
is the upper bound of the DGM's negative conditional log-likelihood}. DGM 
solves these likelihood optimization problems via the Expectation-
Maximization (EM) approach. In the E-step, inference in DGM finds 
the optimal latent variables $\z^{*}$. This inference has the form of 
a CNN as shown in Theorem \ref{theorem:jmap_inference_theorem}. In the 
M-step, given $\z^{*}$, DGM maximizes the corresponding likelihood 
objective functions or their lower bounds as in the case of cross-entropy loss. There is no closed-form M-step update for deep models 
like DGM, so DGM employs the generalized EM instead~\cite{Rubin-1977, Bishop:2006:PRM:1162264}. In generalized EM, the M-step seeks to increase value of the likelihood objective function 
instead of maximizing it. In particular, in the M-step, DGM uses 
gradient-based methods such as Stochastic Gradient Descent (SGD)~\cite{robbins1985stochastic, kiefer1952stochastic} to update its 
parameters. The following theorem derives the learning objectives for 
DGM in both supervised and unsupervised settings.

\begin{theorem} \label{theorem:learning_lddrm}
Denote $\theta := \biggr(\left\{\muy(\y)\right\}_{\y=1}^{K},\left\{\template(l)\right\}_{l=1}^{L},\left\{\pi_{\y}\right\}_{\y=1}^{K},\left\{\bias(\ell)\right\}_{l=1}^{L}\biggr)$.
For any $n \geq 1$, let $\x_{1},\ldots,\x_{n}$ be i.i.d. samples from DGM. Assume that the final rendered template $\h(\y,\z;0)$ is normalized such that its norm is constant. The following holds: \\
\noindent \textit{(a) Cross-entropy loss for supervised training CNNs with labeled data:}
\vspace{-2mm}
\begin{align}
\centering
\max_{(\z_{i})_{i=1}^{n}, \theta}\dfrac{1}{n} \sum_{i=1}^{n} \log p(\y_{i}|\x_{i}, \z_{i};\theta) 
\geq  \max \limits_{\theta} \dfrac{1}{n} \sum_{i=1}^{n} \log q(\y_{i}|\x_{i}) = - \min_{\theta \in \mathcal{A}_{\gamma}} H_{p,q}(\y|\x) \label{eqn:xentropy}
\end{align}
where $q(\y|\x)$ is the posterior estimated by CNN, and $H_{p,q}(\y|\x)$ is the cross-entropy between $q(\y|\x)$ and the true posterior $p(\y|\x)$ given by the ground truth.
\\[8pt]
\textit{(b) Reconstruction loss with RPN for unsupervised training of CNNs with labeled and unlabeled data:}
\begin{eqnarray}
\label{eqn:asymptotic_equivalent_unsup}
\centering
\max \limits_{\theta} \dfrac{1}{n} \sum_{i=1}^{n} \Expect\left[\log p(\x_{i}, \z_{i}|\y_{i})\right] \overset{\text{asymp}}{\approx} - \left(\min \limits_{\theta} \dfrac{1}{n} \sum \limits_{i=1}^{n} \dfrac{\| \x_{i} - \h(\y_{i},\z^{*}_{i};0)\|^{2}}{2} + \text{RPN}\right),\, \text{when } \sigma \to 0
\end{eqnarray}
where the latent variable $\z_{i}^{*}$ is estimated by the CNN as described in Theorem \ref{theorem:jmap_inference_theorem}, $\h(\y_{i},\z^{*}_{i};0)$ is the reconstructed image, and the RPN regularization is the negative log prior defined as follows:
\vspace{-2mm}
\begin{align}
    \text{RPN} = - \frac{1}{n}\sum \limits_{i=1}^{n} \log p(\z^{*}_{i}|\y_{i}) = -\frac{1}{n}\sum \limits_{i=1}^{n}\log \text{Softmax}\left(\rpn(\y_{i},\z_{i}^{*}) \right) \label{defn:rpn}.
\end{align}
\vspace{-6mm}
\end{theorem}

\paragraph{Cross-Entropy Loss for Training Convolutional Neural Networks with Labeled Data:} Part (a) of Theorem~\ref{theorem:learning_lddrm} establishes the cross-entropy loss in the 
context of CNNs as an upper bound of the DGM's negative conditional log-likelihood $L_{sup}:=-\frac{1}{n} \sum_{i=1}^{n} \log p(\y_{i}|\x_{i}, 
\z_{i}; \theta)$. Different from other derivations of cross-entropy 
loss via logistic regression, Theorem~\ref{theorem:learning_lddrm}(a) 
derives the cross-entropy loss in conjunction with the architecture of 
CNNs since the estimation of the optimal latent variables $\z^{*}$ is 
part of the optimization in Eqn.~\ref{eqn:xentropy}. In other 
word, Theorem~\ref{theorem:learning_lddrm}(a) ties feature extraction 
and learning for classification in CNNs into an end-to-end conditional 
likelihood estimation problem in DGM. This new interpretation of 
the cross-entropy loss suggests an interesting direction in which 
better losses for training CNNs with labeled data for supervised 
classification tasks can be derived from tighter upper bounds for 
$L_{sup}$. The Max-Min cross-entropy in Section \ref{sec:max_min} is an example. Note that the assumption that the 
rendered image $h(0)$ has constant norm is solely for the ease of presentation. 
Later, in Appendix~\ref{Sec:appendix_B}, we extend the result of Theorem~\ref{theorem:learning_lddrm}(a) to 
the setting in which the norm of rendered image $h(0)$ is bounded. 

In order to estimate how tight the 
cross-entropy upper bound is, we prove
the lower bound for $L_{sup}$. The gap between this lower bound and the 
cross-entropy upper bound suggests the quality of the estimation in 
Theorem~\ref{theorem:learning_lddrm}(a). In particular, this gap is 
given by:
\begin{align}
\dfrac{1}{n} \sum_{i=1}^{n} \biggr\{\max \limits_{\y}\biggr(\max \limits_{\z}\biggr(\dfrac{\dotprod{\h(\y,\z;0)}{\x_{i}} + \rpn(\y,\z)}{\sigma^{2}}\biggr) - \dfrac{\dotprod{\h(\y,\overline{\z};0)}{\x_{i}} + \eta(\y,\overline{\z})}{\sigma^{2}} \biggr)\biggr\} + \log K,
\end{align}
where $\overline{\z}_{i} = \mathop{\arg \max} \limits_{\z'}p(\y|\x_{i},\z';\theta)$ for $1 \leq i \leq n$. More details can be found in Appendix~\ref{Sec:appendix_B} while its detail proof is deferred to Appendix~\ref{Sec:appendix_C}. 

\paragraph{Reconstruction Loss with the Rendering Path Normalization (RPN) Regularization for Unsupervised Learning with Both Labeled and Unlabeled Data:}
Part (b) of Theorem~\ref{theorem:learning_lddrm} suggests that DGM 
learns from both labeled and unlabeled data by maximizing its expected complete-data 
log-likelihood, $\Expect\left[\log p(\x_{i}, \z_{i}|\y_{i};\theta)\right]$, which is the sum of a reconstruction loss and the RPN 
regularization. Deriving the E-step and M-step of generalized EM when 
$\sigma > 0$ is rather complicated; therefore, for the simplicity of 
the paper, we only focus on the setting in which $
\sigma$ goes to 0. Under that setting,  in the M-step, DGM minimizes the objective function $
\dfrac{1}{n} \sum \limits_{i=1}^{n} \dfrac{\| \x_{i} - \h(\y_{i},
\z^{*}_{i};0)\|^{2}}{2}+ \text{RPN}$ with respect to the parameters $
\theta$ of the model. The first term in this objective function is the 
reconstruction loss between the input image $\x_{i}$ and the 
reconstructed template $\h(\y_{i}, \z^{*}_{i};0)$. The second term is 
the Rendering Path Normalization (RPN) defined in Eqn. \ref{defn:rpn}. 
RPN encourages the $(\y_{i}, \z^{*}_{i})$ inferred in the bottom-up E-
step to have higher prior among all possible values of $(\y, \z)$. Due 
to the parametric form of $p_{\z|\y}$ as in Eqn. \ref{eqn:joint-prior}, 
RPN also enforces the dependencies among latent variables $
(\sparse(\ell), \trans(\ell))$ at different layers in DGM. An 
approximation to this RPN regularization is discussed in Appendix~\ref{Section:relax_unsup_LD_DRM}.

\paragraph{Semi-Supervised Learning with the Deconvolutional Generative Model:} DGM learns from both labeled and unlabeled data 
simultaneously by maximizing a weighted combination of the cross-entropy loss for supervised learning and the reconstruction loss with 
RPN regularization for unsupervised learning as in Theorem~\ref{theorem:learning_lddrm}. We now formulate the semi-supervised 
learning problem in DGM. In particular, let $\x_{1},\ldots,\x_{n}$ 
be i.i.d. samples from DGM and assume that the labels $\y_{1}, 
\ldots, \y_{n_{1}}$ are unknown for some $0 \le n_{1} \le n$, DGM 
utilizes the following model to determine optimal parameters employed 
for the semi-supervised classification task: 

\begin{align}
\label{eqn:semisup_objective_ld_drm}
 \min \limits_{\theta} \left\{\dfrac{\alpha_{\text{RC}}}{n} \sum \limits_{i=1}^{n} \biggr( \dfrac{\| \x_{i} - \h(\y_{i},\z^{*}_{i};0)\|^{2}}{2} + \text{RPN}\biggr) - \dfrac{\alpha_{\text{CE}}}{n - n_{1}} \sum \limits_{i=n_{1} + 1}^{n} \log q_{\theta}(\y_{i}|\x_{i})\right\},
\end{align}
where $\alpha_{\text{RC}}$ and $\alpha_{\text{CE}}$ are non-negative weights 
associated with the reconstruction loss/RPN regularization and the cross-entropy loss, respectively. Again, the optimal latent variables $\z_{i}^{*} = \mathop{\arg \max}\limits_{\z'} \biggr(\dotprod{\h(\y_{i},\z';0)}{\x_{i}}+\rpn(\y_{i},\z')\biggr)$ are inferred in the E-step as in Theorem \ref{theorem:jmap_inference_theorem}. For unlabeled data, $y_{i}$ is set to $\mathop{\arg \max}\limits_{\y'} \biggr(\dotprod{\h(\y',\z^{*}_{i};0)}{\x_{i}}+\rpn(\y',\z^{*}_{i}) + \ln \pi_{y'}\biggr)$.\\

In summary, combining Theorem~\ref{theorem:learning_lddrm}(a) and~\ref{theorem:jmap_inference_theorem}, DGM allows us to derive CNNs 
with convolution layer, the ReLU non-linearity, and the MaxPool layer. These CNNs optimize the cross-entropy loss for supervised classification 
tasks with labeled data. Combining Theorem~\ref{theorem:learning_lddrm}(b) and~\ref{theorem:jmap_inference_theorem}, DGM extends the traditional 
CNNs for unsupervised learning tasks in which the
networks optimize the reconstruction loss with the RPN regularization. 
DGM does semi-supervised learning by optimizing the weighted 
combination of the losses in Theorem~\ref{theorem:learning_lddrm}(a) 
and Theorem~\ref{theorem:learning_lddrm}(b). Inference in the semi-supervised learning setup still follows Theorem~\ref{theorem:jmap_inference_theorem}. DGM can also be extended to explain other variants of CNNs, including ResNet and DenseNet, as well as other components in CNNs such as Leaky ReLU and batch normalization.
\section{Statistical Guarantees for the Deconvolutional Generative Model in the Supervised Setting}
\label{sec:stats}
We provide statistical guarantees for DGM to establish that DGM 
is well defined statistically. First, we prove that DGM is 
consistent under a supervised learning setup. Second, we provide a 
generalization bound for DGM, which is proportional to the ratio of 
the number of active rendering paths and the total number of rendering
paths in the trained DGM. A rendering path is a
configuration of all latent variables in DGM as defined in Table~\ref{tbl:notations-lddrm}, and active rendering paths are those
among rendering paths $(\widehat{\y}, \widehat{\z})$ whose corresponding 
rendered image is sufficiently close to one of the data point from the 
input data distribution. Our key results are summarized below. More details and proofs are deferred to Appendix B and C.

Governed by the connection between the cross-entropy and the posterior 
class probabilities $p(\y|\x,\z)$ under DGM, for the supervised 
setting of i.i.d. data $(\x_{1},\y_{1}), \ldots, (\x_{n},\y_{n}) \sim 
Q$, where $Q$ is the data distribution, we utilize the following model 
to determine optimal parameters employed for the classification task
\begin{align}
\label{eqn:sup_objective_ld_drm}
 \min \limits_{\theta} \left\{\dfrac{\alpha_{\text{RC}}}{n} \sum \limits_{i=1}^{n} \biggr( \dfrac{\| \x_{i} - \h(\y_{i},\z^{*}_{i};0)\|^{2}}{2} - \log p(\y_{i},\z^{*}_{i})\biggr) - \dfrac{\alpha_{\text{CE}}}{n} \sum \limits_{i=1}^{n} \log q_{\theta}(\y_{i}|\x_{i})\right\}
\end{align}
where $\z_{i}^{*} = \mathop{\arg \max}\limits_{\z'} 
\biggr(\dotprod{\h(\y_{i},\z';0)}{\x_{i}}+\eta(\y_{i},\z')\biggr)$, $
\alpha_{\text{RC}}$ and $\alpha_{\text{CE}}$ are non-negative weights 
associated with the reconstruction loss and the cross-entropy loss 
respectively. Here, the approximate posterior $q_{\theta}(\y|\x) = 
\text{Softmax} \biggr(\max \limits_{\z}\biggr(\dotprod{\h(\y,\z;0)}{\x}+
\eta(\y,\z)\biggr) + \log \pi_{\y}\biggr)$ is chosen according to 
Theorem \ref{theorem:learning_lddrm}(a) under the regime $\sigma \to 0$. 
The optimal solutions of objective function \eqref{eqn:sup_objective_ld_drm} induce a corresponding set of optimal 
(active) rendering paths that play a central role 
for an understanding of generalization bound regarding the 
classification tasks. 

Before proceeding to the generalization bound, we first
state the informal result regarding the consistency of optimal 
solutions of~\eqref{eqn:sup_objective_ld_drm} when the sample size $n$ 
goes to infinity.
\begin{theorem} (Informal) \label{proposition:consistency_sup_ld_drm}
Under the appropriate conditions regarding parameter spaces of $\theta$, the optimal solutions of objective function \eqref{eqn:sup_objective_ld_drm} converge almost surely to those of the following population objective function
\begin{align}
\min \limits_{\theta} \left\{\alpha_{\text{RC}} \biggr(\int \biggr( \dfrac{\| x - \h(\y,\z^{*};0)\|^{2}}{2} - \log p(\y,\z^{*})\biggr)dQ(\x,\y)\biggr)- \alpha_{\text{CE}} \int \log q_{\theta}(\y|\x)dQ(\x,\y)\right\} \nonumber
\end{align}
where $\z^{*} = \mathop{\arg \max}\limits_{z'} \biggr(\dotprod{\h(\y,\z';0)}{\x}+\eta(\y,\z')\biggr).$
\end{theorem}
In Appendix C, we 
provide detail formulations of Theorem~\ref{proposition:consistency_sup_ld_drm} for the supervised learning. 
Additionally, the detail proof of this result is presented in Appendix D. The 
statistical guarantee regarding optimal solutions of 
\eqref{eqn:sup_objective_ld_drm} validates their usage for the 
classification task. Given that DGM is consistent under a supervised learning setup, the following theorem establish a generalization bound for the model.
\begin{theorem} (Informal)
\label{theorem:generalization_bound_informal}
Let $L_{\mathcal{A}}$ and $L_{\mathcal{D}}$ denote the population and empirical losses on the data population $\mathcal{A}$ and the training set $\mathcal{D}$ of DGM, respectively. Under the margin-based loss, the generalization gap of the classification framework with optimal solutions from \eqref{eqn:sup_objective_ld_drm} is controlled by the following term
\begin{align}
L_{\mathcal{A}} \lesssim L_{\mathcal{D}} +  \dfrac{8K(2K-1)}{\sqrt{n}}\biggr(2\overline{\sratio}_{n}|\mathcal{L}|(R^{2}+1)+|\log \overline{\gamma}| \biggr) + \sqrt{\dfrac{\log(2\delta^{-1})}{2n}}\nonumber
\end{align}
with probability $1-\delta$. Here, $\overline{\sratio}_{n} \in (0,1)$ 
denotes the ratio of active optimal rendering paths among all the 
optimal rendering paths, $|\mathcal{L}|$ is the total number of 
rendering paths, $\overline{\gamma}$ is the lower bound of prior 
probability $\pi(\y)$ regarding labels, and $R$ is the radius of the 
sphere that the rendered images belong to.
\end{theorem} 
The detail formulations of the above theorem are postponed to Appendix~\ref{Sec:appendix_C}. The dependence of generalization bound on the 
number of active rendering paths $\overline{\sratio}_{n}|\mathcal{L}|$ helps to justify our 
modeling assumptions. In particular, DGM helps to reduce the number of active rendering paths thanks to the dependencies 
among its latent variables, thereby tightening the generalization 
bound. 
Nevertheless, there is a limitation regarding the current generalization bound. In particular, the bound involves the number of rendering paths $|\mathcal{L}|$, which is usually large. This is mainly because our bound has not fully taken into 
account the structures of CNNs,
which is the limitation shared among other latest generalization bounds for CNN. It is interesting to explore if techniques in  works by
\cite{Bartlett-2017} and \cite{Rakhlin-2018} can be employed to improve the term $|\mathcal{L}|$ in
our bound.

\paragraph{Extension to unsupervised and semi-supervised settings} 
Apart from the statistical guarantee and generalization bound 
established for the supervised setting, we also provide careful 
theoretical studies as well as detailed proofs regarding these results 
for the unsupervised and semi-supervised setting in Appendix~\ref{Sec:appendix_B}, Appendix~\ref{Sec:appendix_C},
Appendix~\ref{Sec:appendix_D}, and Appendix~\ref{Sec:appendix_E}. 
\section{New Max-Min Cross Entropy From The Deconvolutional Generative Model} 
\label{sec:max_min}
\begin{figure}[h]
\centering
    \includegraphics[width=0.8\textwidth]{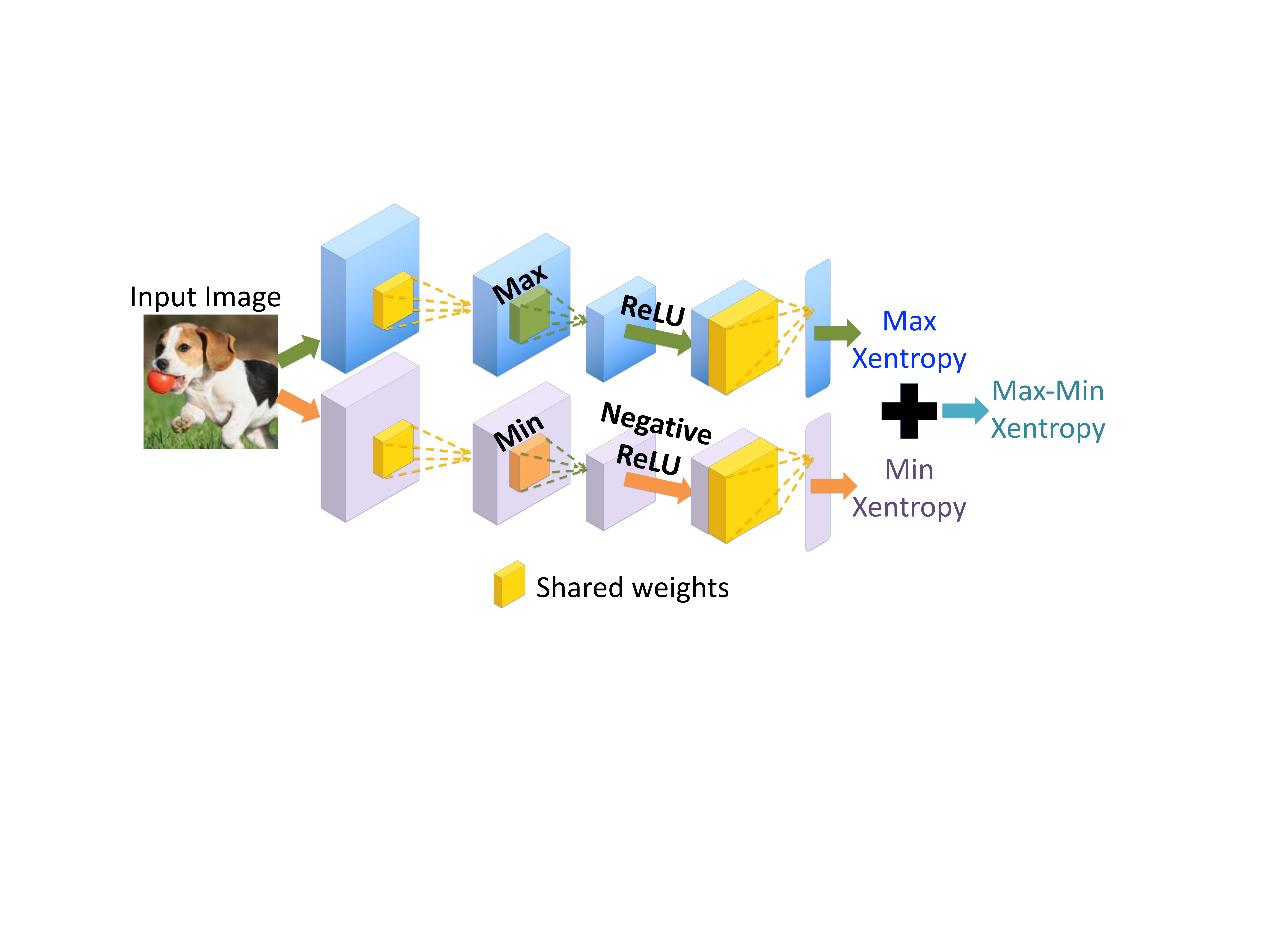} 
    \caption{The Max-Min network. The Max branch in the Max-Min network maximizes the correct target posterior while the Min branch minimizes the incorrect target posterior. These two networks share weights. The Max-Min cross-entropy loss is the weighted average of the cross-entropy losses from the Max and the Min networks.}
    \label{fig:max-min}
\end{figure}

In this section, we explore a particular way to derive an alternative to cross-entropy inspired by the results in Theorem \ref{theorem:learning_lddrm}(a). In particular, denoting $z^{\text{max}} \triangleq \mathop{\arg \max}_{\z}\left\{\dotprod{\h(\y,\z;0)}{\x} + \rpn(\y,\z) \right\}$ and $z^{\text{min}} \triangleq \mathop{\arg \min}_{\z}\left\{\dotprod{\h(\y,\z;0)}{\x} + \rpn(\y,\z) \right\}$, the new cross-entropy $H^{M\&M}$, which is called the \emph{Max-Min cross-entropy}, is the weighted average of the cross-entropy losses from $z^{\text{max}}$ and $z^{\text{min}}$:
\vspace{-0.5mm}
\begin{align*}
    H^{\text{M\&M}} \triangleq  \alpha^{\text{max}}H_{p,q}(y|x,z^{\text{max}}) + \alpha^{\text{min}}H_{p,q}(y|x,z^{\text{min}}) = \alpha^{\text{max}}H^{\text{max}}_{p,q}(y|x) + \alpha^{\text{min}}H^{\text{min}}_{p,q}(y|x). 
\end{align*}
Here the Max cross-entropy $H^{\text{max}}_{p,q}$ and Min cross entropy $H^{\text{min}}_{p,q}$  maximizes the correct target posterior and minimizes the incorrect target posterior, respectively. Similar to the cross-entropy loss, the Max-Min cross-entropy can also be shown to be an upper bound to the negative conditional log-likelihood $L_{\text{sup}}$ of the DGM and has the same generalization bound derived in Section \ref{sec:stats}. The Max-Min networks in Figure~\ref{fig:max-min} realize this new loss. These networks have two CNN-like branches that share weights. The max branch estimates $z^{\text{max}}$ using ReLU and Max-Pooling, and the min branch estimates $z^{\text{min}}$ using the Negative ReLU, i.e., $\min(\cdot,0)$, and Min-Pooling. The Max-Min networks can be interpreted as a form of knowledge distillation like the Born Again networks~\cite{Furlanello-2018} and the Mean Teacher networks. However, instead of a student network learning from a teacher network, in Max-Min networks, two students networks, the Max and the Min networks, cooperate and learn from each other during the training.

\section{Experiments}
\label{sec:experiments}
\subsection{Semi-Supervised Learning}
\label{ssl-experiments}
We show DGM armed with Max-Min cross-entropy and Mean Teacher regularizer achieves SOTA on benchmark datasets. We discuss the experimental results for CIFAR10 and CIFAR100 here. The results for SVHN, training losses, and training details, can be found in the Appendix~\ref{Sec:appendix_A} \& \ref{Sec:appendix_F}.
\paragraph{CIFAR-10:}
\begin{table}[t!]
\centering
\small
\caption{\label{tbl:cifar10}%
Error rate percentage on CIFAR-10 over 3 runs.
}
\begin{tabular}{ l l l l l }
\makecell[lb]{} &\
\makecell[lb]{\ze1K labels\\50K images} &\
\makecell[lb]{\ze2K labels\\50K images} &\
\makecell[lb]{\ze4K labels\\50K images} &\
\makecell[lb]{50K labels\\50K images}\\
\Xhline{1pt}\noalign{\smallskip}
Adversarial Learned Inference \cite{dumoulin2016adversarially} &\
$ 19.98 \pm 0.89 $ & \tss$ 19.09 \pm 0.44 $ & \tss$ 17.99 \pm 1.62 $ & \tss$ $ \\
Improved GAN \cite{salimans2016improved} &\
$ 21.83 \pm 2.01 $ & \tss$ 19.61 \pm 2.09 $ & \tss$ 18.63 \pm 2.32 $ & \tss$ $ \\
Ladder Network \cite{rasmus2015semi} &\ $ $ & \tss$ $ & \tss$ 20.40 \pm 0.47 $ & \tss$ $ \\
$\Pi$ model \cite{laine2016temporal} &\
$ 27.36 \pm 1.20 $ & \tss$ 18.02 \pm 0.60 $ & \tss$ 13.20 \pm 0.27 $ & \tss$ 6.06 \pm 0.11 $ \\
\multicolumn{2}{l}{Temporal Ensembling \cite{laine2016temporal}} &\
\tss$ $ & \tss$ 12.16 \pm 0.31 $ & \tss$ 5.60 \pm 0.10 $ \\
Mean Teacher \cite{tarvainen2017mean} &\ 
$ 21.55 \pm 1.48 $ & \tss$ 15.73 \pm 0.31 $ & \tss$ 12.31 \pm 0.28 $ & \tss$ 5.94 \pm 0.15 $ \\
VAT+EntMin \cite{miyato2017virtual} &\
$ $ & \tss$ $ & \tss$ \pmb{10.55} $ & \tss$ $ \\
DRM \cite{patel2016probabilistic, patel2015arxiv} &\
$ 27.67 \pm 1.86 $ & \tss$ 20.71 \pm 0.30 $ & \tss$ 15.36 \pm 0.34 $ & \tss$ 5.75 \pm 0.24 $ \\
\Xhline{1pt}\noalign{\smallskip}
Supervised-only &\
$ 46.43 \pm 1.21 $ & \tss$ 33.94 \pm 0.73 $ & \tss$ 20.66 \pm 0.57 $ & \tss$ 5.82 \pm 0.15 $ \\
DGM without RPN &\
$ 24.88 \pm 0.76 $ & \tss$ 18.97 \pm 0.80 $ & \tss$ 14.41 \pm 0.19 $ & \tss$ 5.57 \pm 0.07 $ \\
DGM+RPN &\
$ 24.48 \pm 0.43 $ & \tss$ 18.62 \pm 0.70 $ & \tss$ 14.18 \pm 0.46 $ & \tss$ 5.35 \pm 0.08 $ \\
DGM+RPN+Max-Min &\
$ 21.55 \pm 0.46 $ & \tss$ 16.24 \pm 0.17 $ & \tss$ 12.50 \pm 0.35 $ & \tss$ \pmb{4.85 \pm 0.10} $ \\
DGM+RPN+Max-Min+Mean Teacher &\
$ \pmb{19.79 \pm 0.74} $ & \tss$ \pmb{15.11 \pm 0.51} $ & \tss$ 11.81 \pm 0.13 $ & \tss$ 4.88 \pm 0.09 $ \\
\Xhline{1pt}
\end{tabular}
\end{table}
Table \ref{tbl:cifar10} shows comparable results of DGM to SOTA methods. DGM is also better than the best methods that do not use consistency regularization like GAN, Ladder network, and ALI when using only $N_l$=2K and 4K labeled images. DGM outperform DRM in all settings. Also, among methods in our comparison, DGM achieves the best test accuracy when using all available labeled data ($N_l$=50K). We hypothesize that DGM has the advantage over consistency regularization methods like Temporal Ensembling and Mean Teacher when there are enough labeled data is because the consistency regularization in those methods tries to match the activations in the network, but does not take into account the available class labels. On the contrary, DGM employs the class labels, if they are available, in its reconstruction loss and RPN regularization as in Eqns. \ref{eqn:asymptotic_equivalent_unsup} and \ref{defn:rpn}. 
In all settings, RPN regularizer improves DGM performance. Even though the improvement from RPN is small, it is consistent across the experiments. Furthermore, using Max-Min cross-entropy significantly reduces the test errors. When combining with Mean-Teacher, our Max-Min DGM improves upon Mean-Teacher and consistently achieves either SOTA results or second best results in all settings. This consistency in performance is only observed in our method and Mean-Teacher. Also, like with Mean-Teacher, DGM can potentially be combined with other consistency regularization methods, e.g., the Virtual Adversarial Training (VAT) \cite{miyato2017virtual}, to obtain better results.
\paragraph{CIFAR-100:}
Table \ref{tbl:cifar100} shows DGM's comparable results to $\Pi$ model and Temporal Ensembling, as well as better results than DRM. Same as with CIFAR10, using the RPN regularizer results in a slightly better test accuracy, and DGM achieves better results than $\Pi$ model and Temporal Ensembling method when using all available labeled data. 
Notice that combining with Mean-Teacher just slightly improves DGM's performance when training with 10K labeled data. This is again because consistency regularization methods like Mean-Teacher do not add much advantage when there are enough labeled data. However, DGM+Max-Min still yields better test errors and achieves SOTA result in all settings. Note that since combining with Mean-Teacher does not help much here, we only show result for DGM+Max-Min. 
\begin{table}[h]
\centering
\small
\caption{\label{tbl:cifar100}%
Error rate percentage on CIFAR-100 over 3 runs.
}
\begin{tabular}{ l l l l l}
\multicolumn{2}{l}{\makecell[lb]{}} &\
\makecell[cb]{10K labels\\50K images} &\
\makecell[cb]{50K labels\\50K images} \\
\Xhline{1pt}\noalign{\smallskip}
\multicolumn{2}{l}{$\Pi$ model \cite{laine2016temporal}}  &\
$ 39.19 \pm 0.36 $ &
\tss$ 26.32 \pm 0.04$ \\
\multicolumn{2}{l}{Temporal Ensembling \cite{laine2016temporal}} &\ $ 38.65 \pm 0.51$ &
\tss$ 26.30 \pm 0.15$ \\
\multicolumn{2}{l}{DRM \cite{patel2016probabilistic, patel2015arxiv}} &\
$ 41.09 \pm 0.31 $ & \tss$ 27.06 \pm 0.19 $ \\
\Xhline{1pt}\noalign{\smallskip}
\multicolumn{2}{l}{Supervised-only} &\
$ 44.56 \pm 0.30 $ & \tss$ 26.42 \pm 0.17 $ \\
\multicolumn{2}{l}{DGM without RPN} &\
$ 40.70 \pm 1.13 $ & \tss$ 26.27 \pm 0.09 $ \\
\multicolumn{2}{l}{DGM+RPN} &\
$ 39.85 \pm 0.46 $ & \tss$ 25.84 \pm 0.10 $ \\
\multicolumn{2}{l}{DGM+RPN+Mean Teacher} &\
$ 39.84 \pm 0.32 $ & \tss$ 25.98 \pm 0.35 $ \\
\multicolumn{2}{l}{DGM+RPN+Max-Min} &\
$ \pmb{37.75 \pm 0.66} $ & \tss$ \pmb{24.38 \pm 0.29} $ \\
\Xhline{1pt}
\end{tabular}
\end{table}

\subsection{Supervised Learning with Max-Min Cross-Entropy}
The Max-Min cross-entropy can be applied not only to improve semi-supervised learning on deep models including CNNs but also to enhance their supervised learning performance. In our experiments, we indeed observe Max-Min cross-entropy reduces the test error for supervised object classification on CIFAR10. In particular, using the Max-Min cross-entropy loss on a 29-layer ResNet \cite{xie2017aggregated} trained with the Shake-Shake regularization \cite{gastaldi2017shake} and Cutout data augmentation \cite{devries2017improved}, we are able to achieve SOTA test error of 2.30\% on CIFAR10, an improvement of 0.26\% over the test error of the baseline architecture trained with the traditional cross-entropy loss. While 0.26\% improvement seems small, it is a meaningful enhancement given that our baseline architecture (ResNeXt + Shake-Shake + Cutout) is the second best model for supervised learning on CIFAR10. Such small improvement over an already very accurate model is significant in applications in which high accuracy is demanded such as self-driving cars or medical diagnostics. 
Similarly, we observe Max-Min improves the top-5 test error of the Squeeze-and-Excitation ResNeXt-50 network \cite{hu2017squeeze} on ImageNet by 0.17\% compared to the baseline (7.04\% vs. 7.21\%).  For a fair comparison, we re-train the baseline models and report the scores in the re-implementation.
\section{Discussion}
\label{sec:conclusion}
We present the DGM, a general and an effective framework for semi-supervised learning that combines generation and prediction in an end-to-end optimization. Using DGM, we can explain operations used in CNNs and develop new features that help learning in CNNs. For example, we derive the new Max-Min cross-entropy loss for training CNNs, which outperforms the traditional cross-entropy.

In addition to the results discussed above, there are still many open problems related to DGM that have not been addressed in the paper. We give several examples below:
\begin{itemize}
\item An adversarial loss like in GANs can be incorporated into the DGM so that the model can generate realistic images. Furthermore, more knowledge of image generation from graphics and physics can be integrated in DGM so that the model can employ more structures to help learning and generation.
\item The unsupervised and (semi)-supervised models that we consider throughout the paper are under the assumption that the noise $\sigma$ of DGM goes to 0. Governed by this assumption, we are able to derive efficient inference algorithms as well as rigorous statistical guarantees with these models. For the setting that $\sigma$ is not close to 0, the inference with these models will rely on vanilla Expectation-Maximization (EM) algorithm for mixture models to obtain reliable estimators for the rendering templates. Since the parameters of interest are being shared among different rendering templates and have high dimensional structures, it is of practical interest to develop efficient EM algorithm to capture these properties of parameters under that setting of $\sigma$ in DGM. 
\item Thus far, the statistical guarantees with parameter estimation in the paper are established under the ideal assumptions that the optimal global solutions are obtained. However, it happens in practice that the inference algorithms based on SGD with the unsupervised and (semi)-supervised models usually lead to (bad) local minima. As a consequence, investigating sufficient conditions for the inference algorithms to avoid being trapped at bad local minima is an important venue of future work.
\item DGM hinges upon the assumption that the data is generated from mixture of Gaussian distributions with the mean parameters characterizing the complex rendering templates. However, it may happen in reality that the underlying distribution of each component of mixtures is not Gaussian distribution. Therefore, extending the current understandings with DGM under Gaussian distribution to other choices of underlying distributions is an interesting direction to explore.
\end{itemize}

 We would like to end the paper with a remark that DGM is a flexible framework that enables us to introduce new components in the generative process and the corresponding features for CNNs can be derived in the inference. This hallmark of DGM provide a more fundamental and systematic way to design and study CNNs.
 
\section{Acknowledgements}
\label{sec:acknowledge}
First of all, we are very grateful to Amazon AI for providing a highly stimulating research environment for us to start this research project and further supporting our research through their cloud credits program. We would also like to express our sincere thanks to Gautam Dasarathy for great discussions. Furthermore, we would also like to thank Doris Y. Tsao for suggesting and providing references for connections between our model and feedforward and feedback connections in the brain. 

Many people during Tan Nguyen's internship at Amazon AI have helped by providing comments and suggestions on our work, including Stefano Soatto, Zack C. Lipton, Yu-Xiang Wang, Kamyar Azizzadenesheli, Fanny Yang, Jean Kossaifi, Michael Tschannen, Ashish Khetan, and Jeremy Bernstein. We also wish to thank Sheng Zha who has provided immense help with MXNet framework to implement our models.

Finally, we would like to thank members of DSP group at Rice, Machine Learing group at UC Berkeley, and Anima Anandkumar's TensorLab at Caltech who have always been supportive throughout the time it has taken to finish this project.
\newpage
\setlength{\bibsep}{0pt plus 0.3ex}
\bibliography{Arxiv_LD_DRM}
\newpage
\appendix
\begin{center}
\textbf{\Large{Supplementary Material}}
\end{center}
\appendix
\section{Appendix A}
\label{Sec:appendix_A}
This appendix contains semi-supervised learning results of DGM on SVHN compared to other methods.

\vspace{3mm}
\noindent
\textbf{Semi-Supervised Learning Results on SVHN:}
\begin{table}[h!]
\small
\centering
\caption{\label{tbl:svhn}%
Error rate percentage on SVHN in comparison with other state-of-the-art methods. All results are averaged over 2 runs (except for DGM+RPN when using all labels, 1 run)
}
\begin{tabular}{ l l l l l }
\noalign{\bigskip}
\makecell[lb]{} &\
\makecell[lb]{250 labels\\73257 images} &\
\makecell[lb]{500 labels\\73257 images} &\
\makecell[lb]{1000 labels\\73257 images} &\
\makecell[lb]{73257 labels\\73257 images}\\
\Xhline{1pt}\noalign{\smallskip}
ALI \cite{dumoulin2016adversarially} &\
\tss $ $ & \tss $ $ & \tss $ 7.42 \pm 0.65 $ & \tss $ $ \\
Improved GAN \cite{salimans2016improved} &\
\tss $ $ & \tss $ 18.44 \pm 4.8$ & \tss $8.11 \pm 1.3$ & \\
\quad + Jacob.-reg + Tangents \cite{kumar2017semi} &\
\tss$ $ & \tss$4.87 \pm 1.60 $ & \tss$4.39 \pm 1.20 $ & \tss$ $ \\
$\Pi$ model \cite{laine2016temporal} &\
\tss $  9.69 \pm 0.92 $ & \tss $ 6.83 \pm 0.66 $ & \tss $  4.95 \pm 0.26 $ & \tss $ 2.50 \pm 0.07 $ \\
Temporal Ensembling \cite{laine2016temporal} &\
&\tss $ 5.12 \pm 0.13 $ & \tss $ 4.42 \pm 0.16 $ & \tss $ 2.74 \pm 0.0 6$ \\
Mean Teacher \cite{tarvainen2017mean} &\
\tss $ 4.35 \pm 0.50 $ & \tss $ 4.18 \pm 0.27 $ & \tss $  3.95 \pm 0.19 $ & \tss $ \pmb{ 2.50 \pm 0.05} $ \\
VAT+EntMin \cite{miyato2017virtual} &\
\tss $ $ & \tss $ $ & \tss $ 3.86 $ & \tss $ $ \\
DRM \cite{nguyen2016semi} &\
 & \tss$ 9.85 $ & \tss$ 6.78 $ & \\
\Xhline{1pt}\noalign{\smallskip}
Supervised-only &\  
\tss $27.77 \pm 3.18$ & \tss $16.88 \pm 1.30$ & \tss $12.32 \pm 0.95$ & \tss $2.75 \pm 0.10$ \\
DGM without RPN &\
\tss$ 9.78 \pm 0.24 $ & \tss$ 7.42 \pm 0.61  $ & \tss$5.64 \pm 0.13 $ & \tss$ 3.46 \pm 0.04 $ \\
DGM+RPN &\
\tss$ 9.28 \pm 0.01 $ & \tss$ 6.56 \pm 0.88  $ & \tss$5.47 \pm 0.14 $ & \tss$ 3.57 $ \\
DGM+RPN+Max-Min+MeanTeacher &\
\tss$ \pmb{3.97 \pm 0.21} $ & \tss$ \pmb{3.84 \pm 0.34}  $ & \tss$ \pmb{3.70 \pm 0.04} $ & \tss$ 2.87 \pm 0.05 $ \\
\Xhline{1pt}\noalign{\smallskip}
\end{tabular}
\end{table}

\newpage
\section{Appendix B}
\label{Sec:appendix_B}
In this appendix, we give further connection of DGM to cross entropy as well as additional derivation of DGM to various models under both unsupervised and (semi)-supervised setting of data being mentioned in the main text. We also formally present our results on consistency and generalization bounds for DGM in supervised and semi-supervised learning settings. In addition, we explain how to extend DGM to derive ResNet and DenseNet. For the simplicity of the presentation, we denote $\theta=\biggr(\left\{\muy(\y)\right\}_{\y=1}^{K},\left\{\template(\ell)\right\}_{\ell=1}^{L},\left\{\pi_{\y}\right\}_{\y=1}^{K},\left\{\bias(\ell)\right\}_{\ell=1}^{L}\biggr)$ to represent all the parameters that we would like to estimate from DGM where $\mathcal{L}$ is the set of all possible values of latent (nuisance) variables $\z=(\trans(\ell),\sparse(\ell))_{\ell=1}^{L}$. Additionally, for each $(\y,\z) \in \mathcal{J} := \left\{1,\ldots,K\right\} \times \mathcal{L}$, we denote $\theta_{\y,\z} = \biggr(\muy(\y),\left\{\template(\ell)\right\}_{\ell=1}^{L},\left\{\bias(\ell)\right\}\biggr)$, i.e., the subset of parameters corresponding to specific label $\y$ and latent variable $\z$. Furthermore, to stress the dependence of $\eta(\y,\z)$ on $\theta$, we define the following function
\begin{eqnarray}
\tau(\theta_{\y,\z}) : = \eta(\y,\z) = \sum \limits_{\ell=1}^{L} \bias^{\top}(\ell)\left(\sparse(\ell)\odot z(\ell)\right) = \sum \limits_{\ell=1}^{L} \bias^{\top}(\ell) M(\sparse;\ell) z(\ell)\nonumber
\end{eqnarray}
for each $(\y,\z) \in \mathcal{J}$ where $M(\sparse;\ell) = \text{diag}(\sparse(\ell))$ is a masking matrix associated with $\sparse(\ell)$. Throughout this supplement, we will use $\tau(\theta_{\y,\z})$ and $\eta(\y,\z)$ interchangeably as long as the context is clear. Furthermore, we assume that $\template(\ell) \in \theta_{\ell}$, which is a subset of $\mathbb{R}^{F(\ell) \times D(\ell)}$ for any $1 \leq  \ell \leq L$,  $\muy(\y) \in\Omega$, which is a subset of $\mathbb{R}^{D(\ell)}$, for $1 \leq \y \leq K$, and $\bias(\trans;\ell) \in \Xi(\ell)$, which is a subset of $\mathbb{R}^{D(\ell)}$ for all choices of $\trans(\ell)$ and $1 \leq  \ell \leq L$. Last but not least, we say that $\theta$ satisfies the \textit{non-negativity assumption} if the intermediate rendered images $z(\ell)$ satisfy $z(\ell) \geq 0$ for all $1 \leq  \ell \leq L$. Finally, we use $A^{\top}$ to denote transpose of the matrix $A$.
\subsection{Connection between DGM and cross entropy} \label{Section:further_connection_LD_DRM_Cross_entropy}
As being established in part (a) of Theorem \ref{theorem:learning_lddrm}, the cross entropy is the lower bound of maximizing the conditional log likelihood. In the following full theorem, we will show both the upper bound and the lower bound of maximizing the conditional log likelihood in terms of the cross entropy.

\begin{theorem} \label{theorem:cross_entropy_LD_DRM_full} 
Given any $\gamma>0$, we denote $\mathcal{A}_{\gamma} = \left\{\theta: \ \|\h(\y,\z;0)\| = \gamma \right\}$. For any $n \geq 1$ and $\sigma>0$, let $\x_{1},\ldots,\x_{n}$ be i.i.d. samples from the DGM. Then, the following holds \\
(a) (Lower bound)
\begin{eqnarray}
& & \hspace{- 6 em} \max_{(\z_{i})_{i=1}^{n}, \theta \in \mathcal{A}_{\gamma}}\dfrac{1}{n} \sum_{i=1}^{n} \log p(\y_{i}|\x_{i}, \z_{i};\theta) \nonumber \\
& \geq & \max_{\theta \in \mathcal{A}_{\gamma}} \dfrac{1}{n} \sum_{i=1}^{n} \log \biggr(\text{Softmax}\biggr(\max_{\z_{i}}\biggr(\dfrac{\h^{\top}(\y_{i}, \z_{i}; 0)\x_{i} + \eta(\y_{i},\z_{i})}{\sigma^{2}}\biggr) + \bias_{\y_{i}} \biggr)\biggr) \nonumber \\
& = &  \max_{\theta \in \mathcal{A}_{\gamma}} \dfrac{1}{n} \sum_{i=1}^{n} \log q(\y_{i}|\x_{i}) = - \min_{\theta \in \mathcal{A}_{\template}} H_{p,q}(\y|\x) \nonumber
\end{eqnarray}
where $\bias_{\y} = \log \pi_{\y}$ for all $1 \leq \y \leq K$, $q(\y|\x) = \text{Softmax} \biggr(\max \limits_{\z}\biggr(\h^{\top}(\y,\z;0)\x+\eta(\y,\z)\biggr)/\sigma^{2}+\bias_{\y}\biggr)$ for all $(\x,\y)$, and $H_{p,q}(\y|\x)$ is the cross-entropy between the estimated posterior $q(\y|\x)$ and the true posterior given by the ground-true labels $p(\y|\x)$. \\
(b) (Upper bound)
\begin{eqnarray}
& & \hspace{- 6 em} \max_{(\z_{i})_{i=1}^{n}, \theta \in \mathcal{A}_{\gamma}}\dfrac{1}{n} \sum_{i=1}^{n} \log p(\y_{i}|\x_{i}, \z_{i};\theta) \nonumber \\
& & \leq  \max_{\theta \in \mathcal{A}_{\gamma}} \dfrac{1}{n} \sum_{i=1}^{n} \biggr\{\log q(\y_{i}|\x_{i}) + \max \limits_{\y}\biggr(\max \limits_{\z_{i}}\biggr(\dfrac{\h^{\top}(\y,\z_{i}; 0)\x_{i} + \eta(\y,\z_{i})}{\sigma^{2}}\biggr) \nonumber \\ 
& & \hspace{12 em} - \dfrac{\h^{\top}(\y,\overline{\z}_{i}; 0)\x_{i} + \eta(\y,\overline{\z}_{i})}{\sigma^{2}} \biggr)\biggr\} + \log K \nonumber
\end{eqnarray}
where $\overline{\z}_{i} = \mathop{\arg \max} \limits_{\z_{i}} p(\y_{i}|\x_{i},\z_{i};\vec
{\theta})$ for $1 \leq i \leq n$. 
\end{theorem}
\begin{remark}
As being demonstrated in Theorem \ref{theorem:jmap_inference_theorem}, $\max \limits_{\z}\left(\h^{\top}(\y,\z;0)\x + \eta(\y,\z)\right)/\sigma^{2}$, approximately, has the form of a CNN. If we further have the non-negativity assumption with $\theta$, then this is exact. Therefore, the cross entropy $H_{p,q}$ obtained in Theorem \ref{theorem:cross_entropy_LD_DRM_full} has a strong connection with CNN. 
\end{remark}
\begin{remark}
The gap between the upper bound and the lower bound of maximizing the conditional log likelihood in terms of cross entropy function suggests how good the estimation in Theorem~\ref{theorem:cross_entropy_LD_DRM_full} is. In particular, this gap is given by:
\begin{align}
\dfrac{1}{n} \sum_{i=1}^{n} \biggr\{\underbrace{\max \limits_{\y}\biggr(\max \limits_{\z_{i}}\biggr(\dfrac{\h^{\top}(\y,\z_{i}; 0)\x_{i} + \eta(\y,\z_{i})}{\sigma^{2}}\biggr) - \dfrac{\h^{\top}(\y,\overline{\z}_{i}; 0)\x_{i} + \eta(\y,\overline{\z}_{i})}{\sigma^{2}} \biggr)}_{\text{trade-off loss}} \biggr\} + \log K, \nonumber
\end{align}
where $\overline{\z}_{i} = \mathop{\arg \max} \limits_{\z_{i}} p(\y_{i}|\x_{i},\z_{i};\theta)$ for $1 \leq i \leq n$. As long as the number of labels is not too large and the trade-off loss is sufficiently small, the gap between the upper bound and the lower bound in Theorem \ref{theorem:cross_entropy_LD_DRM_full} is small. 
\end{remark}
\subsection{Learning in the DGM without noise for unsupervised setting under non-negativity assumption} \label{Section:relax_unsup_LD_DRM} 
To ease the presentation of the inference with DGM without noise, we first assume that rendered images $\h(\y,\z;0)$ satisfy the non-negativity assumption (Later in Section \ref{Section:relaxation_inference_non_negativity}, we will discuss the relaxation of this assumption for the inference of the DGM). With this assumption, as being demonstrated in Theorem \ref{theorem:jmap_inference_theorem}, we have:
\begin{eqnarray}
\hspace{-2 em} \max \limits_{\y,\z}\left\{\h^{\top}(\y,\z;0)\x + \eta(\y,\z) \right\} = \max \limits_{\y} \h^{\top}(\y)\acts(L) \label{eqn:CNN_structure}
\end{eqnarray} 
where we define
\begin{align}
\acts(L) & = & \max \limits_{\z(L)} \render^{\top}(\z(L))\left(\max \limits_{\z(L-1)}\left( \render^{\top}(\z;L-1) \cdots \left(\max \limits_{\z(1)} \render^{\top}(\z;1)\x + \bias(1)\right)\ldots \right)+\bias(L-1)\right) \nonumber \\
& & \hspace{16 em} +\bias(\ell). \nonumber
\end{align} 
Now, we will provide careful derivation of part (b) of Theorem \ref{theorem:learning_lddrm} in the main text. Remind that, for the unsupervised setting, we have data $\x_{1},\ldots,\x_{n}$ are i.i.d. samples from DGM. The complete-data log-likelihood of the DGM is given as follows:
\begin{align}
\Expect_{y,z}\left[\log p(\x, (\y, \z))\right] &= \sum_{i=1}^{n}\sum_{(\y,\z) \in \Jcal} P(\y,\z|\x_{i})\{\log \pi_{\y,\z} + \log{\N(\x_{i}|\h(\y,\z;0))}\} \nonumber
\end{align}
where we have
\begin{eqnarray}
P(\y,\z|\x_{i})  & = &  \dfrac{\pi_{\y,\z} \N(\x_{i}|\h(\y,\z;0))}{\sum \limits_{(\y',\z') \in \Jcal} \pi_{\y',\z'} \N(\x_{i}|\h(\y',\z';0))} \nonumber \\ 
& = & \dfrac{\pi_{\y}\exp\biggr(-\dfrac{\|\x_{i}-\h(\y,\z;0)\|^{2}-2\eta(\y,\z)}{2\sigma^{2}}\biggr)}{\sum \limits_{(\y',\z') \in \Jcal} \pi_{\y'}\exp\biggr(-\dfrac{\|\x_{i}-\h(\y',\z';0)\|^{2}-2\eta(\y',z')}{2\sigma^{2}}\biggr)}. \nonumber
\end{eqnarray} 
At the zero-noise limit, i.e., $\sigma \to 0$, it is clear that $P(\y,\z|\x_{i}) = 1$ as $(\y,\z) = \mathop {\arg \min} \limits_{(\y',\z') \in \Jcal} \biggr\{\|\x_{i}-\h(\y',\z';0)\|^{2}-2\eta(\y',z')\biggr\}$ and $P(\y,\z|\x_{i}) = 0$ otherwise. Therefore, we can asymptotically view the complete log-likelihood of the DGM under the zero-noise limit as
\begin{eqnarray}
& & \hspace{- 6 em} \sum_{i=1}^{n}\sum_{(\y,\z) \in \Jcal} r_{\y,\z}\biggr(\log \pi_{\y,\z} -\frac{1}{2}\|\x_{i} - \h(\y,\z;0)\|^{2} \biggr) \nonumber \\
& &=  \underbrace{\sum_{i=1}^{n}\sum_{(\y,\z) \in \Jcal} -\frac{1}{2}r_{\y,\z}\|x_{n} - \h(\y,\z;0)\|^{2}}_{\textrm{Reconstruction Loss}} + \underbrace{\sum_{i=1}^{n}\sum_{(\y,\z) \in \Jcal} r_{\y,\z}\log \pi_{\y,\z}}_{\textrm{Path Normalization Regularizer}} \nonumber
\end{eqnarray}
where 
\begin{align}
r_{\y,\z} \equiv \left\{
	\begin{array}{ll}
		1,  & \mbox{if } (\y,\z) = \mathop {\arg \min} \limits_{(\y',\z') \in \Jcal} \biggr\{\|\x_{i}-\h(\y',\z';0)\|^{2}-2\eta(\y',z')\biggr\} \\
		0, & \mbox{otherwise} \nonumber
	\end{array}
\right.
\end{align}
With the above formulation, we have the following objective function 
\begin{eqnarray}
U_{n} =  \min \limits_{\theta} \dfrac{1}{n} \sum_{i=1}^{n}\sum_{(\y,\z) \in \Jcal} r_{\y,\z}\biggr(\frac{1}{2}\|\x_{i} - \h(\y,\z;0)\|^{2} - \log \pi_{\y,\z} \biggr) \label{eqn:nll-DGMm}
\end{eqnarray}
where $\theta=\biggr(\left\{\muy(\y)\right\}_{\y=1}^{K},\left\{\template(\ell)\right\}_{\ell=1}^{L},\left\{\pi_{\y}\right\}_{\y=1}^{K},\left\{\bias(\ell)\right\}_{\ell=1}^{L}\biggr)$. We call the above objective function to be \textit{unsupervised DGM without noise}.
\paragraph{Relaxation of unsupervised DGM without noise:} Unfortunately, the inference with unsupervised DGM without noise is intractable in practice due to two elements: the involvement of $\|\h(\y',\z';0)\|^{2}$ to determine the value of $r_{\y,\z}$ and the summation $\sum \limits_{(\y',\z') \in \Jcal} \exp(\eta(\y',z')+\log \pi_{\y'})$ in the denominator of $\pi_{\y,\z}$ for all $(\y,\z) \in \Jcal$. Therefore, we need to develop a tractable version of this objective function. 
\begin{theorem} \label{theorem:approximation_unsupervised_LD_DRM} (Relaxation of unsupervised DGM without noise)
Assume that $\pi_{\y} \geq \overline{\gamma}$ for all $1 \leq \y \leq K$ for some given $\overline{\gamma} \in (0,1/2)$. Denote
\begin{eqnarray}
V_{n} : = \min \limits_{\theta} \dfrac{1}{n} \sum \limits_{i=1}^{n} \sum \limits_{(\y',\z') \in \mathcal{J}}1_{\left\{(\y',\z') = \mathop{\arg \max}\limits_{(\y,\z) \in \mathcal{J}} \biggr(\h^{\top}(\y,\z;0)\x_{i}+\eta(\y,\z)\biggr) \right\}} \biggr( \dfrac{\| \x_{i} - \h(\y',\z';0)\|^{2}}{2} - \log(\pi_{\y',\z'})\biggr) \nonumber
\end{eqnarray} 
where $p_{\y',\z'} = \exp\biggr(\eta(\y',z')+\log \pi_{\y'}\biggr)/\biggr(\sum \limits_{\y=1}^{K}\exp\biggr(\max \limits_{\z \in \mathcal{L}} \eta(\y,\z) +\log \pi_{\y}\biggr)\biggr)$ for all $(\y',\z') \in \Jcal$. For any $\theta$, we define 
\begin{align}
(\overline{\y}_{i},\overline{\z}_{i}) = \mathop {\arg \min} \limits_{(\y,\z) \in \Jcal} \biggr\{\|\x_{i}-\h(\y,\z;0)\|^{2}-2\eta(\y,\z)\biggr\} \nonumber
\end{align}
and 
\begin{align}
(\widetilde{\y}_{i},\widetilde{\z}_{i}) = \mathop{\arg \max}\limits_{(\y,\z) \in \mathcal{J}} \biggr(\h^{\top}(\y,\z;0)\x_{i}+\eta(\y,\z)\biggr) \nonumber
\end{align}
as $1 \leq i \leq n$. Then, the following holds \\
(a) Upper bound:
\begin{eqnarray}
U_{n} & \leq & \min \limits_{\theta} \dfrac{1}{n} \sum \limits_{i=1}^{n}  \biggr\{\biggr( \dfrac{\| \x_{i} - \h(\widetilde{\y}_{i},\widetilde{\z}_{i}; 0)\|^{2}}{2} - \log(\pi_{\widetilde{\y}_{i},\widetilde{\z}_{i}})\biggr) + \underbrace{\biggr(\log \pi_{\widetilde{\y}_{i}} - \log \pi_{\overline{\y}_{i}}\biggr)}_{\text{prior loss}} \biggr\} + \log |\mathcal{L}| \nonumber \\
& \leq & V_{n} + \log \biggr(\dfrac{1}{\overline{\gamma}}-1\biggr) + \log |\mathcal{L}| \nonumber
\end{eqnarray}
(b) Lower bound: 
\begin{eqnarray}
\hspace { - 5 em} U_{n} & \geq & \min \limits_{\theta} \dfrac{1}{n} \sum \limits_{i=1}^{n}  \biggr\{\biggr( \dfrac{\| \x_{i} - \h(\widetilde{\y}_{i},\widetilde{\z}_{i}; 0)\|^{2}}{2} - \log(\pi_{\widetilde{\y}_{i},\widetilde{\z}_{i}})\biggr) + \underbrace{\biggr(\log \pi_{\overline{\y}_{i}} - \log \pi_{\widetilde{\y}_{i}} \biggr)}_{\text{prior loss}} \nonumber \\
& & \hspace{ 16 em} + \underbrace{\dfrac{1}{2}\biggr(\|\h(\overline{\y}_{i},\overline{\z}_{i}; 0)\|^{2} - \|\h(\widetilde{\y}_{i},\widetilde{\z}_{i}; 0)\|^{2}\biggr)}_{\text{norm loss}} \biggr\} \nonumber \\
& & \geq V_{n} + \log \biggr(\dfrac{\overline{\gamma}}{1-\overline{\gamma}}\biggr) + \min \limits_{\theta} \dfrac{1}{n} \sum \limits_{i=1}^{n} \dfrac{1}{2}\biggr(\|\h(\overline{\y}_{i},\overline{\z}_{i}; 0)\|^{2} - \|\h(\widetilde{\y}_{i},\widetilde{\z}_{i}; 0)\|^{2}\biggr) \nonumber
\end{eqnarray}
\end{theorem}
Unlike $U_{n}$, the inference with objective function of $V_{n}$ is tractable. According to the upper bound and lower bound of $U_{n}$ in terms of $V_{n}$, we can use $V_{n}$ as a tractable approximation of $U_{n}$ for the inference purpose with unsupervised setting of data when the noise is treated to be 0. Therefore, we achieve the conclusion of part (b) of Theorem \ref{theorem:learning_lddrm} in the main text. The algorithm for determined (local) minima of $V_{n}$ is summarized in Algorithm \ref{alg:relax_unsup_LD_DRM}.
\subsection{Relaxation of non-negativity assumption with rendered images} \label{Section:relaxation_inference_non_negativity}
It is clear that the inference with $V_{n}$ relies on the non-negativity assumption such that equation \eqref{eqn:CNN_structure} holds.  Now, we will argue that when the non-negativity assumption with rendered images $\h(\y,\z;0)$ does not hold, we can relax $V_{n}$ to a more tractable version under that setting.
\begin{theorem} \label{theorem:approximation_unsupervised_LD_DRM_no_constraint} (Relaxation of objective function $V_{n}$ when non-negativity assumption does not hold) 
Assume that $\pi_{\y} \geq \overline{\gamma}$ for all $1 \leq \y \leq K$ for some given $\overline{\gamma} \in (0,1/2)$. Denote 
\begin{eqnarray}
W_{n} : = \min \limits_{\theta} \dfrac{1}{n} \sum \limits_{i=1}^{n} \sum \limits_{\y'=1}^{K} 1_{\left\{\y' = \mathop{\arg \max}\limits_{\y \in \mathcal{J}} g(\y,\x_{i}) \right\}} \biggr( \dfrac{\| \x_{i} - g(\y',\overline{\z}_{i})\|^{2}}{2} - \log(\pi_{\y',\overline{\z}_{i}})\biggr) \nonumber
\end{eqnarray}
where
\begin{eqnarray}
& & \hspace{ - 3 em} g(\y,\x) = \h^{\top}(\y) \maxpool\biggr(\relu\left(\convo\biggr(\template(\ell),\cdots \maxpool\biggr(\relu\biggr(\convo\biggr(\template(1), I \right)+\bias(1)\biggr)\biggr)\biggr) \nonumber \\
& & \hspace{ 30 em}\cdots +\bias(\ell) \biggr)\biggr) \nonumber
\end{eqnarray}
for all $(\x,\y)$. Additionally, $\overline{\z}_{i}$ is the maximal value of $\z$ in the CNN structure of $g(\y,\x_{i})$ for $1 \leq i \leq n$. For any $\theta$, we define 
\begin{align}
(\widetilde{\y}_{i},\widetilde{\z}_{i}) = \mathop{\arg \max}\limits_{(\y,\z) \in \mathcal{J}} \biggr(\h^{\top}(\y,\z;0)\x_{i}+\eta(\y,\z)\biggr) \nonumber
\end{align}
and $\overline{\y}_{i} = \mathop{\arg \max} \limits_{\y} g(\y,\x_{i})$ as $1 \leq i \leq n$. Then, the following holds \\
(a) Upper bound:
\begin{eqnarray}
V_{n} & \leq & \min \limits_{\theta} \dfrac{1}{n} \sum \limits_{i=1}^{n}  \biggr\{\biggr( \dfrac{\| \x_{i} - \h(\overline{\y}_{i},\overline{\z}_{i}; 0)\|^{2}}{2} - \log(\pi_{\overline{\y}_{i},\overline{\z}_{i}})\biggr) + \underbrace{\biggr(\log \pi_{\overline{\y}_{i}} - \log \pi_{\widetilde{\y}_{i}} \biggr)}_{\text{prior loss}} \nonumber \\
& & \hspace{ 16 em} +  \underbrace{\dfrac{1}{2}\biggr(\|\h(\widetilde{\y}_{i},\widetilde{\z}_{i}; 0)\|^{2} - \|\h(\overline{\y}_{i},\overline{\z}_{i}; 0)\|^{2}\biggr)}_{\text{norm loss}} \biggr\} \nonumber \\
& \leq & W_{n} + \log \biggr(\dfrac{1}{\overline{\gamma}}-1\biggr) + \min \limits_{\theta} \dfrac{1}{n} \sum \limits_{i=1}^{n}\dfrac{1}{2}\biggr(\|\h(\widetilde{\y}_{i},\widetilde{\z}_{i}; 0)\|^{2} - \|\h(\overline{\y}_{i},\overline{\z}_{i}; 0)\|^{2}\biggr) \nonumber
\end{eqnarray}
(b) Lower bound:
\begin{eqnarray}
V_{n} & \geq &  \min \limits_{\theta} \dfrac{1}{n} \sum \limits_{i=1}^{n}  \biggr\{\biggr( \dfrac{\| \x_{i} - \h(\overline{\y}_{i},\overline{\z}_{i}; 0)\|^{2}}{2} - \log(\pi_{\overline{\y}_{i},\overline{\z}_{i}})\biggr)  + \underbrace{\biggr(\log \pi_{\overline{\y}_{i}} - \log \pi_{\widetilde{\y}_{i}} \biggr)}_{\text{prior loss}} \nonumber \\
& & +  \underbrace{\dfrac{1}{2}\biggr(\|\h(\widetilde{\y}_{i},\widetilde{\z}_{i}; 0)\|^{2} - \|\h(\overline{\y}_{i},\overline{\z}_{i}; 0)\|^{2}\biggr)}_{\text{norm loss}} + \underbrace{\biggr(g(\overline{\y}_{i},\x_{i}) - \left\{\h(\widetilde{\y}_{i},\widetilde{\z}_{i}; 0)^{\top}\x_{i}+\eta(\widetilde{\y}_{i},\widetilde{\z}_{i})\right\} \biggr)}_{\text{CNN loss}} \biggr\} \nonumber \\
& & \geq W_{n} + \log \biggr(\dfrac{\overline{\gamma}}{1-\overline{\gamma}}\biggr) + \min \limits_{\theta} \dfrac{1}{n} \sum \limits_{i=1}^{n} \dfrac{1}{2}\biggr(\|\h(\widetilde{\y}_{i},\widetilde{\z}_{i}; 0)\|^{2} - \|\h(\overline{\y}_{i},\overline{\z}_{i}; 0)\|^{2}\biggr) \nonumber \\
& & + \min \limits_{\theta} \dfrac{1}{n} \sum \limits_{i=1}^{n} \biggr(g(\overline{\y}_{i},\x_{i}) - \left\{\h(\widetilde{\y}_{i},\widetilde{\z}_{i}; 0)^{\top}\x_{i}+\eta(\widetilde{\y}_{i},\widetilde{\z}_{i})\right\} \biggr) \nonumber
\end{eqnarray}
\end{theorem}
The proof argument of the above theorem is similar to that of Theorem \ref{theorem:approximation_unsupervised_LD_DRM}; therefore, it is omitted. The upper bound and lower bound of $V_{n}$ in terms of $W_{n}$ in Theorem \ref{theorem:approximation_unsupervised_LD_DRM_no_constraint} implies that we can use $W_{n}$ as a relaxation of $V_{n}$ when the non-negativity assumption with rendered images $\h(\y,\z;0)$ does not hold. The algorithm for achieving the (local) minima of $W_{n}$ is similar to Algorithm \ref{alg:relax_unsup_LD_DRM}. 
\begin{algorithm}[tb]
   \caption{Relaxation of unsupervised DGM without noise}
   \label{alg:relax_unsup_LD_DRM}
\begin{algorithmic}
   \STATE {\bfseries Input:} Data $\x_{i}$, translation matrices $\ttrans(\trans;\ell)$, zero padding matrices $\pad(\ell)$, number of labels $K$, number of layers $L$. 
   \STATE {\bfseries Output:} Parameters $\theta$.
   \STATE Initialize $\theta=\biggr(\left\{\muy(\y)\right\}_{\y=1}^{K},\left\{\template(\ell)\right\}_{\ell=1}^{L},\left\{\pi_{\y}\right\}_{\y=1}^{K},\left\{\bias(\ell)\right\}_{\ell=1}^{L}\biggr)$.
   \WHILE{$\theta$ has not converged}
   \STATE 1. E-Step: Update labels $(\y,\z)$ of each data
   \FOR {$i=1$ {\bfseries to} $n$}
   \STATE $(\widehat{\y}_{i},\widehat{\z}_{i}) = \mathop {\arg \max} \limits_{\y,\z} \left(\h^{\top}(\y, \z)\x_{i} + \log(\pi_{\y,\z})\right)$.\\
	\ENDFOR
	\STATE 2. M-Step: By using Stochastic Gradient Descent (SGD), update $\theta$ that minimizes $\sum \limits_{i=1}^{n} \biggr( \dfrac{\| \x_{i} - \h(\widehat{\y}_{i},\widehat{\z}_{i})\|^{2}}{2} - \log(\pi_{\widehat{\y}_{i},\widehat{\z}_{i}})\biggr)$.
   \ENDWHILE
\end{algorithmic}
\end{algorithm}
\subsection{DGM with (semi)-supervised setting} \label{sec:inference-NRM_semi_sup}
In this section, we consider the application of DGM to the (semi)-supervised setting of the data. Under that setting, only a (full) portion of labels of data $\x_{1},\ldots,\x_{n}$ is available. Without loss of generality, we also assume that the rendering path $\h(\y,\z;0)$ satisfies the non-negativity assumption. For the case that $\h(\y,\z;0)$ does not satisfy this assumption, we can argue in the same fashion as that of Theorem \ref{theorem:approximation_unsupervised_LD_DRM_no_constraint}. Now, we assume that only the labels $(\y_{n_{1}+1},\ldots,\y_{n})$ are unknown for some $n_{1} \geq 0$. When $n_{1}=0$, we have the supervised setting of data while we have the semi-supervised setting of data when $n-n_{1}$ is small. Our goal is to build a semi-supervised model based on DGM such that the clustering information from data $\x_{1},\ldots,\x_{n_{1}}$ can be used efficiently to increase the accuracy of classifying the labels of data $\x_{n_{1}+1},\ldots,\x_{n}$. For the sake of simple inference with that purpose, we only consider the setting of DGM when the noise goes to 0. Our idea of constructing the semi-supervised model based on DGM is insprired by an approximation of the upper bound of maximizing the conditional log likelihood of DGM in terms of the cross entropy and reconstruction loss in part (b) of Theorem \ref{theorem:learning_lddrm}. In particular, we combine the tractable version of reconstruction loss from the unsupervised setting in Theorem \ref{theorem:learning_lddrm}b and the cross entropy of approximate posterior in Theorem \ref{theorem:learning_lddrm}a, which can be formulated as follows
\begin{eqnarray}
& & \hspace{- 3 em} \min \limits_{\theta} \dfrac{\alpha_{\text{RC}}}{n} \biggr\{\sum \limits_{i=1}^{n_{1}} \sum \limits_{(\y',\z') \in \mathcal{J}}1_{\left\{(\y',\z') = \mathop{\arg \max}\limits_{(\y,\z) \in \mathcal{J}} \biggr(\h^{\top}(\y,\z;0)\x_{i}+\tau(\theta_{\y,\z})\biggr) \right\}} \biggr( \dfrac{\| \x_{i} - \h(\y',\z';0)\|^{2}}{2} - \log(\pi_{\y',\z'})\biggr) \nonumber \\
& & + \biggr(\sum \limits_{i=n_{1}+1}^{n} \sum \limits_{\z' \in \mathcal{L}}1_{\left\{\z' = \mathop{\arg \max}\limits_{\z \in \mathcal{L}} \biggr(\h^{\top}(\y_{i},\z;0)\x_{i}+\tau(\theta_{\y_{i},\z})\biggr) \right\}} \biggr( \dfrac{\| \x_{i} - \h(\y_{i},\z';0)\|^{2}}{2} - \log(\pi_{\y_{i},\z'})\biggr)\biggr\} \nonumber \\
& & - \dfrac{\alpha_{\text{CE}}}{n-n_{1}} \sum \limits_{i=n_{1}+1}^{n} \log q_{\theta}(\y_{i}|\x_{i}) \nonumber
\end{eqnarray}
where $\alpha_{\text{RC}}$ and $\alpha_{\text{CE}}$ are non-negative weights associated with reconstruction loss and cross entropy respectively. Additionally, the approximate posterior $q_{\theta}(\y|\x_{i})$ is chosen as 
\begin{eqnarray} 
q_{\theta}(\y|\x_{i}) = \dfrac{\exp\biggr(\max \limits_{\z \in \mathcal{L}} \biggr\{\h^{\top}(\y,\z;0)\x_{i}+ \tau(\theta_{\y,\z})\biggr\}+ \log \pi_{\y} \biggr)}{\sum \limits_{\y'=1}^{K} \exp\biggr(\max \limits_{\z \in \mathcal{L}} \biggr\{\h^{\top}(\y',\z; 0)\x_{i}+ \tau(\theta_{\y',\z}) \biggr\}+ \log \pi_{\y'} \biggr)}. \nonumber
\end{eqnarray} Note that, since the labels $(\y_{n_{1}+1},\ldots,\y_{n})$ are known, the reconstruction loss for clustering data $\x_{n_{1}+1},\ldots,\x_{n}$ in the above objective function indeeds incorporate these information to improve the accuracy of estimating the parameters. We call the above objective function to be \textit{(semi)-supervised DGM without noise}.
\paragraph{Boosting the accuracy of (semi)-supervised DGM without noise:} In practice, it may happen that the accuracy of classifying data by using the parameters from (semi)-supervised DGM without noise is not very high. To account for that problem, we consider the following general version of (semi)-supervised DGM without noise that includes the variational inference term and the moment matching term
\begin{eqnarray}
& & \hspace{ - 3 em} \min \limits_{\theta} \dfrac{\alpha_{\text{RC}}}{n} \biggr\{\sum \limits_{i=1}^{n_{1}} \sum \limits_{(\y',\z') \in \mathcal{J}}1_{\left\{(\y',\z') = \mathop{\arg \max}\limits_{(\y,\z) \in \mathcal{J}} \biggr(\h^{\top}(\y,\z;0)\x_{i}+\tau(\theta_{\y,\z})\biggr) \right\}} \biggr( \dfrac{\| \x_{i} - \h(\y',\z';0)\|^{2}}{2} - \log(\pi_{\y',\z'})\biggr) \nonumber \\
& & + \biggr(\sum \limits_{i=n_{1}+1}^{n} \sum \limits_{\z' \in \mathcal{L}}1_{\left\{\z' = \mathop{\arg \max}\limits_{\z \in \mathcal{L}} \biggr(\h^{\top}(\y_{i},\z;0)\x_{i}+\tau(\theta_{\y_{i},\z})\biggr) \right\}} \biggr( \dfrac{\| \x_{i} - \h(\y_{i},\z';0)\|^{2}}{2} - \log(\pi_{\y_{i},\z'})\biggr)\biggr\} \nonumber \\
& & - \dfrac{\alpha_{\text{CE}}}{n-n_{1}} \sum \limits_{i=n_{1}+1}^{n} \log q_{\theta}(\y_{i}|\x_{i})  + \dfrac{\alpha_{\text{KL}}}{n} \sum \limits_{i=1}^{n} \sum \limits_{\y=1}^{K} q_{\theta}(\y|\x_{i}) \log\biggr(\dfrac{q_{\theta}(\y|\x_{i})}{\pi_{\y}}\biggr) \nonumber \\
& & + \alpha_{\text{MM}} \sum \limits_{\ell=1}^{L} \text{D}_{\text{KL}} \left(\N(\muy_{h(\ell)}, \sigma^{2}_{h(\ell)})||\N(\muy_{\acts(\ell)}, \sigma^{2}_{\acts(\ell)})\right).\label{eqn:full_objective_semi_sup_SPLD_K_means}
\end{eqnarray}
Here, $\alpha_{\text{KL}}$ and $\alpha_{MM}$ are non-negative weights associated with the variational inference loss and moment matching loss respectively. Additionally, $\muy_{h(\ell)}, \sigma^{2}_{h(\ell)}, \muy_{\acts(\ell)}, \sigma^{2}_{\acts(\ell)}$ in the moment matching loss are defined as follows:
\begin{align}
\label{eqn:moment_matching}
    \muy_{h(\ell)} & = \frac{1}{n}\sum \limits_{i=1}^{n}\hat{\h}(\ell)_{i}, \,\, &\sigma^{2}_{\h(\ell)} &= \frac{1}{n}\sum \limits_{i=1}^{n}(\hat{\h}(\ell)_{i} - \muy_{h(\ell)})^{2} \nonumber \\
    \muy_{\acts(\ell)} &= \frac{1}{n}\sum \limits_{i=1}^{n}\acts(\ell)_{i}, \, \, &\sigma^{2}_{\acts(\ell)} &= \frac{1}{n}\sum \limits_{i=1}^{n}(\acts(\ell)_{i} - \muy_{\acts(\ell)})^{2} 
\end{align}
where $\hat{\h}(\ell)_{i}$ is the estimated value of $\h(\ell)$ given the optimal latent variables $\hat{\sparse}^{(\ell)}$ and $\hat{\trans}^{(\ell)}$ inferred from the image $\x_{i}$ for $1 \leq i \leq n$. It is clear that when $\alpha_{\text{KL}}=\alpha_{MM} = 0$, we return to (semi)-supervised DGM without noise. In Appendix C, we provide careful theoretical analyses regarding statistical guarantees of model \eqref{eqn:full_objective_semi_sup_SPLD_K_means}. 

Now, we will provide heuristic explanations about the improvement in terms of performance of model \eqref{eqn:full_objective_semi_sup_SPLD_K_means} based on the variational inference term and the moment matching term. 

{\bf Regarding the variational term:} The DGM inference algorithm developed thus far ignores uncertainty in the latent nuisance posterior $p(\y, \z | \x)$ due to the max-marginalization over $(\y, \z)$ in the E-step bottom-up inference. We would like to properly account for this uncertainty for two main reasons: (i) our fundamental hypothesis is that the brain performs probabilistic inference and (ii) uncertainty accounting is very important for good generalization in the semi-supervised setting since we have very little labeled data.

One approach attempts to approximate the true class posterior $p(\y|\x)$ for the DGM. We employ \textit{variational inference}, a technique that enables the approximate inference of the latent posterior. Mathematically, for the DGM this means we would like to approximate the true class posterior $p(\y|\x) \approx q(\y|\x)$, where the approximate posterior $q$ is restricted to some tractable family of distributions (e.g. Gaussian or Categorical). We strategically choose the tractable family to be $q(\y|\x) \equiv p(\y | \hat{\z}, \x)$, where $\hat{\z} \equiv \underset{\z}{\amax}\, p(\y,\z|\x)$. In other words, we choose $q$ to be restricted to the DGM family of nuisance max-marginalized class posteriors. Note that this is indeed an approximation, since the true DGM class posterior has nuisances that are \textit{sum}-marginalized out $p(\y | \x) = \underset{\z}{\sum}\, p(\y,\z|\x)$, whereas the approximating variational family has nuisances that are \textit{max}-marginalized out.

Given our choice of variational family $q$, we derive the variational term for the loss function, starting from the principled goal of minimizing the KL-distance $D_{KL}[q(\y|\x) || p(\y|\x)]$ between the true and approximate posteriors with respect to the parameters of $q$. As a result, such an optimized $q$ will tilt towards better approximating $p(\y|\x)$, which in turn means that it will account for \textit{some} of the uncertainty in $p(\z|\x)$. The variational terms in the loss are defined as \cite{blei2016variational}: 
\begin{align}
    \mathcal{L}_{VI} &\equiv \mathcal{L}_{RC} + \beta_{KL} \mathcal{L}_{KL} \nonumber \\
    &\equiv -\Expect_q \left[ \ln p(\x|\y) \right] + \beta_{KL} D_{KL}[ q(\y|\x) || p(\y) ].
    \label{eqn:var-inf}
\end{align}
This term is quite similar to that used in variational autoencoders (VAE) \cite{kingma2013auto}, except for two key differences: (i) here the latent variable $\y$ is discrete categorical rather than continuous Gaussian and (ii) we have employed a slight relaxation of the VAE by allowing for a penalty parameter $\beta_{KL} \neq 1$. The latter is motivated by recent experimental results showing that such freedom enables optimal disentangling of the true intrinsic latent variables from the data.

{\bf Regarding the moment matching term:} Batch Normalization can potentially be derived by normalizing the intermediate rendered images $\h(\ell)$, $\ell=1,2,\dots,L$ in the DGM by subtracting their means and dividing by their standard deviations under the assumption that the means and standard derivations of $\h(\ell)$ are close to those of the activation $\acts(\ell)$ in the CNNs. From this intuition, in Section~\ref{sec:inference-NRM_semi_sup} of Appendix A, we introduce the moment-matching loss to improve the performance of the DGM/CNNs trained for semi-supervised learning tasks.
\subsection{Statistical guarantees for (semi)-supervised setting}
For the sake of simplicity with proof argument, we only provide detail theoretical analysis for statistical guarantee with the setting of \eqref{eqn:full_objective_semi_sup_SPLD_K_means} when the moment matching term is skipped. In particular, we are interested in the following (semi)-supervised model
\begin{eqnarray}
& & \hspace{ - 2 em} Y_{n}  : = \min \limits_{\theta} \dfrac{\alpha_{\text{RC}}}{n} \biggr\{\sum \limits_{i=1}^{n_{1}} \sum \limits_{(\y',\z') \in \mathcal{J}}1_{\left\{(\y',\z') = \mathop{\arg \max}\limits_{(\y,\z) \in \mathcal{J}} \biggr(\h^{\top}(\y,\z;0)\x_{i}+\tau(\theta_{\y,\z})\biggr) \right\}} \nonumber \\
& & \hspace{18 em} \times \biggr( \dfrac{\| \x_{i} - \h(\y',\z';0)\|^{2}}{2} - \log(\pi_{\y',\z'})\biggr) \nonumber \\
& & + \biggr(\sum \limits_{i=n_{1}+1}^{n} \sum \limits_{\z' \in \mathcal{L}}1_{\left\{\z' = \mathop{\arg \max}\limits_{\z \in \mathcal{L}} \biggr(\h^{\top}(\y_{i},\z;0)\x_{i}+\tau(\theta_{\y_{i},\z})\biggr) \right\}} \biggr( \dfrac{\| \x_{i} - \h(\y_{i},\z';0)\|^{2}}{2} - \log(\pi_{\y_{i},\z'})\biggr)\biggr\} \nonumber \\
& & - \dfrac{\alpha_{\text{CE}}}{n-n_{1}} \sum \limits_{i=n_{1}+1}^{n} \log q_{\theta}(\y_{i}|\x_{i})  + \dfrac{\alpha_{\text{KL}}}{n} \sum \limits_{i=1}^{n} \sum \limits_{\y=1}^{K} q_{\theta}(\y|\x_{i}) \log\biggr(\dfrac{q_{\theta}(\y|\x_{i})}{\pi_{\y}}\biggr) \label{eqn:objective_semi_sup_SPLD_K_means}
\end{eqnarray}
where the approximate posterior $q_{\theta}(\y|\x_{i})$ is chosen as 
\begin{eqnarray} 
q_{\theta}(\y|\x_{i}) : = \dfrac{\exp\biggr(\max \limits_{\z \in \mathcal{L}} \biggr\{\h^{\top}(\y,\z;0)\x_{i}+ \tau(\theta_{\y,\z})\biggr\}+ \log \pi_{\y} \biggr)}{\sum \limits_{y'=1}^{K} \exp\biggr(\max \limits_{\z \in \mathcal{L}} \biggr\{\h^{\top}(\y',\z; 0)\x_{i}+ \tau(\theta_{\y',\z}) \biggr\}+ \log \pi_{\y'} \biggr)}. \nonumber
\end{eqnarray} 
Here, $\alpha_{\text{RC}}$, $\alpha_{\text{CE}}$, and $\alpha_{\text{KL}}$ are non-negative weights associated with reconstruction loss, cross entropy, and variational inference respectively. As being indicated in the formulation of objection function $Y_{n}$, the only difference between $Y_{n}$ and~\eqref{eqn:full_objective_semi_sup_SPLD_K_means} is the weight $\alpha_{\text{MM}}$ regarding moment matching loss in~\eqref{eqn:full_objective_semi_sup_SPLD_K_means} is set to be 0. To ease the presentation with theoretical analyses later, we call the objective function with $Y_{n}$ to be \textit{partially labeled latent dependence regularized cross entropy (partially labeled LDCE)}. 
\paragraph{Consistency of partially labeled LDCE:}
Firstly, we demonstrate that the objective function of partially labeled LDCE enjoys the consistency guarantee. 
\begin{theorem} \label{theorem:consistency_objective_LDAP_cross_entropy} (Consistency of objective function of partially labeled LDCE) 
Assume that $n_{1}$ is a function of $n$ such that $n_{1}/n \to \overline{\lambda}$ as $n \to \infty$. Furthermore, $\mathbb{P}(\|\x\| \leq R) = 1$ as $\x \sim P$ for some given $R>0$. We denote the population version of partially labeled LDCE as follows
\begin{eqnarray}
& & \hspace{ - 2 em} \overline{Y}  : = \min \limits_{\theta} \alpha_{\text{RC}}\biggr\{\overline{\lambda}\biggr(\int \sum \limits_{(\y',\z') \in \mathcal{J}}1_{\left\{(\y',\z') = \mathop{\arg \max}\limits_{(\y,\z) \in \mathcal{J}} \biggr(\h^{\top}(\y,\z;0)x+\tau(\theta_{\y,\z})\biggr)\right\}} \biggr( \dfrac{\| x - \h(\y',\z';0)\|^{2}}{2} \nonumber \\
& & \hspace{ - 2 em} - \log(\pi_{\y',\z'})\biggr)dP(x)\biggr) + (1-\overline{\lambda}) \biggr(\int \sum \limits_{\z' \in \mathcal{L}}1_{\left\{\z' = \mathop{\arg \max}\limits_{\z \in \mathcal{L}} \biggr(\h^{\top}(\y,\z;0)x+\tau(\theta_{\y,\z})\biggr)\right\}} \biggr( \dfrac{\| x - \h(\y,\z';0)\|^{2}}{2} \nonumber \\
& & \hspace{ - 2 em} - \log(\pi_{\y,\z'})\biggr)dQ(x,c)\biggr)\biggr\} - \alpha_{\text{CE}} \int \log q_{\theta}(\y|x)dQ(x,c) + \alpha_{\text{KL}} \int \sum \limits_{\y=1}^{K} q_{\theta}(\y|x)\log \biggr(\dfrac{q_{\theta}(\y|x)}{\pi_{\y}}\biggr)dP(x). \nonumber
\end{eqnarray}
Then, we obtain that $Y_{n} \to \overline{Y}$ almost surely as $n \to \infty$.
\end{theorem}
The detail proof of Theorem~\ref{theorem:consistency_objective_LDAP_cross_entropy} is deferred to Appendix C. Now, we denote 
\begin{align*}
\widetilde{\theta} := \biggr(\left\{\widetilde{\muy}(y) \right\}_{\y=1}^{K},\left\{\widetilde{\template}(\ell)\right\}_{\ell=1}^{L},\left\{\widetilde{\pi}_{y}\right\}_{\y=1}^{K},\left\{\widetilde{\bias}(\ell) \right\}_{\ell=1}^{L} \biggr)
\end{align*} 
the optimal solutions of objective function \eqref{eqn:objective_semi_sup_SPLD_K_means}. Note that, the existence of these optimal solutions is guaranteed due to the compactness assumption of the parameter spaces $\Theta_{\ell}$, $\Omega$, and $\Xi_{l}$ for $1 \leq  \ell \leq L$. The optimal solutions $\left\{\widetilde{\muy}(y) \right\}_{\y=1}^{K}$ and $\left\{\widetilde{\template}(\ell)\right\}_{\ell=1}^{L}$ lead to corresponding set of optimal rendered images $\widetilde{S}_{n}$. Similar to the case of SPLD regularized K-means, our goal is to guarantee the consistency of $\widetilde{S}_{n}$ as well as $\left\{\widetilde{\pi}_{y}\right\}_{\y=1}^{K},\left\{\widetilde{\bias}(\ell) \right\}_{\ell=1}^{L}$. 

In particular, we denote $\widetilde{\mathcal{F}}_{0}$ the set of all optimal solutions $\widetilde{\theta}^{0}$ of population partially labeled LDCE where $\widetilde{\theta}^{0} : = \biggr(\left\{\widetilde{\muy}^{0}(y) \right\}_{\y=1}^{K},\left\{\widetilde{\template}_{0}(\ell)\right\}_{\ell=1}^{L},\left\{\widetilde{\pi}_{y}^{0}\right\}_{\y=1}^{K},\left\{\widetilde{\bias}_{0}(\ell) \right\}_{\ell=1}^{L} \biggr)$. For each $\widetilde{\theta}^{0}\in \widetilde{\mathcal{F}}_{0}$, we define $\widetilde{S}_{0}$ the set of optimal rendered images associated with $\widetilde{\theta}^{0}$. We denote $\Gcal(\widetilde{\mathcal{F}}_{0})$ the corresponding set of all optimal rendered images $\widetilde S_{0}$, optimal prior probabilities $\left\{\widetilde{\pi}_{\y}^{0}\right\}_{\y=1}^{K}$, and optimal biases $\left\{\widetilde{\bias}_{0}(\ell) \right\}_{\ell=1}^{L}$. 
\begin{theorem} \label{theorem:consistency_optimal_solutions_LDCE} (Consistency of optimal rendering paths and optimal solutions of partially labeled LDCE) 
Assume that $\mathbb{P}(\|\x\| \leq R) = 1$ as $\x \sim P$ for some given $R>0$. Then, we obtain that 
\begin{eqnarray}
\inf \limits_{\left(\widetilde{S}_{0}, \left\{\widetilde{\pi}_{y}^{0}\right\},\left\{\widetilde{\bias}_{0}(\ell) \right\} \right) \in \Gcal(\widetilde{\mathcal{F}}_{0})} \biggr\{H(\widetilde{S}_{n},\widetilde{S}_{0}) + \sum \limits_{\y=1}^{K} |\widetilde{\pi}_{y} - \widetilde{\pi}_{y}^{0}| + \sum \limits_{\ell=1}^{L} \|\widetilde{\bias}(\ell) - \widetilde{\bias}_{0}(\ell)\|\biggr\}  \to 0 \nonumber
\end{eqnarray}
almost surely as $n \to \infty$. 
\end{theorem}
The detail proof of Theorem~\ref{theorem:consistency_optimal_solutions_LDCE} is postponed to Appendix C. 
\subsection{Generalization bound for classification framework with (semi)-supervised setting} \label{Section:semi_supervised_LD_DRMM}
In this section, we provide a simple generalization bound for certain classification function with the optimal solutions $\widetilde{\theta} = \biggr(\left\{\widetilde{\muy}(y) \right\}_{\y=1}^{K},\left\{\widetilde{\template}(\ell)\right\}_{\ell=1}^{L},\left\{\widetilde{\pi}_{y}\right\}_{\y=1}^{K},\left\{\widetilde{\bias}(\ell) \right\}_{\ell=1}^{L} \biggr)$ of~\eqref{eqn:full_objective_semi_sup_SPLD_K_means}. In particular, we denote the following function $f:\mathbb{R}^{D^{(0)}} \times \left\{1,\ldots,K\right\} \to \mathbb{R}$ as 
\begin{eqnarray}
f(\x,\y) = \max \limits_{\z \in \mathcal{L}} \biggr\{\widetilde \h^{\top}(\y,\z;0)x+\tau(\widetilde \theta_{\y,\z}) \biggr\} + \log \widetilde{\pi}_{y} \nonumber
\end{eqnarray} 
for all $(\x,\y) \in \mathbb{R}^{D^{(0)}} \times \left\{1,\ldots,K\right\}$ where 
\begin{align}
\widetilde \h(\y,\z; 0) & : = \widetilde \Lambda(\z;1)\ldots \widetilde \Lambda(\z;L)\widetilde \muy(\y), \nonumber \\
\widetilde \Lambda(\z;\ell) & : = \sum_{\p\in \pp(\ell)}  \sparse(\ell, \p)\ttrans(t;\ell,\p) \pad(\ell, \p)\widetilde \template(\ell, \p) \nonumber
\end{align}
for all $(\y,\z)$ and $1 \leq \ell \leq L$. To achieve the generalization bound regarding that classification function, we rely on the study of generalization bound with margin loss. For the simplicity of argument, we assume that the true labels of $\x_{1},\ldots, \x_{n}$ are $\y_{1},\ldots,\y_{n}$ while $\y_{1},\ldots,\y_{n_{1}}$ are not available to train. The margin of a labeled example $(\x,\y)$ based on $f$ can be defined as
\begin{eqnarray}
\rho(f,\x,\y) = f(\x,\y) - \max \limits_{l \neq \y} f(\x,l). \nonumber
\end{eqnarray}
Therefore, the classification function $f$ misspecifies the labeled example $(\x,\y)$ as long as $\rho(f,\x,\y) \leq 0$. The empirical margin error of $f$ at margin coefficient $\template \geq 0$ is 
\begin{eqnarray}
R_{n,\gamma}(f) = \dfrac{1}{n} \sum \limits_{i=1}^{n} 1_{\left\{\rho(f,\x_{i},\y_{i}) \leq \gamma \right\}}. \nonumber
\end{eqnarray}
It is clear that $R_{n,0}(f)$ is the empirical risk of 0-1 loss, i.e., we have
\begin{eqnarray}
R_{n,0}(f) = \dfrac{1}{n} \sum \limits_{i=1}^{n} 1_{\left\{\mathop {\arg \max} \limits_{1 \leq \y \leq K} f(\x_{i},\y) \neq \y_{i}\right\}}. \nonumber
\end{eqnarray}
Similar to the argument in the case of SPLD regularized K-means, the optimal solutions $\widetilde{\theta}$ of partially labeled LDCE lead to a set of rendered images $\widetilde{\h}(\y,\z; 0)$ for all $(\y,\z) \in \Jcal$. However, only a small fraction of rendering paths are indeed active in the following sense. 
There exists a subset $\mathcal{L}_{n}$ of $\mathcal{L}$ such that $|\mathcal{L}_{n}| \leq \overline{\tau}_{n}|\mathcal{L}|$ where $\overline{\tau}_{n} \in (0,1]$, which is independent of data $(\x_{1},\y_{1}),\ldots,(\x_{n},\y_{n})$, and the following holds
\begin{eqnarray}
\max \limits_{\z \in \mathcal{L}} \biggr\{\widetilde \h^{\top}(\y,\z;0)x+\tau(\widetilde \theta_{\y,\z}) \biggr\} + \log \widetilde{\pi}_{y} = \max \limits_{\z \in \mathcal{L}_{n}} \biggr\{\widetilde \h^{\top}(\y,\z;0)x + \tau(\widetilde \theta_{\y,\z}) \biggr\} + \log \widetilde{\pi}_{y} \nonumber
\end{eqnarray}
for all $1 \leq \y \leq K$. The above equation implies that
\begin{eqnarray}
R_{n,\template}(f) = R_{n,\gamma}(f_{\overline{\gamma}_{n}}) \nonumber
\end{eqnarray}
for all $\template \geq 0$ and $n \geq 1$ where $f_{\overline{\tau}_{n}}(\x,\y) = \max \limits_{\z \in \mathcal{L}_{n}} \biggr\{\widetilde \h^{\top}(\y,\z;0)x + \tau(\widetilde \theta_{\y,\z}) \biggr\} + \log \widetilde{\pi}_{y}$ for all $(\x,\y)$. With that connection, we denote the expected margin error of classification function $f_{\overline{\tau}_{n}}$ at margin coefficient $\template \geq 0$ is 
\begin{eqnarray}
R_{\gamma}(f_{\overline{\tau}_{n}})  =  \mathbb{E} 1_{\left\{\rho(f_{\overline{\tau}_{n}},\x,\y) \leq \gamma \right\}}. \nonumber
\end{eqnarray}
The generalization bound that we establish in this section will base on the gap between the expected margin error $R_{0}(f_{\overline{\tau}_{n}})$ and its corresponding empirical version $R_{n,\template}(f_{\overline{\tau}_{n}})$, which is also $R_{n,\template}(f)$. 
\begin{theorem} \label{theorem:generalization_gap_classification_LDCE} (Generalization bound for margin-based classification)
Assume that $P(\|\x\| \leq R)=1$ for some given $R>0$ and $\x \sim P$. Additionally, the parameter spaces $\Theta_{\ell}$ and $\Omega$ are chosen such that $\|\h(\y,\z; 0)\| \leq R$ for all $(\y,\z) \in \Jcal$. For any $\delta>0$, with probability at least $1-\delta$, we have
\begin{eqnarray}
& & \hspace{-2 em} R_{0}(f_{\overline{\tau}_{n}})  \leq  \inf \limits_{\gamma \in (0,1]} \biggr\{R_{n,\gamma}(f_{\overline{\tau}_{n}})+\dfrac{8K(2K-1)}{\template\sqrt{n}}\biggr(2\overline{\tau}_{n}|\mathcal{L}|(R^{2}+1)+|\log \overline{\gamma}| \biggr) \nonumber \\
& & \hspace{ 2em} +\biggr(\dfrac{\log\log_{2}(2\gamma^{-1})}{n}\biggr)^{1/2} + \sqrt{\dfrac{\log(2\delta^{-1})}{2n}}\biggr\} \nonumber
\end{eqnarray}
where $\overline{\gamma}$ is the lower bound of prior probability $\pi_{\y}$ for all $\y$.
\end{theorem}
\begin{remark}The result of Theorem \ref{theorem:generalization_gap_classification_LDCE} gives a simple characterization for the generalization bound of classification setup from optimal solutions of partially labeled LDCE based on the number of active rendering paths, which is inherent to the structure of DGM. Such dependence of generalization bound on the 
number of active rendering paths $\overline{\sratio}_{n}|\mathcal{L}|$ 
is rather interesting and may provide a new perspective on understanding the 
generalization bound. Nevertheless, there are certain limitations 
regarding the current generalization gap: (1) the active ratio $\overline{\sratio}_{n}$ may change with the sample size unless we put certain constraints 
on the sparsity of switching variables $a$ to reduce the number of 
active optimal rendering paths; (2) the generalization bound is depth-
dependent due to the involvement of the number of rendering paths $|
\mathcal{L}|$. This is mainly because we have not fully taken into 
account all the structures of CNNs for the studying of 
generalization bound. Given some current progress on depth-independent 
generalization bound \cite{Bartlett-2017,Rakhlin-2018}, it is an 
interesting direction to explore whether the techniques in 
these work can be employed to improve $|\mathcal{L}|$ in the 
generalization bound in Theorem \ref{theorem:generalization_gap_classification_LDCE}.
\end{remark}
\subsection{Deconvolutional Generative Model is the Unifying Framework for Networks in the Convnet Family}
\label{sec:unifying-framework}
The structure of the rendering matrices $\render(\ell)$ gives rise to MaxPooling, ReLU, and convolution operators in the CNNs. By modifying the structure of $\render(\ell)$, we can derive different types of networks in the convolutional neural network family. In this section, we define and explore several other interesting variants of DGM: the Residual DGM (ResDGM) and the Dense DGM (DenseDGM). Inference algorithms in these DGMs yield ResNet \cite{he2016deep} and DenseNet \cite{huang2017densely}, respectively. Proofs for these correspondence are given in Appendix C. Both ResNet and DenseNet are among state-of-the-art neural networks for object recognition and popularly used for other visual perceptual inference tasks. These two architectures employ skip connections (a.k.a., shortcuts) to create short paths from early layers to later layers. During training, the short paths help avoid the vanishing-gradient problem and allow the network to propagate and reuse features.

\subsubsection{Residual Deconvolutional Generative Model Yields ResNet}\label{sec:resdrm}
\begin{figure}[t]
    \centering
    \includegraphics[width=0.5\textwidth]{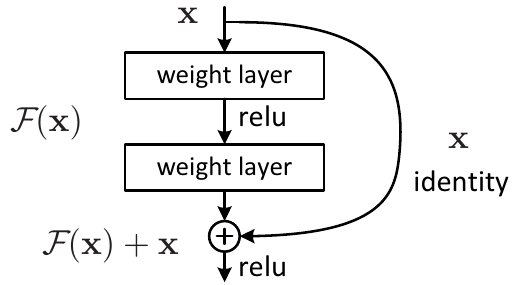} 
    \caption{ResNet building block as in \cite{he2016deep}}
    \vspace*{-3mm}
    \label{fig:resnetblock}
\end{figure}

In a ResNet, layers learn residual functions with reference to the layer inputs. In particular, as illustrated in Fig.~\ref{fig:resnetblock}, layers in a ResNet are reformulated to represent the mapping $F(\acts) + \w_{\text{skip}}\acts$ and the layers try to fit the residual mapping $F(\acts)$ where $\acts$ is the input feature. The term $\w_{\text{skip}}\acts$ accounts for the skip connections/shortcuts \cite{he2016deep}. In order to derive the ResNet, we rewrite the rendering matrix $\render(\ell)$ as the sum of a shortcut matrix $\render_{\text{skip}}(\ell)$ and a rendering matrix, both of which can be updated during the training. The shortcut matrices yields skip connections in \cite{he2016deep}. Note that $\render_{\text{skip}}(\ell)$ depends on the template selecting latent variables $\sparse(\ell)$. In the rest of this section, for clarity, we will refer to $\render(\ell)$ and $\render_{\text{skip}}(\ell)$ as $\render(\trans,\sparse;\ell)$ and $\render_{\text{skip}}(\sparse;\ell)$, respectively, to show their dependency on latent variables in DGM. We define the Residual Deconvolutional Generative Model as follows:
\begin{definition} \label{def:resdrm}
The Residual Deconvolutional Generative Model (ResDGM) is the Deconvolutional Generative Model whose rendering process from layer $\ell$ to layer $\ell -1$, for some $\ell \in \{1,2, \cdots, L\}$, has the residual form as follows:
\begin{align}
    \h(\ell - 1) & : = \left(\render(t,s;\ell) + \render_{\text{skip}}(s;\ell)\right)\h(\ell), \label{eqn:lambdares}
\end{align}
where $\render_{\text{skip}}(t,s;\ell)$ is the shorcut matrices that results in skip connections in the corresponding ResNet. In particular, $\render_{\text{skip}}(t,s;\ell)$ has the following form:
\begin{align}
    \render_{\text{skip}}(s;\ell) = \tilde{\render}_{\text{skip}}(\ell)M(\sparse;\ell),
\end{align}
where $M(\sparse;\ell)\equiv \textrm{diag}\left(\sparse(\ell)\right) \in \R^{D(\ell) \times D(\ell)}$ is a diagonal matrix whose diagonal is the vector $\sparse(\ell)$. This matrix selects the templates for rendering. Furthermore, $\tilde{\render}_{\text{skip}}(\ell)$ is a rendering matrix that is independent of latent variables $\trans$ and $\sparse$.
\end{definition}
The following theorem show that similar to how CNNs can be derived from DGM, ResNet can be derived as a bottom-up inference in ResDGM.
\begin{theorem}
\label{theorem:resldrdrm-resnet}
\normalfont
Inference in ResDGM yields skip connetions. In particular, if the rendering process at layer $\ell$ has the residual form as in Definition \ref{def:resdrm}, the inference at this layer takes the following form:
\begin{align} 
    \acts(\ell) &\equiv \max_{\trans(\ell), \sparse(\ell)}\left\{\left(\render^{\top}(\trans, \sparse;\ell) + \render^{\top}_{\text{skip}}(\sparse;\ell)\right) \acts(\ell-1) + \bias(\ell)\right\} \nonumber\\
	&= \maxpool \left(\relu \left(\convo(\template^{\top}(\ell),\acts(\ell-1)) + \bias(\ell) +\underbrace{\tilde{\render}^{\top}_{\text{skip}}(\ell)\acts(\ell-1)}_{\text{skip connection}}\right)\right) \nonumber \\
	&\overset{d}{=} \maxpool \left(\relu \left(\convo(\w(\ell),\acts(\ell-1)) + \bias(\ell) +\underbrace{ \w_{\text{skip}}(\ell)\acts(\ell-1)}_{\text{skip connection}}\right)\right).
	\label{eqn:res-req}
\end{align}
\end{theorem}
\noindent Here, when $\acts(\ell-1)$ and $\acts(\ell)$ have the same dimensions, $\tilde{\render}_{\text{skip}}(\ell)$ is chosen to be an constant identity matrix in order to derive the parameter-free, identity shortcut among layers of the same size in the ResNet. When $\acts(\ell-1)$ and $\acts(\ell)$ have the different dimensions, $\tilde{\render}_{\text{skip}}(\ell)$ is chosen to be a learnable shortcut matrix which yields the projection shortcut $\w_{\text{skip}}(\ell)$ among layers of different sizes in the ResNet. As mentioned above, identity shortcuts and projection shortcuts are two types of skip connections in the ResNet. The operator $\overset{d}{=}$ implies that discriminative relaxation is applied. 

In practice, the skip connections are usually across two layers or three layers. This is indeed a straightforward extension from the ResDGM. In particular, the ResDGM building block that corresponds to the ResNet building block in \cite{he2016deep} (see Fig.~\ref{fig:resnetblock}) takes the following form:
\begin{align}
\h(\ell - 2) & : = \left(\render(t,s;\ell-1)\render(t,s;\ell) + \render_{\text{skip}}(s;\ell)\right)\h(\ell) \nonumber.
\end{align}
In inference, this ResDGM building block yields the ResNet building block in Fig.~\ref{fig:resnetblock}:
\begin{align} 
	\acts(\ell) &= \relu\biggr(\convo \biggr(\template^{\top}(\ell), \relu \biggr(\convo\left(\template^{\top}(\ell-1),\acts(\ell-2)\right) + \bias(\ell-1)\biggr)\biggr) \nonumber\\
	&\hspace{1in}+ \bias(\ell) + \tilde{\render}^{\top}_{\text{skip}}(\ell)\acts(\ell-2)\biggr) \nonumber \\
	&\overset{d}{=} \relu\biggr(\convo \biggr(\w(\ell), \relu \biggr(\convo\biggr(\w(\ell-1),\acts(\ell-2)\biggr) + \bias(\ell-1)\biggr)\biggr) \nonumber\\
	&\hspace{1in}+ \bias(\ell) + \w_{\text{skip}}(\ell)\acts(\ell-2)\biggr).
	\label{eqn:res-req-block}
\end{align}
\subsubsection{Dense Deconvolutional Generative Model Yields DenseNet}\label{sec:densedrm} 
\begin{figure}[h!]
    \centering
    \includegraphics[width=0.35\textwidth]{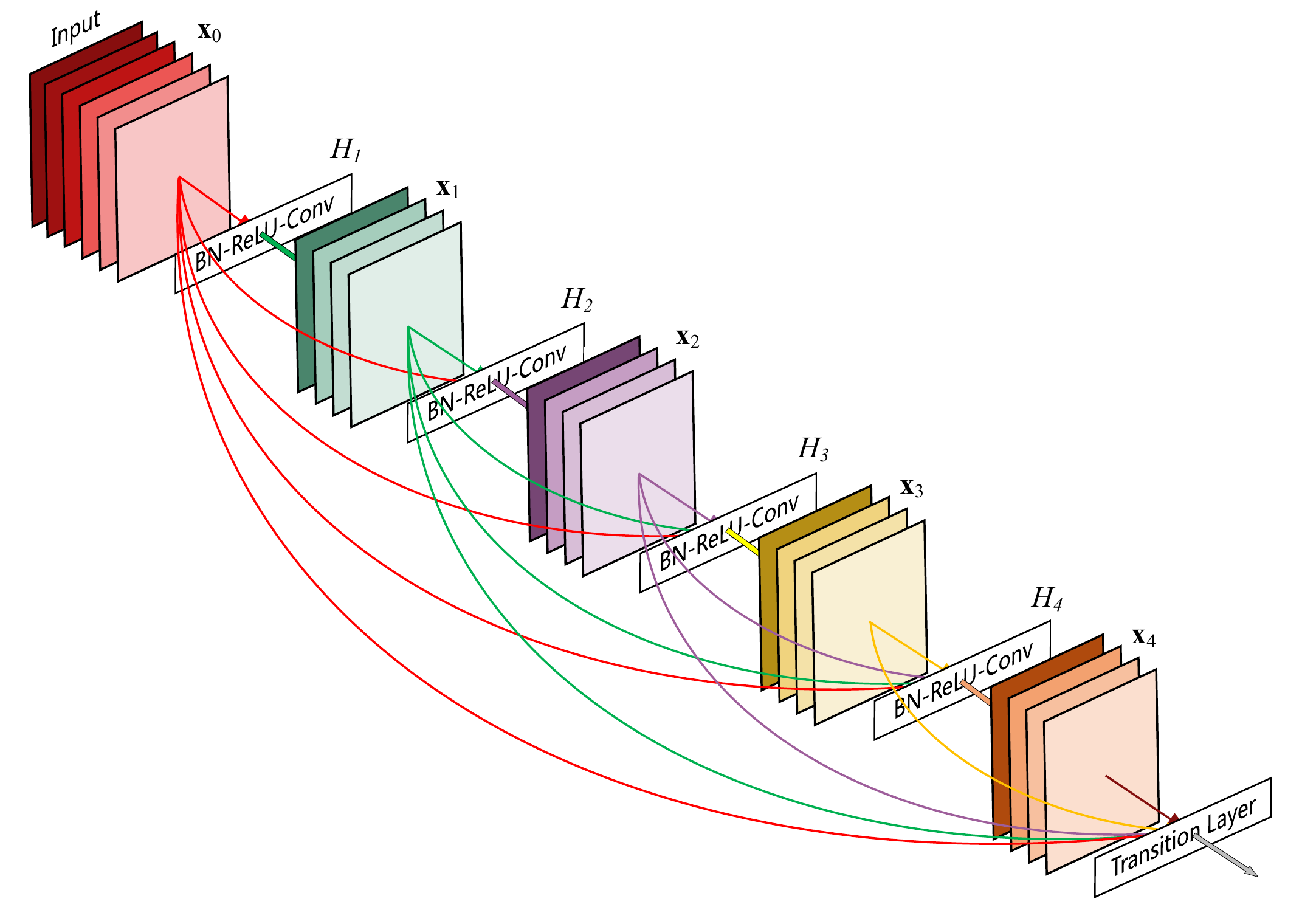} 
    \caption{DenseNet building block as in \cite{huang2017densely}}
    \label{fig:denseblock}
\end{figure}

In a DenseNet \cite{huang2017densely}, instead of combining features through summation, the skip connections concatenate features. In addition, within a building block, all layers are connected to each other (see Fig.~\ref{fig:denseblock}). Similar to how ResNet can be derived from ResDGM, DenseNet can also be derived from a variant of DGM, which we call the Dense Deconvolutional Generative Model (DenseDGM). In DenseDGM, the rendering matrix $\render(\ell)$ is concatenated by an identity matrix. This extra identity matrix, in inference, yields the skip connections that concatenate features at different layers in a DenseNet. We define DenseDGM as follows.

\begin{definition} \label{def:densedrm}
The Dense Deconvolutional Generative Model (DenseDGM) is the Deconvolutional Generative Model  whose rendering process from layer $\ell$ to layer $\ell -1$, for some $\ell \in \{1,2, \cdots, L\}$, has the residual form as follows:
\begin{align}
    \h(\ell - 1) & : = \left[\render(t,s;\ell)\h(\ell), \, \mathbf{1}_{D(\ell)}\h(\ell)\right]. \label{eqn:lambdadense}
\end{align}
\end{definition}
\noindent We again denote $\render(\ell)$ as $\render(t,s;\ell)$ to show the dependency of $\render(\ell)$ on the latent variables $\trans(\ell)$ and $\sparse(\ell)$. The following theorem establishes the connections between DenseNet and DenseDGM.
\begin{theorem} 
\label{theorem:denseldrdrm-densenet}
\normalfont
Inference in DenseDGM yields DenseNet building blocks. In particular, if the rendering process at layer $\ell$ has the dense form as in Definition \ref{def:densedrm}, the inference at this layer takes the following form:
\begin{align} 
	\acts(\ell)&\equiv\left[
\begin{array}{c}
\max \limits_{\trans(\ell), \sparse(\ell)}\left\{\render^{\top}(\trans, \sparse;\ell)\acts(\ell -1) + \bias(\ell)\right\}  \\
\acts(\ell -1) 
\end{array}\right]\nonumber \\
&=\left[
\begin{array}{c}
\maxpool (\relu (\convo(\template^{\top}(\ell),\acts(\ell-1)) + \bias(\ell))) \\
\acts(\ell -1)
\end{array}\right] \nonumber \\
&\overset{d}{=}\left[
\begin{array}{c}
\maxpool (\relu (\convo(\w(\ell),\acts(\ell-1)) + \bias(\ell))) \\
\acts(\ell -1)
\end{array}\right].
	\label{eqn:dense-req}
\end{align}
\end{theorem}
\noindent In Eqn.~\ref{eqn:dense-req}, we concatenate the output $\maxpool \relu (\convo(\w(\ell), \acts(\ell-1))+ \bias(\ell)))$ at layer $\ell$ with the input feature $\acts(\ell -1)$ at layer $\ell -1$ to generate the input to the next layer $\acts(\ell)$, just like in the DenseNet. Proofs for Theorem \ref{theorem:resldrdrm-resnet} and \ref{theorem:denseldrdrm-densenet} can be found in Appendix B. The approach to proving Theorem \ref{theorem:resldrdrm-resnet} can be used to prove the result in Eqn.~\ref{eqn:res-req-block}. 
\newpage 
\section{Appendix C}
\label{Sec:appendix_C}
In this appendix, we provide the proofs for key results in the main text as well as Appendix~\ref{Sec:appendix_A} and Appendix~\ref{Sec:appendix_B}. 
\subsection{Proof for Theorem \ref{theorem:jmap_inference_theorem}: Deriving Convolutional Neural Networks from the Deconvolutional Generative Model} 
\label{sec:proof-lddrm2cnn}
To ease the clarity of the proof presentation, we ignore the normalizing factor $\frac{1}{\sigma^{2}}$ and only consider the proof for two layers, i.e., $L=2$. The argument for $L \geq 3$ is similar and can be derived recursively from the proof for $L=2$. Similar proof holds when the normalizing factor $\frac{1}{\sigma^{2}}$ is considered.
Now, we obtain that
\begin{align}
	& \max \limits_{\z}\left\{\h^{\top}(\y,\z;0)\x + \eta(\y,\z) \right\} \nonumber \\
    = &\max_{\substack{\trans(1),\sparse(1) \\ \trans(2), \sparse(2)}} \biggr\{\sum_{\p_{1} \in \pp(1)}h(1,\p_{1})\sparse(1,\p_{1})\template^{\top}(1,\p_{1})\pad^{\top}(1,\p_{1})\ttrans^{\top}(\trans;1,\p_{1})\x \nonumber \\
    &\hspace{0.5in}+ \sum_{\p_{1} \in \pp(1)} \h(1,\p_{1})\sparse(1,\p_{1})\bias(\trans;1,\p_{1}) + \sum_{\p_{2} \in \pp(2)} \muy(\y;\p_{2})\sparse(2,\p_{2})\bias(\trans;2,\p_{2})\biggr\}  \nonumber \\
    = &\max_{\substack{\trans(1),\sparse(1) \\ \trans(2), \sparse(2)}} \biggr\{\sum_{\p_{1} \in \pp(1)}\h(1,\p_{1})\sparse(1,\p_{1})\left(\template^{\top}(1,\p_{1})\pad^{\top}(1,\p_{1})\ttrans^{\top}(\trans;1,\p_{1})\x + \bias(\trans;1,\p_{1})\right) \nonumber \\
    &\hspace{0.5in} + \sum_{\p_{2} \in \pp(2)} \muy(\y;\p_{2})\sparse(2,\p_{2})\bias(\trans;2,\p_{2})\biggr\}  \nonumber \\
    \overset{(a)}{=} &\max_{\trans(2),\sparse(2)} \biggr\{\sum_{\p_{1} \in \pp(1)}\h(1,\p_{1})\max_{\trans(1,\p_{1}), \sparse(1,\p_{1})}\sparse(1,\p_{1})\left(\template^{\top}(1,\p_{1})\pad^{\top}(1,\p_{1})\ttrans^{\top}(\trans;1,\p_{1})\x + \bias(\trans;1,\p_{1})\right) \nonumber \\
    &\hspace{0.5in} + \sum_{\p_{2} \in \pp(2)} \muy(\y;\p_{2})\sparse(2,\p_{2})\bias(\trans;2,\p_{2})\biggr\}=A,  \nonumber
\end{align}
where equation in (a) is due to the non-negativity assumption that $\h(1,\p_{1}) \ge 0$, just as in max-sum and max-product message passing. We define $\acts(1,\p_{1})$ as follows:
\begin{align}
\acts(1,\p_{1})=\max_{\trans(1,\p_{1}), \sparse(1,\p_{1})}\sparse(1,\p_{1})\left(\template^{\top}(1,\p_{1})\pad^{\top}(1,\p_{1})\ttrans^{\top}(\trans;1,\p_{1})\x + \bias(\trans;1,\p_{1})\right), \nonumber
\end{align}
and let $\acts(1)=(\acts(1,\p_{1}))_{\p_{1}\in \pp(1)}$ be the vector of $\acts(1,\p_{1})$. The following holds:
\begin{align}
    A = &\max_{\trans(2),\sparse(2)} \biggr\{\h^{\top}(1)\acts(1) + \sum_{\p_{2} \in \pp(2)} \muy(\y;\p_{2})\sparse(2,\p_{2})\bias(\trans;2,\p_{2})\biggr\} \nonumber \\
    \overset{(b)}{=}&\max_{\trans(2),\sparse(2)} \biggr\{\sum_{\p_{2} \in \pp(2)}\muy(y; \p_{2})\sparse(2,\p_{2})\template^{\top}(2,\p_{2})\pad^{\top}(2,\p_{2})\ttrans^{\top}(\trans;2,\p_{2})\acts(1) \nonumber \\
    & \hspace{16 em} + \sum_{\p_{2} \in \pp(2)} \muy(\y;\p_{2})\sparse(2,\p_{2})\bias(\trans;2,\p_{2})\biggr\} \nonumber \\
   \overset{(c)}{=}& \sum_{\p_{2} \in \pp(2)}\muy(y; \p_{2})\underbrace{\max_{\trans(2),\sparse(2)}\sparse(2,\p_{2})\biggr(\template^{\top}(2,\p_{2})\pad^{\top}(2,\p_{2})\ttrans^{\top}(\trans;2,\p_{2})\acts(1) + \bias(\trans;2,\p_{2})\biggr)}_{\acts(2,\p_{2})} \nonumber \\
    = &\muy^{\top}(\y)\acts(2)  \nonumber.
\end{align}
Here $\acts(2)=(\acts(2,\p_{2}))_{\p_{2}\in \pp(2)}$ is the vector of $\acts(2,\p_{2})$, and in line (b) we substitute $\h(1)$ by:
\begin{align}
\h(1)=\sum \limits_{\p\in \pp(2)}  \sparse(2, \p_{2})\ttrans(t;2,\p_{2}) \pad(2, \p_{2})\template(2, \p_{2}) \muy(\y;\p_{2}). \nonumber
\end{align}
Notice that line (a) and (c) form a recursion. Therefore, we finish proving that the feedforward step in CNNs is the latent variable inference in DGM if $\acts(1)$ has the structure of the building block of CNNs, i.e.  $\maxpool\left(\relu\left(\convo()\right)\right)$. Indeed,
\begin{align}
    \acts(1) &= (\acts(1, \p(1)))_{\p(1)\in \pp(1)} \nonumber \\
    &= \left(\max_{\trans(1,\p_{1}), \sparse(1,\p_{1})}\sparse(1,\p_{1})\left(\template^{\top}(1,\p_{1})\pad^{\top}(1,\p_{1})\ttrans^{\top}(\trans;1,\p_{1})\x + \bias(\trans;1,\p_{1})\right)\right)_{\p(1)\in \pp(1)} \nonumber \\
    & = \left(\max\left(\left(\relu\left(\template^{\top}(1,\p_{1})\pad^{\top}(1,\p_{1})\ttrans^{\top}(\trans;1,\p_{1})\x + \bias(\trans;1,\p_{1})\right)\right)_{\trans(1,\p_{1})=0,1,2,3}\right)\right)_{\p(1)\in \pp(1)} \label{eqn:relu-maxpool} \\
    & = \maxpool\left(\relu\left(\convo\left(\template^{\top}(1), \x\right)+ \bias(\trans;1)\right)\right) \nonumber \\
    & \overset{d}{=} \maxpool\left(\relu\left(\convo\left(W(1), \x\right)+ \bias(\trans;1)\right)\right), \nonumber
\end{align}
where $W(1)=\template^{\top}(1)$ corresponds to the weights at layer 1 of the CNN, and $\overset{d}{=}$ implies that discriminative relaxation is applied. In Eqn.~\ref{eqn:relu-maxpool}, since $\sparse(1,\p_{1})\in \{0,1\}$, $\max \limits_{\sparse(1,\p_{1})}(\sparse(1,\p_{1}) \times .)$ is equivalent to $\max(\,.\,,0)$ and, therefore, yields the ReLU function. Also, in order to compute $\max \limits_{\trans(1,\p_{1})}()$, we take the brute force approach, computing the function inside the parentheses for all possible values of $\trans(1,\p_{1})$ and pick the maximum one. This procedure is equivalent to the MaxPool operator in CNNs. Here we can make the bias term $\bias(\trans;1)$ independent of $\trans$ and $\bias(\trans;1,\p_{1})$ are the same for all pixels $\p_{1}$ in the same feature map of $\h(1)$ as in CNNs. 
Similarly, $\acts(2)= \maxpool\left(\relu\left(\convo\left(W(2), \acts(1)\right)+ \bias(2)\right)\right)$. Thus, we obtain the conclusion of the theorem for $L=2$.

\subsection{Proof for Theorem \ref{theorem:resldrdrm-resnet}: Deriving the Residual Networks from the Residual Deconvolutional Generative Model}
Similar to the proof in Section \ref{sec:proof-lddrm2cnn} above, when the rendering process at each layer in ResDGM is as in Eqn.~\ref{eqn:lambdares}, the activations $\acts(\ell)$ is given by:
\begin{align}
\acts(\ell)&=\left(\max_{\trans(\ell,\p), \sparse(\ell,\p)}\sparse(\ell,\p)\left(\left(\template^{\top}(\ell,\p)\pad^{\top}(\ell,\p)\ttrans^{\top}(\trans;\ell,\p)+\tilde{\render}^{\top}_{\text{skip}}(\ell,\p)\right)\acts(\ell-1)+ \bias(\trans;\ell,\p)\right)\right)_{\p\in \pp(\ell)} \nonumber \\
    & = \biggr(\max\biggr(\biggr(\relu\biggr(\template^{\top}(\ell,\p)\pad^{\top}(\ell,\p)\ttrans^{\top}(\trans;\ell,\p)\acts(\ell-1) + \bias(\trans;\ell,\p)  \nonumber \\ &\hspace{1.25in}+\tilde{\render}^{\top}_{\text{skip}}(\ell,\p)\acts(\ell-1)\biggr)\biggr)_{\trans(\ell,\p)=0,1,2,3}\biggr)\biggr)_{\p\in \pp(\ell)} \nonumber \\
    & = \maxpool\left(\relu\left(\convo\left(\template^{\top}(\ell,\p), \acts(\ell-1)\right)+ \bias(\trans;\ell) + \tilde{\render}^{\top}_{\text{skip}}(\ell)\acts(\ell-1)\right)\right) \nonumber \\
    &\overset{d}{=} \maxpool \left(\relu \left(\convo(W^{(\ell)},\acts(\ell-1)) + \bias(\trans;\ell) +\underbrace{ W_{\text{skip}}(\ell)\acts(\ell-1)}_{\text{skip connection}}\right)\right).
\end{align}
Here we can again make the bias term $\bias(\trans;\ell)$ independent of $\trans$ and $\bias(\trans;\ell,\p)$ are the same for all pixels $\p_{\ell}$ in the same feature map of $\h(\ell)$ as in CNNs. We obtain the conclusion of the Theorem \ref{theorem:resldrdrm-resnet}.
\subsection{Proof for Theorem \ref{theorem:denseldrdrm-densenet}: Deriving the Densely Connected Networks from the Dense Deconvolutional Generative Model}
Similar to the proof for Theorem \ref{theorem:resldrdrm-resnet} above, when the rendering process at each layer in ResDGM is as in Eqn.~\ref{eqn:lambdadense}, the activations $\acts(\ell)$ is given by:
\begin{align}
    \acts(\ell)&\equiv\left[
        \begin{array}{c}
        \left(\max \limits_{\sparse(\ell,\p),\trans(\ell,\p)}\left(\sparse(\ell,\p)\left(\template^{\top}(\ell,\p)\pad^{\top}(\ell,\p)\ttrans^{\top}(\trans;\ell,\p)\acts(\ell -1)+\bias(\trans;\ell,\p)\right)\right)\right)_{\p\in \pp(\ell)}  \\
        \acts(\ell -1) 
        \end{array}\right]\nonumber \\
     &=\left[
        \begin{array}{c}
        \left(\max\left(\left(\relu\left(\template^{\top}(\ell,\p)\pad^{\top}(\ell,\p)\ttrans^{\top}(\trans;\ell,\p)\acts(\ell-1) + \bias(\trans;\ell,\p)\right)\right)_{\trans(\ell,\p)=0,1,2,3}\right)\right)_{\p\in \pp(\ell)} \\
        \acts(\ell -1) 
        \end{array}\right]\nonumber \\
    &=\left[
        \begin{array}{c}
        {\rm MaxPool} ({\rm ReLu} ({\rm Conv}(\template^{\top}(\ell,\p),\acts(\ell-1)) + \bias(\trans;\ell))) \\
        \acts(\ell -1)
        \end{array}\right] \nonumber \\
        &\overset{d}{=}\left[
        \begin{array}{c}
        {\rm MaxPool} ({\rm ReLu} ({\rm Conv}(W^{(\ell)},\acts(\ell-1)) + \bias(\trans;\ell))) \\
        \acts(\ell -1)
        \end{array}\right].
\end{align}
Again we can make the bias term $\bias(\trans;\ell)$ independent of $\trans$ and $\bias(\trans;\ell,\p)$ are the same for all pixels $\p_{\ell}$ in the same feature map of $\h(\ell)$ as in CNNs. We obtain the conclusion of the Theorem \ref{theorem:denseldrdrm-densenet}.

\subsection{Proving that the Parametrized Joint Prior p(y,z) is a Conjugate Prior}
\label{sec:conjugate-prior-proof}
Again, for simplicity, we only consider the proof for two layers. The argument for $L \ge 3$ is similar and can be derived recursively from the proof for $L = 2$. In the derivation below, $h(2)$ is $\muy(\y)$. As in Eqn.\ref{eqn:joint-prior} in the definition of the DGM, the joint prior of $y$ and $z$ is given by:

\begin{align}
    p(y,z) &\propto \exp \left(
     \frac{1}{\sigma^{2}}\bias^{\top}(\trans;2)(\sparse(2)\odot \h(2)) + \frac{1}{\sigma^{2}}\bias^{\top}(\trans;1)(\sparse(1)\odot \h(1)) +\ln\pi_{\y} \right) \nonumber  \\
     &= \exp \left(
     \frac{1}{\sigma^{2}}\h^{\top}(2)(\bias(\trans;2)\odot \sparse(2)) + \frac{1}{\sigma^{2}}\h^{\top}(1)(\bias(\trans;1)\odot \sparse(1)) +\ln\pi_{\y} \right) \nonumber \\
     &= \exp \left(
     \frac{1}{\sigma^{2}}\h^{\top}(2)(\bias(\trans;2)\odot \sparse(2)) + \frac{1}{\sigma^{2}}\h^{\top}(2)\render^{\top}(2)(\bias(\trans;1)\odot \sparse(1)) +\ln\pi_{\y} 
     \right) \nonumber\\
     &= \exp \left(
     \frac{1}{\sigma^{2}}\h^{\top}(2)\left[\render^{\top}(2)(\bias(\trans;1)\odot \sparse(1))+\bias(\trans;2)\odot \sparse(2)\right] +\ln\pi_{\y} 
     \right) \label{eqn:prior-form}
\end{align}

Furthermore, as explained in Section \ref{Section:proof_cross_entropy} of Appendix D,  due to the constant norm assumption with $h(y,z;0)$, the likelihood $p(x|y,z)$ is estimated as follows:
\begin{align}
     p(x|y,z) &\propto \exp \left(\frac{1}{\sigma^{2}} h^{\top}(y,z;0)x\right) \nonumber \\
     &= \exp \left(\frac{1}{\sigma^{2}} h^{\top}(2)\render^{\top}(2)\render^{\top}(1)x\right) \nonumber
\end{align}
The posterior $p(y,z|x)$ is given by:
\begin{align}
    p(y,z|x) &\propto \exp \left(
     \frac{1}{\sigma^{2}}\h^{\top}(2)\left[\render^{\top}(2)(\bias(\trans;1)\odot \sparse(1))+ \render^{\top}(2)\render^{\top}(1)x +\bias(\trans;2)\odot \sparse(2)\right] +\ln\pi_{\y} 
     \right) \nonumber\\
     &= \exp \left(
     \frac{1}{\sigma^{2}}\h^{\top}(2)\left[\render^{\top}(2)(\bias(\trans;1)\odot \sparse(1)+ \render^{\top}(1)x) +\bias(\trans;2)\odot \sparse(2)\right] +\ln\pi_{\y} 
     \right) \label{eqn:posterior-form}
\end{align}

Comparing Eqn. \ref{eqn:prior-form} and Eqn. \ref{eqn:posterior-form}, we see that the prior and the posterior have the same functional form. This completes the proof.

\subsection{Deriving Other Components in the Convolutional Neural Networks}
\subsubsection{Deriving the Leaky Rectified Linear Unit (Leaky ReLU)}
The Leaky ReLU can be derived from the DGM in the same way as we derive the ReLU in Section \ref{sec:proof-lddrm2cnn}, but instead of $\sparse(\ell) \in \{0,1\}$, we now let $\sparse(\ell) \in \{\alpha,1\}$, where $\alpha$ is a small positive constant. Then, in Eqn.~\ref{eqn:relu-maxpool}, $\max \limits_{\sparse(\ell,\p)}(\sparse(\ell,\p) \times .)$ is equivalent to $[.<0](\alpha \times .)+[.>=0](.)$, which is the LeakyReLU function. Note that compared to Eqn.~\ref{eqn:relu-maxpool}, here we replace the layer number $(1)$ by $(\ell)$ since the same derivation can be applied at any layer.

\subsection{Proofs for Batch Normalization}
\label{sec:deriving-bn}
In this section we will derive the batch normalization from a 2-layer DGM. The same derivation can be generalized to the case of K-layer DGM.

In order to derive the Batch Normalization, we normalize the intermediate rendering image $\h(\ell)$.
\begin{align}
    &\h(\y,\z;0) \nonumber \\
    &= \sum_{p_{1}\in \pp(1)} \sparse(1,\p_{1})\ttrans(\trans;1,\p_{1}) \pad(1,\p_{1})\template(1,\p_{1})\frac{1}{\sigma_{h}(1,\p_{1})} \left(\h(1,\p_{1}) - \Expect_\h(1,\p_{1})\right) - \Expect_{\h(\y,\z;0)} \nonumber \\
    &= \sum_{p_{1}\in \pp(1)}  \sparse(1,\p_{1})\ttrans(\trans;1,\p_{1}) \pad(1,\p_{1})\template(1,\p_{1})\frac{1}{\sigma_{h}(1,\p_{1})} \nonumber \\
    &\hspace{0.5in}\times\biggr(\biggr(\sum_{p_{2}\in \pp(2)}  \sparse(1,\p_{2})\ttrans(\trans;2,\p_{2}) \pad(2,\p_{2})\template(2,\p_{2})\frac{1}{\sigma_\h(\y,\z;0)} \h(\y,\z;0)\biggr)(1,\p_{1}) - \Expect_\h(1,\p_{1})\biggr) \nonumber \\
    &\hspace{0.5in}- \Expect_{\h(\y,\z;0)}
\end{align}

During inference, we de-mean the input image and find $z^{*}=\arg\max\limits_{z}\h^{\top}(\y,\z;0)(\x-\Expect_{\h(\y,\z;0)})$. In particular, the inference can be derived as follows:
\begin{align}
&\max\limits_{z}\h^{\top}(\y,\z;0)(\x-\Expect_{\h(\y,\z;0)}) \nonumber \\
=&\max\limits_{z}\biggr(\sum_{p_{1}\in \pp(1)}  \sparse(1,\p_{1})\ttrans(\trans;1,\p_{1}) \pad(1,\p_{1})\template(1,\p_{1})\frac{1}{\sigma_{h}(1,\p_{1})} \biggr(\h(1,\p_{1}) - \Expect_\h(1,\p_{1})\biggr)\biggr)^{\top}(\x-\Expect_{\h(\y,\z;0)}) \nonumber \\
&- \underbrace{\Expect_{\h(\y,\z;0)}^{\top}(\x-\Expect_{\h(\y,\z;0)})}_{\text{const w.r.t } \y, \z} \nonumber \\
\overset{a}{\approx}&\max\limits_{z}\underbrace{\sum_{p_{1}\in \pp(1)} \biggr( \underbrace{\sparse(1,\p_{1})\biggr(\h(1,\p_{1}) - \Expect_\h(1,\p_{1}) \biggr)\frac{1}{\sigma_{h}(1,\p_{1})}}_{\text{scalar}}\underbrace{\underbrace{\template^{\top}(1,\p_{1})\pad^{\top}(1,\p_{1})\ttrans^{\top}(\trans;1,\p_{1})}_{\text{row vector}} \underbrace{(\x-\Expect_{i})}_{\text{column vector}}}_{\text{scalar}}\biggr)}_{\text{dot product}} \nonumber \\
&+ \text{ const} : = A + \text{const}. \nonumber
\end{align}
Direct computation leads to
\begin{align}
A =&\max\limits_{z}\sum_{p_{1}\in \pp(1)} \biggr( \sparse(1,\p_{1})\biggr(\h(1,\p_{1}) - \Expect_\h(1,\p_{1}) \biggr)\underbrace{\frac{\sigma_\acts(1,\p_{1})}{\sigma_{h}(1,\p_{1})}}_{\alpha(1,\p_{1})}\underbrace{\frac{1}{\sigma_\acts(1,\p_{1})}}_{\text{normalize}}\nonumber \\
&\times\underbrace{(\template^{\top}(1,\p_{1})\pad^{\top}(1,\p_{1})\ttrans^{\top}(\trans;1,\p_{1})\x-\Expect[\template^{\top}(1,\p_{1})\pad^{\top}(1,\p_{1})\ttrans^{\top}(\trans;1,\p_{1})\x])}_{\text{de-mean}}\biggr) \nonumber \\
=&\max\limits_{\z(2)}\sum_{p_{1}\in \pp(1)}\max\limits_{\sparse(1,\p_{1}), \trans(1,\p_{1})} \biggr( \sparse(1,\p_{1})\biggr(\h(1,\p_{1}) - \Expect_\h(1,\p_{1}) \biggr) \nonumber \\
&\times\text{BatchNorm}(\template^{\top}(1,\p_{1})\pad^{\top}(1,\p_{1})\ttrans^{\top}(\trans;1,\p_{1})\x;\alpha(1,\p_{1}))\biggr)  \nonumber \\
=&\max\limits_{\z(2)}\biggr(\h(1) - \Expect_\h(1) \biggr)^{\top} \biggr( \maxpool(\relu(\text{BatchNorm}(\convo(\template(1),\x);\alpha(1),0)))\biggr) \label{eqn:bn-layer1}
\end{align}

In line (a), we approximate $\Expect_{\h(\y,\z;0)}$ by its empirical value $\Expect_{i}$.  The de-mean and normalize operators with the scale parameter $\alpha(1)$ and the shift parameter $\beta(1)= 0$ in the equations above already have the form of batch normalization. Note that when the model is trained with Stochastic Gradient Descent (SGD), the scale and shift parameters at each layer also account for the error in evaluating statistics of the activations using the mini-batches. Thus, $\beta(1)$ is not 0 any more, but a parameter learned during the training. Also, in the equations above, $\acts(1)$ is the activations at layer 1 in CNNs and given by:
\begin{align}
\acts(1)=\maxpool(\relu(\text{Normalize}(\text{Demean}(\convo(\template(1),\x))))),
\end{align}
where $\sigma_\acts(1)$ is the standard deviation of the $\acts(1)$. Eqn.\ref{eqn:bn-layer1} can be expressed in term of $\acts(1)$ as follows:

\begin{align}
&\max\limits_{\z(2)}\biggr(\h(1) - \Expect_\h(1) \biggr)^{\top}\acts(1) + \text{ const} \nonumber \\
=&\max\limits_{\z(2)}\h^{\top}(1)(\acts(1) - \Expect_\acts(1)) + (\h^{\top}(1)\Expect_\acts(1) - \Expect^{\top}_\h(1)\acts(1)) + \text{ const} \nonumber \\
=&\max\limits_{\z(2)}\sum_{p_{2}\in \pp(2)} \biggr( \sparse(2,\p_{2})\muy(\y,\p_{2})\underbrace{\frac{\sigma_\acts(2,\p_{2})}{\sigma_{\muy(\y)}(\p_{2})}}_{\alpha(2,\p_{2})}\underbrace{\frac{1}{\sigma_\acts(2,\p_{2})}}_{\text{normalize}}\nonumber \\
&\times\biggr(\underbrace{\template^{\top}(2,\p_{2})\pad^{\top}(2,\p_{2})\ttrans^{\top}(\trans;2,\p_{2})\acts(1)-\Expect[\template^{\top}(2,\p_{2})\pad^{\top}(2,\p_{2})\ttrans^{\top}(\trans;2,\p_{2})\acts(1)]}_{\text{de-mean}} \nonumber \\
&\hspace{1 in}+ \underbrace{\h(1,\p_{1})\Expect_\acts(1,\p_{1}) - \Expect_\h(1,\p_{1})\acts(1,\p_{1})}_{\beta(2,\p_{2})}\biggr)\biggr) + \text{ const} \nonumber \\
=&\max\limits_{\z(2)}\mu^{\top}(\y) \biggr( \maxpool(\relu(\text{BatchNorm}(\convo(\template(2),\acts(1));\alpha(2),\alpha(2)\odot\beta(2))))\biggr) + \text{ const}
\end{align}

The batch normalization at this layer of the CNN has the scale parameter $\alpha(2)$ and the shift parameter is the element-wise product of $\alpha(2)$ and $\beta(2)$. 
\subsection{Proofs for connection between DGM and cross entropy}
\label{Section:proof_cross_entropy}
\paragraph{PROOF OF THEOREM \ref{theorem:cross_entropy_LD_DRM_full}}
(a) To ease the presentation of proof, we denote the following key notation
\begin{align}
A  : =  \max_{(\z_{i})_{i=1}^{n}, \theta \in \mathcal{A}_{\gamma}}\dfrac{1}{n} \sum_{i=1}^{n} \ln p(\y_{i}|\x_{i}, \z_{i};\theta). \nonumber
\end{align} 
From the definition of $\mathcal{A}_{\gamma} = \left\{\theta: \ \|\h(\y,\z;0)\| = \gamma \right\}$, we achieve the following equations
\begin{eqnarray}
  A  & =& \max_{\theta \in \mathcal{A}_{\gamma}} \dfrac{1}{n} \sum_{i=1}^{n} \max_{\z_{i}} \log p(\y_{i}|\x_{i}, \z_{i};\theta) \nonumber \\
    &=&\max_{\theta \in \mathcal{A}_{\gamma}} \dfrac{1}{n} \sum_{i=1}^{n} \max_{\z_{i}} \log \frac{ p(\x_{i}|\y_{i}, \z_{i};\theta)p(\y_{i}|\z_{i};\theta)}{ \sum \limits_{\y=1}^{K} p(\x_{i}|\y, \z_{i};\theta)p(\y|\z_{i};\theta)} \nonumber \\
    &=&\max_{\theta \in \mathcal{A}_{\gamma}} \dfrac{1}{n} \sum_{i=1}^{n} \max_{\z_{i}} \biggr\{\log p(\x_{i}|\y_{i}, \z_{i};\theta) + \log p(\y_{i},\z_{i}|\theta) - \log \biggr(\sum \limits_{\y=1}^{K} p(\x_{i}|\y, \z_{i};\theta)p(\y,\z_{i}|\theta)\biggr) \biggr\}. \nonumber
\end{eqnarray}
From the formulation of DGM, we have the following formulation of prior probabilities $p(\y,\z|\theta)$
\begin{eqnarray}
p(\y,\z|\theta) = \frac{\exp\biggr(\dfrac{\eta(\y,\z)}{\sigma^{2}}\biggr)\pi_{\y}}{\sum\limits_{\y',\z'}\exp\biggr(\dfrac{\eta(\y',z')}{\sigma^{2}}\biggr)\pi_{\y'}}. \nonumber
\end{eqnarray}
for all $(\y,\z)$. By means of the previous equations, we eventually obtain that
\begin{eqnarray}
   A & = & \max_{\theta \in \mathcal{A}_{\gamma}} \dfrac{1}{n} \sum_{i=1}^{n} \max_{\z_{i}} \biggr\{\log p(\x_{i}|\y_{i}, \z_{i};\theta) + \log \left(\dfrac{\exp\biggr(\dfrac{\eta(\y_{i},\z_{i})}{\sigma^{2}}\biggr)\pi_{\y_{i}}}{\sum \limits_{\y,\z}\exp\biggr(\dfrac{\eta(\y,\z)}{\sigma^{2}}\biggr)\pi_{\y}}\right) \nonumber \\
    & & \hspace { 6 em} - \log \sum \limits_{\y=1}^{K} \biggr(p(\x_{i}|\y, \z_{i};\theta)\frac{\exp\biggr(\dfrac{\eta(\y,\z_{i})}{\sigma^{2}}\biggr)\pi_{\y}}{\sum\limits_{\y',\z'}\exp\biggr(\dfrac{\eta(\y',\z')}{\sigma^{2}}\biggr)\pi_{\y'}}\biggr) \biggr\} \nonumber \\
    & = &\max_{\theta \in \mathcal{A}_{\gamma}} \dfrac{1}{n} \sum_{i=1}^{n} \max_{\z_{i}}\left( \log p(\x_{i}|\y_{i}, \z_{i};\theta) + \log \biggr(\exp\biggr(\dfrac{\eta(\y_{i},\z_{i})}{\sigma^{2}}\biggr)\pi_{\y_{i}}\biggr)\right) \nonumber \\
    & &  \hspace { 6 em} - \log \biggr(\sum \limits_{\y=1}^{K} \biggr(p(\x_{i}|\y, \z_{i};\theta)\exp\biggr(\dfrac{\eta(\y,\z_{i})}{\sigma^{2}} \biggr)\pi_{\y} \biggr). \nonumber
\end{eqnarray}
By defining $\acts_{i}(\y, \z_{i}) := \log p(\x_{i}|\y, \z_{i};\theta) + \log \biggr(\exp\biggr(\dfrac{\eta(\y,\z_{i})}{\sigma^{2}} \biggr)\pi_{\y}\biggr)$ for all $1 \leq i \leq n$, the above equation can be rewritten as
\begin{eqnarray}
   A & = & \max_{\theta \in \mathcal{A}_{\gamma}} \dfrac{1}{n} \sum_{i=1}^{n} \max_{\z_{i}} \left( \log\left(\exp(\acts_{i}(\y_{i}, \z_{i}))\right) - \log \left(\sum \limits_{\y=1}^{K} \exp(\acts_{i}(\y, \z_{i}))\right)\right) \nonumber \\
    & = & \max_{\theta \in \mathcal{A}_{\gamma}} \dfrac{1}{n} \sum_{i=1}^{n} \max_{\z_{i}} \log \left(\text{Softmax}(\acts_{i}(\y_{i}, \z_{i}))\right) \nonumber \\
   & \geq & \max_{\theta \in \mathcal{A}_{\gamma}} \dfrac{1}{n} \sum_{i=1}^{n} \log \left(\text{Softmax}\left(\max_{\z_{i}}\acts_{i}(\y_{i}, \z_{i}))\right)\right) = B. \nonumber
\end{eqnarray}
By means of direct computation, the following equations hold
\begin{eqnarray}
   B & = & \max_{\theta \in \mathcal{A}_{\gamma}} \dfrac{1}{n} \sum_{i=1}^{n} \log \left(\text{Softmax}\left(\max_{\z_{i}} \log p(\x_{i}|\y_{i}, \z_{i};\theta) + \eta(\y_{i},\z_{i})/\sigma^{2}+\log\pi_{\y_{i}}\right)\right) \nonumber \\
    & = & \max_{\theta \in \mathcal{A}_{\gamma}} \dfrac{1}{n} \sum_{i=1}^{n} \log \left(\text{Softmax}\left(\max_{\z_{i}}-\frac{\|\x_{i} - \muy_{\y_{i},\z_{i}}\|^{2}}{2 \sigma^{2}} + \dfrac{\eta(\y_{i},\z_{i})}{\sigma^{2}}+\log\pi_{\y_{i}}\right)\right) \nonumber \\
    & \overset{(i)}{=} & \max_{\theta \in \mathcal{A}_{\gamma}} \dfrac{1}{n} \sum_{i=1}^{n} \log \left(\text{Softmax}\left(\max_{\z_{i}}\left(\dfrac{\h^{\top}(\y_{i}, \z_{i}; 0)\x_{i} + \eta(\y_{i},\z_{i})}{\sigma^{2}}\right) + \bias_{\y_{i}} \right)\right) \nonumber \\\
    & = & - \min_{\theta \in \mathcal{A}_{\gamma}} \dfrac{1}{n} \sum_{i=1}^{n} - \log \left(\text{Softmax}\left(\max_{\z_{i}}\left(\dfrac{\h^{\top}(\y_{i}, \z_{i}; 0)\x_{i} + \eta(\y_{i},\z_{i})}{\sigma^{2}}\right) + \bias_{\y_{i}} \right)\right) \nonumber \\
    & = & - \min_{\theta \in \mathcal{A}_{\gamma}} \dfrac{1}{n} \sum_{i=1}^{n} - \log q(\y_{i}|\x_{i}) = - \min_{\theta \in \mathcal{A}_{\template}} H_{p,q}(\y|\x) \nonumber
\end{eqnarray}
where equation (i) is due to the constant norm assumption with rendered images $\h(\y,\z; 0)$. Therefore, we achieve the conclusion of part (a) of the theorem.

(b) Regarding the upper bound, from the definition of $\overline{\z}_{i}$, we obtain that
\begin{eqnarray}
& & \hspace{ - 6 em} \max_{\z_{i}} \log \left(\text{Softmax}(\acts_{i}(\y_{i}, \z_{i}))\right) - \log \left(\text{Softmax}\left(\max_{\z_{i}}\acts_{i}(\y_{i}, \z_{i}))\right)\right) \nonumber \\
& & = \log \left(\text{Softmax}(\acts_{i}(\y_{i}, \overline{\z}_{i}))\right) - \log \left(\text{Softmax}\left(\max_{\z_{i}}\acts_{i}(\y_{i}, \z_{i}))\right)\right) \nonumber \\
& & \leq \log \dfrac{\max \limits_{\z_{i}} \exp(\acts_{i}(\y_{i}, \z_{i}))}{\sum \limits_{\y=1}^{K} \exp(\acts_{i}(\y, \overline{\z}_{i}))} - \log \left(\text{Softmax}\left(\max_{\z_{i}}\acts_{i}(\y_{i}, \z_{i}))\right)\right) \nonumber \\
& & = \log \biggr( \sum \limits_{\y=1}^{K} \exp(\max \limits_{\z_{i}} \acts_{i}(\y,\z_{i}))\biggr) - \log \biggr(\sum \limits_{\y=1}^{K} \exp(\acts_{i}(\y, \overline{\z}_{i}))\biggr) \nonumber \\
& & \leq \log K + \max \limits_{\y} \max \limits_{\z_{i}} \acts_{i}(\y,\z_{i}) - \max \limits_{\y} \acts_{i}(\y,\overline{\z}_{i}) \nonumber \\
& & \leq \log K + \max \limits_{\y}\biggr(\max \limits_{\z_{i}} \acts_{i}(\y,\z_{i}) - \acts_{i}(\y,\overline{\z}_{i})\biggr) \nonumber
\end{eqnarray}
for any $1 \leq i \leq n$. As a consequence, we obtain the conclusion of part (b) of the theorem.
\paragraph{PROOF OF THEOREM \ref{theorem:approximation_unsupervised_LD_DRM}}
(a) From the definitions of $(\overline{\y}_{i},\overline{\z}_{i})$ and $(\widetilde{\y}_{i},\widetilde{\z}_{i})$, we obtain that
\begin{eqnarray}
U_{n} & = & \min \limits_{\theta} \dfrac{1}{n} \sum \limits_{i=1}^{n} \biggr(\frac{1}{2}\|\x_{i} - \h(\overline{\y}_{i},\overline{\z}_{i}; 0)\|^{2} - \log \pi_{\overline{\y}_{i},\overline{\z}_{i}} \biggr) \nonumber \\
& = & \min \limits_{\theta} \biggr\{\dfrac{1}{n} \sum \limits_{i=1}^{n} \biggr(\frac{1}{2}\|\x_{i} - \h(\overline{\y}_{i},\overline{\z}_{i}; 0)\|^{2} -\eta(\overline{\y}_{i},\overline{\z}_{i}) - \log \pi_{\overline{\y}_{i}} \biggr) \nonumber \\
& & \hspace{ 12 em} + \log \biggr(\sum \limits_{(\y',\z') \in \Jcal} \exp(\eta(\y',z') + \log \pi_{\y'}) \biggr)\biggr\} = A \nonumber
\end{eqnarray}
By means of direct computation, the following inequality holds
\begin{eqnarray}
A & \leq &  \min \limits_{\theta} \biggr\{\dfrac{1}{n} \sum \limits_{i=1}^{n} \biggr(\frac{1}{2}\|\x_{i} - \h(\widetilde{\y}_{i},\widetilde{\z}_{i}; 0)\|^{2} -\eta(\widetilde{\y}_{i},\widetilde{\z}_{i}) - \log \pi_{\overline{\y}_{i}} \biggr)  \nonumber \\
& & \hspace{ 12 em} + \log \biggr(\sum \limits_{(\y',\z') \in \Jcal} \exp(\eta(\y',z')+\log \pi_{\y'}) \biggr)\biggr\}. \nonumber
\end{eqnarray}
It is clear that
\begin{eqnarray}
\sum \limits_{(\y',\z') \in \Jcal} \exp(\eta(\y',z')+\log \pi_{\y'}) \leq |\mathcal{L}| \sum \limits_{y'} \exp(\max \limits_{\z' \in \mathcal{L}} \eta(\y',z') + \log \pi_{\y'}). \nonumber
\end{eqnarray} 
Combining this inequality with the inequality of the term $A$ in the above display, we have
\begin{eqnarray}
U_{n} & \leq & \min \limits_{\theta} \biggr\{\dfrac{1}{n} \sum \limits_{i=1}^{n} \biggr(\frac{1}{2}\|\x_{i} - \h(\widetilde{\y}_{i},\widetilde{\z}_{i}; 0)\|^{2} -\eta(\widetilde{\y}_{i},\widetilde{\z}_{i}) - \log \pi_{\widetilde{\y}_{i}} \biggr)  + (\log \pi_{\widetilde{\y}_{i}} - \log \pi_{\overline{\y}_{i}}) \nonumber \\
& + & \log \biggr(\sum \limits_{\y'} \exp(\max \limits_{\z' \in \mathcal{L}} \eta(\y',z') + \log \pi_{\y'})\biggr) + \log |\mathcal{L}| \biggr\} \nonumber \\
& = & \min \limits_{\theta} \dfrac{1}{n} \sum \limits_{i=1}^{n}  \biggr\{\biggr( \dfrac{\| \x_{i} - \h(\widetilde{\y}_{i},\widetilde{\z}_{i}; 0)\|^{2}}{2} - \log(\pi_{\widetilde{\y}_{i},\widetilde{\z}_{i}})\biggr) + \biggr(\log \pi_{\widetilde{\y}_{i}} - \log \pi_{\overline{\y}_{i}}\biggr) \biggr\} + \log |\mathcal{L}|  \nonumber \\
& \leq & \min \limits_{\theta} \dfrac{1}{n} \sum \limits_{i=1}^{n}  \biggr( \dfrac{\| \x_{i} - \h(\widetilde{\y}_{i},\widetilde{\z}_{i}; 0)\|^{2}}{2} - \log(\pi_{\widetilde{\y}_{i},\widetilde{\z}_{i}})\biggr) + \log \biggr(\dfrac{1}{\overline{\gamma}}- 1\biggr) + \log |\mathcal{L}|  \nonumber \\
& = & V_{n} + \log \biggr(\dfrac{1}{\overline{\gamma}}- 1\biggr) + \log |\mathcal{L}| \nonumber
\end{eqnarray}
where the final inequality is due to the fact that $\pi_{\widetilde{\y}_{i}}/\pi_{\overline{\y}_{i}} \leq (1-\overline{\gamma})/\overline{\gamma}$ for all $1 \leq i \leq n$. Therefore, we achieve the conclusion of part (a) of the theorem. \\
(b) Similar to the proof argument with part (a), we have
\begin{eqnarray}
U_{n} & = & \min \limits_{\theta} \biggr\{\dfrac{1}{n} \sum \limits_{i=1}^{n} \biggr(\frac{1}{2}\|\x_{i} - \h(\overline{\y}_{i},\overline{\z}_{i}; 0)\|^{2} -\eta(\overline{\y}_{i},\overline{\z}_{i}) - \log \pi_{\overline{\y}_{i}} \biggr) \nonumber \\
& & \hspace{ 15 em} + \log \biggr(\sum \limits_{(\y',\z') \in \Jcal} \exp(\eta(\y',z')+\log \pi_{\y'}) \biggr)\biggr\} \nonumber \\
& \geq &  \min \limits_{\theta} \biggr\{\dfrac{1}{n} \sum \limits_{i=1}^{n}\biggr(\dfrac{1}{2}\|\x_{i}\|^{2} - \h(\widetilde{\y}_{i},\widetilde{\z}_{i}; 0)^{\top}\x_{i} - \eta(\widetilde{\y}_{i},\widetilde{\z}_{i}) + \|\h(\overline{\y}_{i},\overline{\z}_{i}; 0)\|^{2}/2 - \log \pi_{\overline{\y}_{i}}\biggr) \nonumber \\
& & \hspace{ 15 em} + \log \biggr(\sum \limits_{y'} \exp(\max \limits_{\z' \in \mathcal{L}} \eta(\y',z') + \log \pi_{\y'})\biggr) = B. \nonumber
\end{eqnarray}
Direct computation with $B$ leads to
\begin{eqnarray}
B & = & \min \limits_{\theta} \dfrac{1}{n} \sum \limits_{i=1}^{n}  \biggr\{\biggr( \dfrac{\| \x_{i} - \h(\widetilde{\y}_{i},\widetilde{\z}_{i}; 0)\|^{2}}{2} - \log(\pi_{\widetilde{\y}_{i},\widetilde{\z}_{i}})\biggr) + \biggr(\log \pi_{\overline{\y}_{i}} - \log \pi_{\widetilde{\y}_{i}} \biggr) \nonumber \\
& & \hspace{ 20 em} + \|\h(\overline{\y}_{i},\overline{\z}_{i}; 0 )\|^{2} - \|\h(\widetilde{\y}_{i},\widetilde{\z}_{i}; 0)\|^{2} \biggr\} \nonumber \\
& \geq & \min \limits_{\theta} \dfrac{1}{n} \sum \limits_{i=1}^{n}  \biggr( \dfrac{\| \x_{i} - \h(\widetilde{\y}_{i},\widetilde{\z}_{i}; 0)\|^{2}}{2} - \log(\pi_{\widetilde{\y}_{i},\widetilde{\z}_{i}})\biggr) + \min \limits_{\theta} \dfrac{1}{n} \sum \limits_{i=1}^{n}\biggr(\log \pi_{\overline{\y}_{i}} - \log \pi_{\widetilde{\y}_{i}} \biggr) \nonumber \\
& & \hspace{ 20 em} + \min \limits_{\theta} \dfrac{1}{n} \sum \limits_{i=1}^{n}\biggr(\|\h(\overline{\y}_{i},\overline{\z}_{i}; 0)\|^{2} - \|\h(\widetilde{\y}_{i},\widetilde{\z}_{i}; 0)\|^{2}\biggr) \nonumber \\
& \geq & V_{n} + \log \biggr(\dfrac{\overline{\gamma}}{1-\overline{\gamma}}\biggr) + \min \limits_{\theta} \dfrac{1}{n} \sum \limits_{i=1}^{n}\biggr(\|\h(\overline{\y}_{i},\overline{\z}_{i}; 0)\|^{2} - \|\h(\widetilde{\y}_{i},\widetilde{\z}_{i}; 0)\|^{2}\biggr) \nonumber
\end{eqnarray}
where the final inequality is due to the fact that $\pi_{\widetilde{\y}_{i}}/\pi_{\overline{\y}_{i}} \geq \overline{\gamma}/(1-\overline{\gamma})$ for all $1 \leq i \leq n$. As a consequence, we achieve the conclusion of part (b) of the theorem.
\subsection{Proofs for statistical guarantee and generalization bound for (semi)-supervised learning}
\label{Section:proof_semi_sup}
\paragraph{PROOF OF THEOREM \ref{theorem:consistency_objective_LDAP_cross_entropy}}
The proof of this theorem relies on several results with uniform laws of large numbers. In particular, we will need to demonstrate the following results
\begin{eqnarray}
& & \hspace{-2 em} \sup \limits_{\theta} \biggr|\dfrac{1}{n_{1}}\sum \limits_{i=1}^{n_{1}}{\max \limits_{(\y,\z) \in \mathcal{J}} \biggr(\h^{\top}(\y,\z;0)\x_{i}+\tau(\theta_{\y,\z})\biggr)} \nonumber \\
& & \hspace{10 em} - \int \max \limits_{(\y,\z) \in \mathcal{J}} \biggr(\h^{\top}(\y,\z;0)x+\tau(\theta_{\y,\z})\biggr) dP(x)\biggr| \to 0, \label{eqn:consistency_spldap_kmeans_first} \\
& & \hspace{-2 em} L_{n_{1}} = \sup \limits_{\theta} \biggr|\dfrac{1}{n_{1}}\sum \limits_{i=1}^{n_{1}} \sum \limits_{(\y',\z') \in \mathcal{J}}{1_{\left\{(\y',\z') = \mathop{\arg \max}\limits_{(\y,\z) \in \mathcal{J}} \biggr(\h^{\top}(\y,\z;0)\x_{i}+\tau(\theta_{\y,\z})\biggr) \right\}}\biggr(\dfrac{\|\h(\y',\z';0)\|^{2}}{2} - \log \pi_{\y'} \biggr)}\nonumber \\
& & \hspace{-4.0 em} - \int \biggr(\sum \limits_{(\y',\z') \in \mathcal{J}}{1_{\left\{(\y',\z') = \mathop{\arg \max}\limits_{(\y,\z) \in \mathcal{J}} \biggr(\h^{\top}(\y,\z;0)x +\tau(\theta_{\y,\z})\biggr) \right\}}\biggr(\dfrac{\|\h(\y',\z';0)\|^{2}}{2} - \log \pi_{\y'} \biggr)}\biggr)dP(x) \biggr| \to 0, \label{eqn:consistency_spldap_kmeans_second} \\
& &  \hspace{- 3 em} \sup \limits_{\theta} \biggr|\dfrac{1}{n-n_{1}}\sum \limits_{i=n_{1}+1}^{n}{\max \limits_{\z \in \mathcal{L}} \biggr(\h^{\top}(\y_{i},\z;0)\x_{i}+\tau(\theta_{\y_{i},\z})\biggr)} \nonumber \\
& & \hspace{12 em} - \int \max \limits_{\z \in \mathcal{L}} \biggr(\h^{\top}(\y,\z;0)x+\tau(\theta_{\y,\z})\biggr) dQ(x,c)\biggr| \to 0, \label{eqn:consistency_objective_LDAP_cross_entropy_first} \\
& & \hspace{- 3 em} E_{n}^{(1)} = \sup \limits_{\theta} \biggr|\dfrac{1}{n-n_{1}}\sum \limits_{i=n_{1}+1}^{n} \sum \limits_{\z' \in \mathcal{L}}{1_{\left\{\z' = \mathop{\arg \max}\limits_{\z \in \mathcal{L}} \biggr(\h^{\top}(\y_{i},\z;0)\x_{i}+\tau(\theta_{\y_{i},\z})\biggr) \right\}}\biggr(\dfrac{\|\h(\y_{i},\z';0)\|^{2}}{2} - \log \pi_{\y_{i}} \biggr)}\nonumber \\
& & \hspace{- 3 em} - \int \biggr(\sum \limits_{\z' \in \mathcal{L}}{1_{\left\{\z' = \mathop{\arg \max}\limits_{g \in \mathcal{J}} \biggr(\h^{\top}(\y,\z;0)x +\tau(\theta_{\y,\z})\biggr) \right\}}\biggr(\dfrac{\|\h(\y,\z';0)\|^{2}}{2} - \log \pi_{\y} \biggr)}\biggr)dQ(x,c) \biggr| \to 0, \label{eqn:consistency_objective_LDAP_cross_entropy_second}
\end{eqnarray}
\begin{eqnarray}
& & \hspace{- 3 em} E_{n}^{(2)} = \sup \limits_{\theta} \biggr|\dfrac{1}{n-n_{1}} \sum \limits_{i=n_{1}+1}^{n} \log q_{\theta}(\y_{i}|\x_{i}) - \int \log q_{\theta}(\y|x)dQ(x,c) \biggr| \to 0, \label{eqn:consistency_objective_LDAP_cross_entropy_third} \\
& & \hspace{- 3 em} E_{n}^{(3)} = \sup \limits_{\theta} \biggr|\dfrac{1}{n}\sum \limits_{i=1}^{n} \sum \limits_{\y=1}^{K} q_{\theta}(\y|\x_{i}) \log\biggr(\dfrac{q_{\theta}(\y|\x_{i})}{\pi_{\y}}\biggr) \nonumber \\
& & \hspace{12 em} - \int \sum \limits_{\y=1}^{K} q_{\theta}(\y|x)dQ(x,c) \log \biggr(\dfrac{q_{\theta}(\y|x)}{\pi_{\y}}\biggr)dP(x) \biggr| \to 0, \label{eqn:consistency_objective_LDAP_cross_entropy_fourth}
\end{eqnarray}
almost surely as $n \to \infty$. The proof for \eqref{eqn:consistency_objective_LDAP_cross_entropy_first} is similar to that of \eqref{eqn:consistency_spldap_kmeans_first}; therefore, it is omitted.
\paragraph{Proof of~\eqref{eqn:consistency_spldap_kmeans_first}:} It is clear that
\begin{eqnarray}
& & \hspace{-2 em}  \sup \limits_{\theta} \biggr|\dfrac{1}{n_{1}}\sum \limits_{i=1}^{n_{1}}{\max \limits_{(\y,\z) \in \mathcal{J}} \biggr(\h^{\top}(\y,\z;0)\x_{i}+\tau(\theta_{\y,\z})\biggr)} - \int \max \limits_{(\y,\z) \in \mathcal{J}} \biggr(\h^{\top}(\y,\z;0)x+\tau(\theta_{\y,\z})\biggr) dP(x)\biggr| \nonumber \\
& & \hspace{10 em} \leq \sup \limits_{|S'| \leq |\Jcal|}\biggr|\dfrac{1}{n_{1}}\sum \limits_{i=1}^{n_{1}} \max \limits_{s \in S'} s^{\top}[\x_{i},1] - \int \max \limits_{s \in S'} s^{\top}[x,1]dP(x) \biggr| \nonumber
\end{eqnarray}
where $[x,1] \in \mathbb{R}^{D^{(0)}+1}$ denotes the vector forms by concatenating 1 to $x \in \mathbb{R}^{D^{(0)}}$ and $S'$ in the above supremum is the set of finite elements in $\mathbb{R}^{D^{(0)}+1}$. Therefore, to achieve the result of \eqref{eqn:consistency_spldap_kmeans_first}, it is sufficient to show that
\begin{eqnarray}
\sup \limits_{|S'| \leq |\Jcal|}\biggr|\dfrac{1}{n_{1}}\sum \limits_{i=1}^{n_{1}} \max \limits_{s \in S'} s^{\top}[\x_{i},1] - \int \max \limits_{s \in S'} s^{\top}[x,1]dP(x) \biggr| \to 0. \label{eqn:consistency_spldap_kmeans_first_first}
\end{eqnarray}
To obtain the conclusion of \eqref{eqn:consistency_spldap_kmeans_first_first}, we utilize the classical result with bracketing entropy to establish the uniform laws of large number (cf. Lemma 3.1 in~\cite{Vandegeer-2000}). In particular, we denote $\mathcal{G}$ to be the family of function on $\mathbb{R}^{D^{(0)}}$ with the form $f_{S'}(x) = \max \limits_{s \in S'} s^{\top}[x,1]$ where $S' \in \mathcal{O}_{|\Jcal|}$, which contains all sets that have at most $|\Jcal|$ elements in $\mathbb{R}^{D^{(0)}+1}$. Due to the assumption with distribution $P$, we can restrict $\mathcal{O}_{|\Jcal|}$ to contain only set $S'$ with elements in $\mathbb{B}(R)$, which is a closed ball of radius $R$ on $\mathbb{R}^{D^{(0)}+1}$. By means of Lemma 2.5 in \cite{Vandegeer-2000}, we can find a finite set $E_{\delta}$ such that each element in $\mathbb{B}(R)$ is within distance $\delta$ to some element of $E_{\delta}$ for all $\delta>0$. We denote $\mathcal{O}_{|\Jcal|}(\delta)$ to be the subset of $\mathcal{O}_{|\Jcal|}$ such that it only contains sets with elements in $E_{\delta}$ for all $\delta >0$. Now, for each set $S'=\left\{s_{1},\ldots,s_{k}\right\} \in \mathcal{O}_{|\Jcal|}$, we can choose corresponding set $\overline{S}=\left\{s_{1}',\ldots,s_{k}'\right\} \in \mathcal{O}_{|\Jcal|}(\delta)$ such that $\|s_{i}-s_{i}'\| \leq \delta$ for all $1 \leq i \leq k$. Now, we denote
\begin{eqnarray}
\overline{f}_{\overline{S}}(x) = \max \limits_{s \in \overline{S}} s^{\top}[x,1]+\delta\|[x,1]\|, \nonumber \\
\underline{f}_{\overline{S}}(x) = \max \limits_{s \in \overline{S}} s^{\top}[x,1] -\delta \|[x,1]\| \nonumber
\end{eqnarray}
for any $\overline{S} \in \mathcal{O}_{|\Jcal|}(\delta)$. It is clear that $\underline{f}_{\overline{S}}(x) \leq f_{S}(x) \leq \overline{f}_{\overline{S}}(x)$ for all $x \in \mathbb{R}^{D^{(0)}}$. Furthermore, we also have that
\begin{eqnarray}
\int (\overline{f}_{\overline{S}}(x) - \underline{f}_{\overline{S}}(x)) dP(x) = 2\delta \int \|[x,1]\| dP(x) \leq 2\delta\biggr(\int \|x\|dP(x) +1\biggr). \nonumber
\end{eqnarray}
For any $\epsilon>0$, by choosing $\delta < \dfrac{\epsilon}{2(\int \|x\| dP(x)+1)}$ then we will have that $\int (\overline{f}_{\overline{S}}(x) - \underline{f}_{\overline{S}}(x)) dP(x) < \epsilon$. It implies that the $\epsilon$-bracketing entropy of $\mathcal{G}$ is finite for the $\mathbb{L}_{1}$ norm with distribution $P$ (for the definition of bracketing entropy, see Definition 2.2 in~\cite{Vandegeer-2000}). According to Lemma 3.1 in~\cite{Vandegeer-2000}, it implies that $\mathcal{G}$ satisfies the uniform law of large numbers, i.e.,~\eqref{eqn:consistency_spldap_kmeans_first_first} holds.  
\paragraph{Proof of~\eqref{eqn:consistency_spldap_kmeans_second}:} To achieve the conclusion of this claim, we will need to rely on the control of Rademacher complexity based on Vapnik-Chervonenkis (VC) dimension. In particular, we firstly demonstrate that
\begin{eqnarray}
\sup \limits_{|S'| = k} \biggr|\dfrac{1}{n_{1}}\sum \limits_{i=1}^{n_{1}}1_{\left\{j = \mathop{\arg \max}\limits_{1 \leq l \leq |S'|} [\x_{i},1]^{\top}s_{l}'\right\}} - \int 1_{\left\{j = \mathop{\arg \max}\limits_{1 \leq l \leq |S'|} [x,1]^{\top}s_{l}' \right\}} dP(x)\biggr| \to 0 \label{eqn:consistency_spldap_kmeans_third}
\end{eqnarray}
almost surely as $n_{1} \to \infty$ for each $1 \leq j \leq k$ and $k \geq 1$ where the supremum is taken with respect to $S'=\left\{s_{1}',\ldots,s_{k}'\right\}$. For each $j$, we denote the Rademacher complexity as follows
\begin{eqnarray}
R_{n_{1}} = \mathbb{E} \sup \limits_{|S'|=k} \biggr|\dfrac{1}{n_{1}}\sum \limits_{i=1}^{n_{1}} \sigma_{i}1_{\left\{j = \mathop{\arg \max}\limits_{1 \leq l \leq |S'|} [\x_{i},1]^{\top}s_{l}' \right\}} \biggr| \nonumber
\end{eqnarray}
where $\sigma_{1},\ldots,\sigma_{n_{1}}$ are i.i.d. Rademacher random variables, i.e., $\mathbb{P}(\sigma_{i}=-1) = \mathbb{P}(\sigma_{i}=1) = 1/2$ for $1 \leq i \leq n_{1}$. Then, for any $n_{1} \geq 1$ and $\delta \geq 0$, according to standard argument with Rademacher complexity \cite{Vershynin_2011},
\begin{eqnarray}
\sup \limits_{|S'| = k} \biggr|\dfrac{1}{n_{1}}\sum \limits_{i=1}^{n_{1}}1_{\left\{j = \mathop{\arg \max}\limits_{1 \leq l \leq |S'|} [\x_{i},1]^{\top}s_{l}'\right\}} - \int 1_{\left\{j = \mathop{\arg \max}\limits_{1 \leq l \leq |S'|} [x,1]^{\top}s_{l}' \right\}} dP(x)\biggr| \leq 2R_{n_{1}} + \delta \nonumber
\end{eqnarray}
with probability at least $1 - 2\exp\biggr(-\dfrac{n_{1}\delta^{2}}{8} \biggr)$. According to Borel-Cantelli's lemma, to achieve \eqref{eqn:consistency_spldap_kmeans_third}, it is sufficient to demonstrate that $R_{n_{1}} \to 0$ as $n_{1} \to \infty$. 

To achieve that result, we will utilize the study of VC dimension with partitions (cf. Section 21.5 in~\cite{Devroye-1996}). In particular, for each set $S' = (s_{1}',\ldots,s_{k}')$, it gives rise to the partition $A_{i} = \left\{x \in \mathbb{R}^{D^{(0)}} : [x,1]^{\top}s_{i}' \geq [x,1]^{\top}s_{l}' \ \forall \ l \in \left\{1,\ldots,k\right\} \right\}$ as $1 \leq i \leq k$. For our purpose with \eqref{eqn:consistency_spldap_kmeans_third}, it is sufficient to consider $\mathcal{P}_{n_{1}} = \left\{A_{j},\bigcup \limits_{i \neq j} A_{i} \right\}$, which is a partition of $\mathbb{B}(R)$, for each set $S'$ with $k$ elements. We denote $\mathcal{F}$ to be the collection of all $\mathcal{P}_{n}$ for all $S'$ with $k$ elements and $\mathcal{B}(\mathcal{P}_{n})$ the collection of all sets obtained from the unions of elements of $\mathcal{P}_{n}$. For each data $(\x_{1},\ldots,\x_{n_{1}})$, we let $N_{\mathcal{F}}(\x_{1},\ldots,\x_{n_{1}})$ the number of different sets in $\left\{(\x_{1},\ldots,\x_{n_{1}}) \cap A: A \in \mathcal{B}(\mathcal{P}_{n_{1}}) \ \text{for} \ \mathcal{P}_{n_{1}} \in \mathcal{F}\right\}$. The shatter coefficient of $\mathcal{F}$ is defined as 
\begin{eqnarray}
\Delta_{n_{1}}(\mathcal{F}) = s(\mathcal{F},n_{1}) = \max \limits_{(\x_{1},\ldots,\x_{n_{1}})} N_{\mathcal{F}}(\x_{1},\ldots,\x_{n_{1}}). \nonumber
\end{eqnarray}
According to Lemma 21.1 in~\cite{Devroye-1996}, $\Delta_{n_{1}}(\mathcal{F}) \leq 4\Delta_{n_{1}}^{*}(\mathcal{F})$ where $\Delta_{n_{1}}^{*}(\mathcal{F})$ is the maximal number of different ways that $n_{1}$ points can be partitioned by members of $\mathcal{F}$. Now, for each element $\mathcal{P}_{n_{1}}=\left\{A_{j},\bigcup \limits_{i \neq j} A_{i} \right\}$ of $\mathcal{F}$, it is clear that the boundaries between $A_{j}$ and $\bigcup \limits_{i \neq j} A_{i}$ are subsets of hyperplanes. From the formulation of $A_{j}$, we have at most $k-1$ boundaries between $A_{j}$ and $\bigcup \limits_{i \neq j} A_{i}$. From the classical result of~\cite{Dudley_1978}, each $n_{1}$ points in $\mathbb{B}(R)$ can be splitted by a hyperplane in at most $n_{1}^{D^{(0)}+1}$ different ways as the VC dimension of the hyperplane is at most $D^{(0)}+1$. As a consequence, we would have $\Delta_{n_{1}}^{*}(\mathcal{F}) \leq n_{1}^{(D^{(0)}+1)(k-1)}$, which leads to $\Delta_{n_{1}}(\mathcal{F}) \leq 4n_{1}^{(D^{(0)}+1)(k-1)}$. 

Going back to our evaluation with Rademacher complexity $R_{n_{1}}$, by means of Massart's lemma, we have that
\begin{eqnarray}
R_{n_{1}} & = & \mathbb{E}\biggr(\mathbb{E}_{\sigma} \sup \limits_{|S'|=k} \biggr|\dfrac{1}{n_{1}}\sum \limits_{i=1}^{n_{1}} \sigma_{i}1_{\left\{j = \mathop{\arg \max}\limits_{1 \leq l \leq |S'|} [\x_{i},1]^{\top}s_{l}' \right\}} \biggr||\x_{1},\ldots,\x_{n_{1}}\biggr) \nonumber \\
& & \hspace{- 2 em} \leq \mathbb{E}\biggr(\sqrt{\dfrac{2\log 2 N_{\mathcal{F}}(\x_{1},\ldots,\x_{n_{1}})}{n_{1}}}\biggr) \leq \sqrt{\dfrac{2 (\log 8+(D^{(0)}+1)(k-1)\log n_{1})}{n_{1}}} \to 0 \label{eqn:consistency_spldap_kmeans_fourth}
\end{eqnarray}
as $n_{1} \to \infty$. Therefore, \eqref{eqn:consistency_spldap_kmeans_third} is proved.
\paragraph{Proof of~\eqref{eqn:consistency_objective_LDAP_cross_entropy_second}:} To achieve the conclusion of this claim, we firstly demonstrate that
\begin{eqnarray}
& & \sup \limits_{|S'| = k} \biggr|\dfrac{1}{n-n_{1}}\sum \limits_{i=n_{1}+1}^{n}1_{\left\{j = \mathop{\arg \max}\limits_{1 \leq l \leq |S'|} [\x_{i},1]^{\top}s_{l}' \right\}}1_{\left\{\y_{i}=l\right\}} \nonumber \\
& & \hspace{ 5 em} - \int 1_{\left\{j = \mathop{\arg \max}\limits_{1 \leq l \leq |S'|} [x,1]^{\top}s_{l}' \right\}}1_{\left\{c=l\right\}}  dQ(x,c)\biggr| \to 0 \label{eqn:consistency_objective_LDAP_cross_entropy_fifth}
\end{eqnarray}
almost surely as $n \to \infty$ for each $1 \leq j \leq k$ and $1 \leq l \leq K$ where $k \geq 1$ and the supremum is taken with respect to $S'=\left\{s_{1}',\ldots,s_{k}'\right\}$. The proof of the above result will rely on VC dimension with Voronoi partitions being established in~\eqref{eqn:consistency_spldap_kmeans_second}. In particular, according to the standard argument with Rademacher complexity, it is sufficient to demonstrate that
\begin{eqnarray}
R_{n}' = \mathbb{E} \sup \limits_{|S'|=k} \biggr|\dfrac{1}{n - n_{1}}\sum \limits_{i=n_{1}+1}^{n} \sigma_{i}1_{\left\{j = \mathop{\arg \max}\limits_{1 \leq l \leq |S'|} [\x_{i},1]^{\top}s_{l}'\right\}}1_{\left\{\y_{i}=l\right\}} \biggr| \to 0. \nonumber
\end{eqnarray}
By means of the inequality with Rademacher complexity in \eqref{eqn:consistency_spldap_kmeans_fourth}, we obtain that
\begin{eqnarray}
R_{n}' & = & \sum \limits_{v=0}^{n-n_{1}}\sum \limits_{\vec{c} \in A_{v}} \mathbb{E}\biggr(\mathbb{E}_{\sigma} \sup \limits_{|S'|=k} \biggr|\dfrac{1}{n-n_{1}}\sum \limits_{i=n_{1}+1}^{n} \sigma_{i}1_{\left\{j = \mathop{\arg \max}\limits_{1 \leq l \leq |S'|} [\x_{i},1]^{\top}s_{l}' \right\}}1_{\left\{\y_{i}=l\right\}} \biggr||\vec{c} \in A_{v}\biggr) \mathbb{P}(\vec{c} \in A_{v}) \nonumber \\
& = & \sum \limits_{v=0}^{n-n_{1}} \mathbb{E} \sup \limits_{|S'|=k} \biggr|\dfrac{1}{n-n_{1}}\sum \limits_{i=n_{1}+1}^{n_{1}+v} \sigma_{i}1_{\left\{j = \mathop{\arg \max}\limits_{1 \leq l \leq |S'|} [\x_{i},1]^{\top}s_{l}' \right\}}\biggr|p_{l}^{v}(1-p_{l})^{n-n_{1}-v} {n-n_{1} \choose v} \nonumber \\
& \leq & \sum \limits_{v=1}^{n-n_{1}}\dfrac{v}{n-n_{1}}\sqrt{\dfrac{2(\log 8+(D^{(0)}+1)(k-1)\log v)}{v}}p_{l}^{v}(1-p_{l})^{n-n_{1}-v} {n-n_{1} \choose v}  \nonumber
\end{eqnarray}
\begin{eqnarray}
& \leq & \sqrt{\dfrac{2\log 8}{n-n_{1}}} \sum \limits_{v=1}^{n-n_{1}}\sqrt{\dfrac{v}{n-n_{1}}}p_{l}^{v}(1-p_{l})^{n-n_{1}-v} {n-n_{1} \choose v} \nonumber \\
& + & \sqrt{\dfrac{2(D^{(0)}+1)(k-1)\log(n-n_{1})}{n-n_{1}}}\sum \limits_{v=1}^{n-n_{1}}\sqrt{\dfrac{v\log v}{(n-n_{1})\log(n-n_{1})}}p_{l}^{v}(1-p_{l})^{n-n_{1}-v} {n-n_{1} \choose v} \nonumber \\
& \leq & \biggr(\sqrt{\dfrac{2\log 8}{n-n_{1}}}+\sqrt{\dfrac{2(D^{(0)}+1)(k-1)\log(n-n_{1})}{n-n_{1}}}\biggr)\sum \limits_{v=1}^{n-n_{1}}p_{l}^{v}(1-p_{l})^{n-n_{1}-v} {n-n_{1} \choose v} \nonumber \\
& = & \biggr(\sqrt{\dfrac{2\log 8}{n-n_{1}}}+\sqrt{\dfrac{2(D^{(0)}+1)(k-1)\log(n-n_{1})}{n-n_{1}}}\biggr) (1 - (1-p_{l})^{n-n_{1}}) \to 0 \nonumber
\end{eqnarray}
as $n \to \infty$ where $\vec{c} = (c_{n-n_{1}+1},\ldots, c_{n})$ and $A_{v}$ is the set of $\vec{c}$ such that there are exactly $v$ values of $\y_{i}$ to be $l$ for $0 \leq v \leq n-n_{1}$. The final inequality is due to the fact that $v/(n-1) \leq 1$ and $v\log v/\left\{(n-1)\log(n-1)\right\} \leq 1$ for all $1 \leq v \leq n-n_{1}$. Therefore, we achieve the conclusion of \eqref{eqn:consistency_objective_LDAP_cross_entropy_fifth}.

Now, coming back to \eqref{eqn:consistency_objective_LDAP_cross_entropy_second}, by means of triangle inequality, we achieve that
\begin{eqnarray}
& & \hspace{ -2 em} E_{n}^{(1)} \leq 
\sum \limits_{l=1}^{K} \sup \limits_{\theta} \biggr|\dfrac{1}{n-n_{1}}\sum \limits_{i=n_{1}+1}^{n} \sum \limits_{\z' \in \mathcal{L}}{1_{\left\{\z' = \mathop{\arg \max}\limits_{\z \in \mathcal{L}} \biggr(\h^{\top}(\y_{i},\z;0)\x_{i}+\tau(\theta_{\y_{i},\z})\biggr) \right\}}1_{\left\{\y_{i}=l\right\}}\biggr(\dfrac{\|\h(\y_{i},\z';0)\|^{2}}{2} - \log \pi_{\y_{i}} \biggr)}\nonumber \\
& & - \int \biggr(\sum \limits_{\z' \in \mathcal{L}}{1_{\left\{\z' = \mathop{\arg \max}\limits_{g \in \mathcal{J}} \biggr(\h^{\top}(\y,\z;0)x +\tau(\theta_{\y,\z})\biggr) \right\}}1_{\left\{\y=l\right\}}\biggr(\dfrac{\|\h(\y,\z';0)\|^{2}}{2} - \log \pi_{\y} \biggr)}\biggr)dQ(x,\y) \biggr| \nonumber \\
& & \leq \sum \limits_{l=1}^{K} \sup \limits_{\theta} \sum \limits_{\z' \in \mathcal{L}} \biggr|\dfrac{\|\h(l,\z';0)\|^{2}}{2} - \log \pi_{l} \biggr| \biggr|\dfrac{1}{n-n_{1}}\sum \limits_{i=n_{1}+1}^{n} 1_{\left\{\z' = \mathop{\arg \max}\limits_{\z \in \mathcal{L}} \biggr(\h^{\top}(\y_{i},\z;0)\x_{i}+\tau(\theta_{\y_{i},\z})\biggr) \right\}}1_{\left\{\y_{i}=l\right\}} \nonumber \\
& & - \int 1_{\left\{\z' = \mathop{\arg \max}\limits_{g \in \mathcal{J}} \biggr(\h^{\top}(\y,\z;0)x +\tau(\theta_{\y,\z})\biggr) \right\}}1_{\left\{\y=l\right\}} dQ(x,\y) \biggr| \to 0 \nonumber
\end{eqnarray}
where the last inequality is due to the results with uniform laws of large numbers from \eqref{eqn:consistency_objective_LDAP_cross_entropy_fifth} and the fact that $\dfrac{\|\h(l,\z';0)\|^{2}}{2} - \log \pi_{l}$ is bounded for all $l$ and $\z' \in \mathcal{L}$. Hence, we obtain the conclusion of \eqref{eqn:consistency_objective_LDAP_cross_entropy_second}. 
\paragraph{Proof of~\eqref{eqn:consistency_objective_LDAP_cross_entropy_third}:} For this claim, we have the following inequality
\begin{eqnarray}
\hspace{- 4 em} E_{n}^{(2)} & \leq &  \sup \limits_{\left\{S_{\y}'\right\},\left\{\pi_{\y}\right\}} \biggr|\dfrac{1}{n-n_{1}} \sum \limits_{i=n_{1}+1}^{n} \log \dfrac{\exp\biggr(\max \limits_{s \in S_{\y_{i}}'} s^{\top}[\x_{i},1]+\log \pi_{\y_{i}}\biggr)}{\sum \limits_{\ell=1}^{K} \exp\biggr(\max \limits_{s \in S_{l}'} s^{\top}[\x_{i},1]+\log \pi_{l}\biggr)} \nonumber \\
& & \hspace{ 4 em} - \int \log \dfrac{\exp\biggr(\max \limits_{s \in S_{\y}'} s^{\top}[x,1]+\log \pi_{\y}\biggr)}{\sum \limits_{\ell=1}^{K} \exp\biggr(\max \limits_{s \in S_{l}'} s^{\top}[x,1]+\log \pi_{l}\biggr)} dQ(x,\y)\biggr| \nonumber
\end{eqnarray}
\begin{eqnarray}
& & \leq \sum \limits_{\ell=1}^{K} \sup \limits_{\left\{S_{\y}'\right\},\left\{\pi_{\y}\right\}} \biggr|\dfrac{1}{n-n_{1}} \sum \limits_{i=n_{1}+1}^{n} \log \dfrac{\exp\biggr(\max \limits_{s \in S_{\y_{i}}'} s^{\top}[\x_{i},1]+\log \pi_{\y_{i}}\biggr)}{\sum \limits_{\ell=1}^{K} \exp\biggr(\max \limits_{s \in S_{l}'} s^{\top}[\x_{i},1]+\log \pi_{l}\biggr)}1_{\left\{\y_{i}=l\right\}} \nonumber \\
& & \hspace{ 4 em} - \int \log \dfrac{\exp\biggr(\max \limits_{s \in S_{\y}'} s^{\top}[x,1]+\log \pi_{\y}\biggr)}{\sum \limits_{\ell=1}^{K} \exp\biggr(\max \limits_{s \in S_{l}'} s^{\top}[x,1]+\log \pi_{l}\biggr)}1_{\left\{c=l\right\}} dQ(x,\y)\biggr| = \sum \limits_{\ell=1}^{K} F_{n,l} \nonumber
\end{eqnarray}
where $\left\{S_{\y}'\right\}$ in the above supremum stands for the collection of sets $S_{1}',\ldots,S_{K}'$ such that $|S_{\y}'| \leq |\mathcal{L}|$ and elements in $S_{\y}'$ are in $\mathbb{R}^{D^{(0)}+1}$. Therefore, to achieve the conclusion of \eqref{eqn:consistency_objective_LDAP_cross_entropy_third}, it is sufficient to demonstrate that $F_{n,l} \to 0$ almost surely as $n \to \infty$ for each $1 \leq l \leq K$.

In fact, for each $1 \leq l \leq K$, we denote $\mathcal{G}$ to be the family of function on $\mathbb{R}^{D^{(0)}} \times \left\{1,\ldots,K\right\}$ of the form 
\begin{eqnarray}
f_{\left\{S_{\y}'\right\},\left\{\pi_{\y}\right\}}(\x,\y) = \log \dfrac{\exp\biggr(\max \limits_{s \in S_{\y}'} s^{\top}[x,1]+\log \pi_{\y}\biggr)}{\sum \limits_{\ell=1}^{K} \exp\biggr(\max \limits_{s \in S_{l}'} s^{\top}[x,1]+\log \pi_{l}\biggr)}1_{\left\{\y=l\right\}} \nonumber
\end{eqnarray}
for all $(\x,\y)$ where $S_{1}',\ldots,S_{K}' \in \mathcal{O}_{|\mathcal{L}|}$, which contains all sets that have at most $|\mathcal{L}|$ elements in $\mathbb{R}^{D^{(0)}+1}$, and $\left\{\pi_{\y}\right\}$ satisfy that $\sum \limits_{\y=1}^{K} \pi_{\y}=1$ and $\pi_{\y} \geq \overline{\gamma}$ for all $1 \leq \y \leq K$. Due to the assumption with distribution $P$, we can restrict $\mathcal{O}_{|\mathcal{L}|}$ to contain only set $S'$ with elements in the ball $\mathbb{B}(R)$ of radius $R$ on $\mathbb{R}^{D^{(0)}+1}$. By means of Lemma 2.5 in \cite{Vandegeer-2000}, we can find a finite set $E_{\delta}$ such that each element in $\mathbb{B}(R)$ is within distance $\delta$ to some element of $E_{\delta}$ for all $\delta>0$. Additionally, there exists a set $\Delta(\delta)$ such that for each $(\pi_{1},\ldots,\pi_{K})$, we can find a corresponding element $(\pi_{1}',\ldots,\pi_{K}') \in \Delta(\delta)$ such that $\|(\pi_{1},\ldots,\pi_{K}) - (\pi_{1}',\ldots,\pi_{K}')\| \leq \delta$. We denote 
\begin{eqnarray}
\mathcal{F}(\delta) = \biggr\{\left\{S_{\y}'\right\},\left\{\pi_{\y}\right\}: \ \text{elements of} \ S_{\y}' \ \text{in} \ E_{\delta}, \ \text{and} (\pi_{1},\ldots,\pi_{K}) \in \Delta(\delta)\biggr\} \nonumber
\end{eqnarray}
for all $\delta>0$. 

For each element $\left\{S_{\y}'\right\}_{\y=1}^{K}, \left\{\pi_{\y}\right\}_{\y=1}^{K}$, we can choose the corresponding element $\left\{\overline{S}_{\y}'\right\}_{\y=1}^{K}, \left\{\overline{\pi}_{\y}\right\}_{\y=1}^{K} \in \mathcal{F}(\delta)$ such that $S_{\y}' = \left\{s_{\y 1},\ldots,s_{\y k_{\y}}\right\}$, $\overline{S}_{\y}' = \left\{s_{\y 1}',\ldots,s_{\y k_{\y}}'\right\}$ satisfy $\|s_{\y j}-s_{\y j}'\| \leq \delta$ for all $1 \leq \y \leq K$ and $1 \leq j \leq k_{\y}$. Additionally, $\|(\pi_{1},\ldots,\pi_{K}) - (\overline{\pi}_{1},\ldots,\overline{\pi}_{K})\| \leq \delta$. With these notations, we define
\begin{eqnarray}
& & \hspace{- 2 em} \overline{f}_{\left\{\overline{S}_{\y}'\right\},\left\{\overline{\pi}_{\y}\right\}}(x,\y) = \log \biggr(\dfrac{\exp(\mathop {\max} \limits_{s \in \overline{S}_{\y}'}s^{\top}[x,1] + \log \overline{\pi}_{\y})}{\sum \limits_{\tau =1}^{K} \exp(\mathop {\max} \limits_{s \in \overline{S}_{\tau}'}s^{\top}[x,1] + \log \overline{\pi}_{\tau})}\biggr)1_{\left\{\y=l\right\}} + 2\delta\|[x,1]\|+2\delta/\overline{\gamma}, \nonumber \\
& & \hspace{- 2 em} \underline{f}_{\left\{\overline{S}_{\y}'\right\},\left\{\overline{\pi}_{\y}\right\}}(x,\y) = \log \biggr(\dfrac{\exp(\mathop {\max} \limits_{s \in \overline{S}_{\y}'}s^{\top}[x,1] + \log \overline{\pi}_{\y})}{\sum \limits_{\tau =1}^{K} \exp(\mathop {\max} \limits_{s \in \overline{S}_{\tau}'}s^{\top}[x,1] + \log \overline{\pi}_{\tau})}\biggr)1_{\left\{\y=l\right\}} - 2\delta\|[x,1]\| - 2\delta/\overline{\gamma} \nonumber
\end{eqnarray}
for any $\left(\left\{\overline{S}_{\y}'\right\},\left\{\overline{\pi}_{\y}\right\}\right) \in \mathcal{F}(\delta)$. By means of Cauchy-Schwarz's inequality, we have
\begin{eqnarray}
s_{\y i}^{\top}[x,1]1_{\left\{\y=l\right\}}-(s_{\y i}')^{\top}[x,1]1_{\left\{\y=l\right\}} & \leq & \|s_{ci}-s_{\y i}'\|\|[x,1]\|1_{\left\{\y=l\right\}} \nonumber \\
& \leq & \delta \|[x,1]\| \nonumber
\end{eqnarray}
for all $x$ and $1 \leq i \leq k$. Additionally, the following also holds
\begin{eqnarray}
s_{\y i}^{\top}[x,1]1_{\left\{\y=l\right\}}-(s_{\y i}')^{\top}[x,1]1_{\left\{\y=l\right\}} \geq -\delta \|[x,1]\|. \nonumber
\end{eqnarray}
Furthermore, $|\log \pi_{\y} - \log \overline{\pi}_{\y}| \leq \log(1+\delta/\overline{\gamma}) \leq \delta/\overline{\gamma}$. Hence, we obtain that
\begin{eqnarray}
\dfrac{\exp(\mathop {\max} \limits_{s \in S_{\y}'}s^{\top}[x,1] + \log \pi_{\y})}{\sum \limits_{\tau =1}^{K} \exp(\mathop {\max} \limits_{s \in S_{\tau}'}s^{\top}[x,1] + \log \pi_{\tau})} &\leq & \dfrac{\exp(\mathop {\max} \limits_{s \in \overline{S}_{\y}'}s^{\top}[x,1] + \log \overline{\pi}_{\y} + \delta\|[x,1]\|+\delta/\overline{\gamma})}{\sum \limits_{\tau=1}^{K} \exp(\mathop {\max} \limits_{s \in \overline{S}_{\tau}'}s^{\top}[x,1] + \log \overline{\pi}_{\tau} - \delta\|[x,1]\| - \delta/\overline{\gamma})} \nonumber \\
& \leq & \dfrac{\exp(\mathop {\max} \limits_{s \in \overline{S}_{\y}'}s^{\top}[x,1] + \log \overline{\pi}_{\y})}{\sum \limits_{\tau =1}^{K} \exp(\mathop {\max} \limits_{s \in \overline{S}_{\tau}'}s^{\top}[x,1] + \log \overline{\pi}_{\tau})}\exp(2\delta\|[x,1]\|+2\delta/\overline{\gamma}). \nonumber
\end{eqnarray}
Similarly, we also have
\begin{eqnarray}
\dfrac{\exp(\mathop {\max} \limits_{s \in S_{\y}'}s^{\top}[x,1] + \log \pi_{\y})}{\sum \limits_{\tau =1}^{K} \exp(\mathop {\max} \limits_{s \in S_{\tau}'}s^{\top}[x,1] + \log \pi_{\tau})} \geq \dfrac{\exp(\mathop {\max} \limits_{s \in \overline{S}_{\y}'}s^{\top}[x,1] + \log \overline{\pi}_{\y})}{\sum \limits_{\tau =1}^{K} \exp(\mathop {\max} \limits_{s \in \overline{S}_{\tau}'}s^{\top}[x,1] + \log \overline{\pi}_{\tau})}\exp(-2\delta\|[x,1]\|-2\delta/\overline{\gamma}). \nonumber
\end{eqnarray}
As a consequence, we achieve that
\begin{eqnarray}
\underline{f}_{\left\{\overline{S}_{\y}'\right\},\left\{\overline{\pi}_{\y}\right\}}(x,\y) \leq f_{\left\{S_{\y}'\right\},\left\{\pi_{\y}\right\}}(x,\y)  \leq \overline{f}_{\left\{\overline{S}_{\y}'\right\},\left\{\overline{\pi}_{\y}\right\}}(x,\y) \nonumber
\end{eqnarray}
for all $(x,\y)$. With the formulations of $\underline{f}_{\left\{\overline{S}_{\y}'\right\},\left\{\overline{\pi}_{\y}\right\}}(x,c)$ and $\overline{f}_{\left\{\overline{S}_{\y}'\right\},\left\{\overline{\pi}_{\y}\right\}}(x,\y)$, we have
\begin{eqnarray}
& & \hspace{- 3em} \int \biggr(\overline{f}_{\left\{\overline{S}_{\y}'\right\},\left\{\overline{\pi}_{\y}\right\}}(x,\y) - \underline{f}_{\left\{\overline{S}_{\y}'\right\},\left\{\overline{\pi}_{\y}\right\}}(x,\y)\biggr) dQ(x,\y) \nonumber \\
& & = 4\delta \int \|[x,1]\| dQ(x,\y)+4\delta/\overline{\gamma} \leq 4\delta\biggr(\int \|x\| dQ(x,\y)+1\biggr)+4\delta/\overline{\gamma}. \nonumber
\end{eqnarray}
For any $\epsilon>0$, by choosing $\delta < \dfrac{\epsilon}{4\int \|x\| dQ(x,\y)+4+4/\overline{\gamma}}$ then we will have $\int (\overline{f}_{\left\{\overline{S}_{\y}'\right\},\left\{\overline{\pi}_{\y}\right\}}(x,\y) - \underline{f}_{\left\{\overline{S}_{\y}'\right\},\left\{\overline{\pi}_{\y}\right\}}(x,\y)) dQ(x,\y) < \epsilon$. It implies that the $\epsilon$-bracketing entropy of $\mathcal{G}$ is finite for the $L_{1}$ norm with distribution $Q$. Therefore, it implies that $\mathcal{G}$ satisfies the uniform law of large numbers, i.e., $F_{n,l} \to 0$ almost surely as $n \to \infty$ for all $1 \leq l \leq K$. As a consequence, the uniform law of large number result \eqref{eqn:consistency_objective_LDAP_cross_entropy_third} holds.

Going back to the original problem, denote $\widetilde{\theta}^{0} = \biggr(\left\{\widetilde{\muy}^{0}(\y)\right\}_{\y=1}^{K},\left\{\widetilde{\template}_{0}(\ell)\right\}_{\ell=1}^{L},\left\{\widetilde{\pi}_{\y}^{0}\right\}_{\y=1}^{K},\left\{\widetilde{\bias}_{0}(\ell)\right\}_{\ell=1}^{L} \biggr)$ the optimal solutions of population partially labeled LDCE (Note that, the existence of these optimal solutions is guaranteed due to the compact assumptions with the parameter spaces $\Theta_{\ell}$, $\Omega$, and $\Xi_{\ell}$ for all $1 \leq  \ell \leq L$). Then, according to the formulation of partially labeled LDCE, we will have that
\begin{eqnarray}
Y_{n} & \leq & \min \limits_{\theta} \dfrac{\alpha_{\text{RC}}}{n} \biggr\{\sum \limits_{i=1}^{n_{1}} \sum \limits_{(\y',\z') \in \mathcal{J}}1_{\left\{(\y',\z') = \mathop{\arg \max}\limits_{(\y,\z) \in \mathcal{J}} \biggr(\x_{i}^{\top}\widetilde{\h}^{0}(\y,\z;0)+\tau(\widetilde{\theta}^{0}_{\y,\z})\biggr) \right\}} \biggr( \dfrac{\| \x_{i} - \widetilde{\h}^{0}(\y',\z';0)\|^{2}}{2} \nonumber \\
& & \hspace{- 2 em} - \log(\widetilde{p}^{0}(\y',\z'))\biggr) + \sum \limits_{i=n_{1}+1}^{n} \sum \limits_{\z' \in \mathcal{L}}1_{\left\{\z' = \mathop{\arg \max}\limits_{\z \in \mathcal{L}} \biggr(\x_{i}^{\top}\widetilde{\h}^{0}(\y,\z;0)+\tau(\widetilde{\theta}^{0}_{\y,\z})\biggr) \right\}} \biggr( \dfrac{\| \x_{i} - \widetilde{\h}^{0}(\y_{i},\z';0)\|^{2}}{2} \nonumber \\
& & \hspace{- 2 em} - \log(\widetilde{p}^{0}(\y_{i},\z'))\biggr)\biggr\} - \dfrac{\alpha_{\text{CE}}}{n-n_{1}} \sum \limits_{i=n_{1}+1}^{n} \log q_{\widetilde{\theta}^{0}}(\y_{i}|\x_{i})  + \dfrac{\alpha_{\text{KL}}}{n} \sum \limits_{i=1}^{n} \sum \limits_{\y=1}^{K} q_{\widetilde{\theta}^{0}}(\y|\x_{i}) \log\biggr(\dfrac{q_{\widetilde{\theta}^{0}}(\y|\x_{i})}{\pi_{\y}}\biggr) = D_{n} \nonumber
\end{eqnarray}
 for all $n \geq 1$ where we have the following formulations
\begin{align}
& \widetilde{\h}^{0}(\y,\z;0)  =  \widetilde{\Lambda}_{0}(\z;1)\ldots\widetilde{\Lambda}_{0}(\z;L)\widetilde{\muy}^{0}(\y), \nonumber \\
& \widetilde{\Lambda}_{0}(\z;\ell)  = \sum_{\p\in \pp(\ell)}  \sparse(\ell, \p)\ttrans(t;\ell,\p) \pad(\ell, \p)\tilde{\template}_{0}(\ell, \p), \nonumber \\
& \widetilde{p}^{0}(\y',\z')  =  \exp\biggr(\tau(\widetilde{\theta}^{0}_{\y',\z'})+\log \widetilde{\pi}_{\y'}^{0} \biggr)/\biggr(\sum \limits_{\y=1}^{K}\exp\biggr(\max \limits_{\z \in \mathcal{L}} \tau(\widetilde{\theta}^{0}_{\y,\z})+\log \widetilde{\pi}_{\y}^{0} \biggr)\biggr). \nonumber
\end{align}
From the results with uniform laws of large numbers in \eqref{eqn:consistency_spldap_kmeans_first}, \eqref{eqn:consistency_spldap_kmeans_second}, \eqref{eqn:consistency_objective_LDAP_cross_entropy_first}, \eqref{eqn:consistency_objective_LDAP_cross_entropy_second}, \eqref{eqn:consistency_objective_LDAP_cross_entropy_third}, and \eqref{eqn:consistency_objective_LDAP_cross_entropy_fourth}, we obtain that
\begin{eqnarray}
& & \hspace{ - 2 em} \dfrac{1}{n_{1}} \sum \limits_{i=1}^{n_{1}} \sum \limits_{(\y',\z') \in \mathcal{J}}1_{\left\{(\y',\z') = \mathop{\arg \max}\limits_{(\y,\z) \in \mathcal{J}} \biggr(\x_{i}^{\top}\widetilde{\h}^{0}(\y,\z;0)+\tau(\widetilde{\theta}^{0}_{\y,\z})\biggr) \right\}} \biggr( \dfrac{\| \x_{i} - \widetilde{\h}^{0}(\y',\z';0)\|^{2}}{2} - \log(\widetilde{p}^{0}(\y',\z'))\biggr) \nonumber \\
& & \hspace{ - 3 em} \to \int \sum \limits_{(\y',\z') \in \mathcal{J}}1_{\left\{(\y',\z') = \mathop{\arg \max}\limits_{(\y,\z) \in \mathcal{J}} \biggr(x^{\top}\widetilde{\h}^{0}(\y,\z;0)+\tau(\widetilde{\theta}^{0}_{\y,\z})\biggr)\right\}} \biggr( \dfrac{\| x - \widetilde{\h}^{0}(\y',\z';0)\|^{2}}{2} - \log(\widetilde{p}^{0}(\y',\z'))\biggr)dP(x), \nonumber \\
& & \hspace{- 2 em} \dfrac{1}{n-n_{1}} \sum \limits_{i=n_{1}+1}^{n} \sum \limits_{\z' \in \mathcal{L}}1_{\left\{\z' = \mathop{\arg \max}\limits_{\z \in \mathcal{L}} \biggr(\x_{i}^{\top}\widetilde{\h}^{0}(\y,\z;0)+\tau(\widetilde{\theta}^{0}_{\y,\z})\biggr) \right\}} \biggr( \dfrac{\| \x_{i} - \widetilde{\h}^{0}(\y_{i},\z';0)\|^{2}}{2} - \log(\widetilde{p}^{0}(\y_{i},\z'))\biggr) \nonumber \\
& & \hspace{ - 3 em} \to \int \sum \limits_{\z' \in \mathcal{L}}1_{\left\{\z' = \mathop{\arg \max}\limits_{\z \in \mathcal{L}} \biggr(x^{\top}\widetilde{\h}^{0}(\y,\z;0)+\tau(\widetilde{\theta}^{0}_{\y,\z})\biggr)\right\}} \biggr( \dfrac{\| x - \widetilde{\h}^{0}(\y,\z';0)\|^{2}}{2} - \log(\widetilde{p}^{0}(\y,\z'))\biggr)dQ(x,\y), \nonumber \\
& & \dfrac{1}{n-n_{1}} \sum \limits_{i=n_{1}+1}^{n} \log q_{\widetilde{\theta}^{0}}(\y_{i}|\x_{i}) \to \int \log q_{\widetilde{\theta}^{0}}(\y|x)dQ(x,\y), \nonumber \\
& &  \dfrac{1}{n} \sum \limits_{i=1}^{n} \sum \limits_{\y=1}^{K} q_{\widetilde{\theta}^{0}}(\y|\x_{i}) \log\biggr(\dfrac{q_{\widetilde{\theta}^{0}}(\y|\x_{i})}{\pi_{\y}}\biggr) \to \int \sum \limits_{\y=1}^{K} q_{\widetilde{\theta}^{0}}(\y|x) \log \biggr(\dfrac{q_{\widetilde{\theta}^{0}}(\y|x)}{\pi_{\y}}\biggr)dP(x) \nonumber 
\end{eqnarray}
almost surely as $n \to \infty$. Combining with the fact that $n_{1}/n \to \overline{\lambda}$, the above results lead to $D_{n} \to \overline{Y}$ almost surely as $n \to \infty$. Therefore, we have $\lim \limits_{n \to \infty} Y_{n} \leq \overline{Y}$ almost surely as $n \to \infty$. On the other hand, with the results of uniform laws of large numbers in \eqref{eqn:consistency_spldap_kmeans_first}, \eqref{eqn:consistency_spldap_kmeans_second}, \eqref{eqn:consistency_objective_LDAP_cross_entropy_first}, \eqref{eqn:consistency_objective_LDAP_cross_entropy_second}, \eqref{eqn:consistency_objective_LDAP_cross_entropy_third}, and \eqref{eqn:consistency_objective_LDAP_cross_entropy_fourth}, we also have that
\begin{eqnarray}
& & \hspace{-2 em} \dfrac{1}{n_{1}} \sum \limits_{i=1}^{n_{1}} \sum \limits_{(\y',\z') \in \mathcal{J}}1_{\left\{(\y',\z') = \mathop{\arg \max}\limits_{(\y,\z) \in \mathcal{J}} \biggr(\x_{i}^{\top}\widetilde{\h}(\y,\z;0)+\tau(\widetilde{\theta}_{\y,\z})\biggr) \right\}} \biggr( \dfrac{\| \x_{i} - \widetilde{\h}(\y',\z';0)\|^{2}}{2} - \log(\widetilde{p}(\y',\z'))\biggr) \nonumber \\
& & \hspace{-2 em} \to \int \sum \limits_{(\y',\z') \in \mathcal{J}}1_{\left\{(\y',\z') = \mathop{\arg \max}\limits_{(\y,\z) \in \mathcal{J}} \biggr(x^{\top}\widetilde{\h}(\y,\z;0)+\tau(\widetilde{\theta}_{\y,\z})\biggr)\right\}} \biggr( \dfrac{\| x - \widetilde{\h}(\y',\z';0)\|^{2}}{2} - \log(\widetilde{p}(\y',\z'))\biggr)dP(x), \nonumber \\
& & \hspace{-2 em} \dfrac{1}{n-n_{1}} \sum \limits_{i=n_{1}+1}^{n} \sum \limits_{\z' \in \mathcal{L}}1_{\left\{\z' = \mathop{\arg \max}\limits_{\z \in \mathcal{L}} \biggr(\x_{i}^{\top}\widetilde{\h}(\y,\z;0)+\tau(\widetilde{\theta}_{\y,\z})\biggr) \right\}} \biggr( \dfrac{\| \x_{i} - \widetilde{\h}(\y_{i},\z';0)\|^{2}}{2} - \log(\widetilde{p}(\y_{i},\z'))\biggr) \nonumber \\
& & \hspace{-2 em} \to \int \sum \limits_{\z' \in \mathcal{L}}1_{\left\{\z' = \mathop{\arg \max}\limits_{\z \in \mathcal{L}} \biggr(x^{\top}\widetilde{\h}^{0}(\y,\z;0)+\tau(\widetilde{\theta}_{\y,\z})\biggr)\right\}} \biggr( \dfrac{\| x - \widetilde{\h}(\y,\z';0)\|^{2}}{2} - \log(\widetilde{p}(\y,\z'))\biggr)dQ(x,\y), \nonumber \\
& & \hspace{-2 em} \dfrac{1}{n-n_{1}} \sum \limits_{i=n_{1}+1}^{n} \log q_{\widetilde{\theta}}(\y_{i}|\x_{i}) \to \int \log q_{\widetilde{\theta}}(\y|x)dQ(x,\y), \nonumber \\
& & \hspace{-2 em} \dfrac{1}{n} \sum \limits_{i=1}^{n} \sum \limits_{\y=1}^{K} q_{\widetilde{\theta}}(\y|\x_{i}) \log\biggr(\dfrac{q_{\widetilde{\theta}}(\y|\x_{i})}{\pi_{\y}}\biggr) \to \int \sum \limits_{\y=1}^{K} q_{\widetilde{\theta}}(\y|x) \log \biggr(\dfrac{q_{\widetilde{\theta}}(\y|x)}{\pi_{\y}}\biggr)dP(x) \nonumber
\end{eqnarray}
almost surely as $n \to \infty$ where $\widetilde{\theta} = \biggr(\left\{\widetilde{\muy}(\y)\right\}_{\y=1}^{K},\left\{\widetilde{\template}(\ell)\right\}_{\ell=1}^{L},\left\{\widetilde{\pi}_{\y}\right\}_{\y=1}^{K},\left\{\widetilde{\bias}(\ell)\right\}_{\ell=1}^{L} \biggr)$ is the optimal solution of partially labeled LDCE and
\begin{align} 
\widetilde{\h}(\y,\z;0) & = \widetilde{\Lambda}(\z;1)\ldots\widetilde{\Lambda}(\z;L)\widetilde{\muy}(\y), \nonumber \\
\widetilde{\Lambda}(\z;\ell) & = \sum_{\p\in \pp(\ell)}  \sparse(\ell, \p)\ttrans(t;\ell,\p) \pad(\ell, \p)\tilde{\template}(\ell, \p), \nonumber \\
\widetilde{p}(\y',\z') & = \exp\biggr(\tau(\widetilde{\theta}_{\y',\z'})+\log \widetilde{\pi}_{\y'} \biggr)/\biggr(\sum \limits_{\y=1}^{K}\exp\biggr(\max \limits_{\z \in \mathcal{L}} \tau(\widetilde{\theta}_{\y',\z'})+\log \widetilde{\pi}_{\y} \biggr)\biggr). \nonumber
\end{align} 
Hence, we eventually achieve that
\begin{eqnarray}
& & \hspace {- 2 em} Y_{n}  \to  \alpha_{\text{RC}}\biggr\{\overline{\lambda}\biggr(\int \sum \limits_{(\y',\z') \in \mathcal{J}}1_{\left\{(\y',\z') = \mathop{\arg \max}\limits_{(\y,\z) \in \mathcal{J}} \biggr(x^{\top}\widetilde{\h}(\y,\z;0)+\tau(\widetilde{\theta}_{\y,\z})\biggr)\right\}} \biggr( \dfrac{\| x - \widetilde{\h}(\y',\z';0)\|^{2}}{2} \nonumber \\
& & \hspace {- 2 em} - \log(\widetilde{p}(\y',\z'))\biggr)dP(x)\biggr) + (1-\overline{\lambda}) \biggr(\int \sum \limits_{\z' \in \mathcal{L}}1_{\left\{\z' = \mathop{\arg \max}\limits_{\z \in \mathcal{L}} \biggr(x^{\top}\widetilde{\h}^{0}(\y,\z;0)+\tau(\widetilde{\theta}_{\y,\z})\biggr)\right\}} \biggr( \dfrac{\| x - \widetilde{\h}(\y,\z';0)\|^{2}}{2} \nonumber \\
& & \hspace {- 2 em} - \log(\widetilde{p}(\y,\z'))\biggr)dQ(x,\y)\biggr)\biggr\} - \alpha_{\text{CE}} \int \log q_{\widetilde{\theta}}(\y|x)dQ(x,\y) \nonumber \\
& & \hspace {18 em} + \alpha_{\text{KL}} \int \sum \limits_{\y=1}^{K} q_{\widetilde{\theta}}(\y|x) \log \biggr(\dfrac{q_{\widetilde{\theta}}(\y|x)}{\pi_{\y}}\biggr)dP(x) \geq \overline{Y} \nonumber
\end{eqnarray}
almost surely as $n \to \infty$. As a consequence, we achieve $Y_{n} \to \overline{Y}$ almost surely as $n \to \infty$. We reach the conclusion of the theorem. 
\paragraph{PROOF OF THEOREM \ref{theorem:consistency_optimal_solutions_LDCE}}
The proof argument of this theorem is a direct application of the results with uniform laws of large numbers in the proof of Theorem \ref{theorem:consistency_objective_LDAP_cross_entropy}. In fact, we define
\begin{eqnarray}
& & \widetilde{\mathcal{F}}_{0}(\epsilon) = \biggr\{\left(S,\left\{\pi_{\y}\right\},\left\{\bias(\ell)\right\}\right): S=\biggr\{\h(\y,\z;0): \ (\y,\z) \in \Jcal \biggr\}, \nonumber \\
& &  \text{and} \inf \limits_{\left(\widetilde{S}_{0}, \left\{\widetilde{\pi}_{\y}^{0}\right\},\left\{\widetilde{\bias}_{0}(\ell)\right\}\right) \in \Gcal(\widetilde{\mathcal{F}}_{0})} \biggr\{H(S,\widetilde{S}_{0}) + \sum \limits_{\y=1}^{K} |\pi_{\y} - \widetilde{\pi}_{\y}^{0}| + \sum \limits_{\z \in \mathcal{L}} \|\bias(\ell) - \widetilde{\bias}_{0}(\ell)\|\biggr\} \geq \epsilon \biggr\} \nonumber
\end{eqnarray}
for any $\epsilon>0$. Since the parameter spaces of $\theta$ are compact sets, the set $\widetilde{\mathcal{F}}_{0}(\epsilon)$ is also a compact set for all $\epsilon>0$. Denote
\begin{eqnarray}
& & \hspace{ - 2 em} g\left(S,\left\{\pi_{\y}\right\},\left\{\bias(\ell)\right\}\right) \nonumber \\
& = & \alpha_{\text{RC}}\biggr\{\overline{\lambda}\biggr(\int \sum \limits_{(\y',\z') \in \mathcal{J}}1_{\left\{(\y',\z') = \biggr(\h^{\top}(\y,\z;0)x+\tau(\theta_{\y,\z})\biggr)\right\}} \biggr( \dfrac{\| x - \h(\y',\z';0)\|^{2}}{2} - \log(\pi_{\y',\z'})\biggr)dP(x)\biggr) \nonumber \\
& & + (1-\overline{\lambda}) \biggr(\int \sum \limits_{\z' \in \mathcal{L}}1_{\left\{\z' = \mathop{\arg \max}\limits_{\z \in \mathcal{L}} \biggr(\h^{\top}(\y,\z;0)x+\tau(\theta_{\y,\z})\biggr)\right\}} \biggr( \dfrac{\| x - \h(\y,\z';0)\|^{2}}{2} - \log(\pi_{\y,\z'})\biggr)dQ(x,\y)\biggr\} \nonumber \\
& & - \alpha_{\text{CE}} \int \log q_{\theta}(\y|x)dQ(x,\y) + \alpha_{\text{KL}} \int \sum \limits_{\y=1}^{K} q_{\theta}(\y|x)dQ(x,c) \log \biggr(\dfrac{q_{\theta}(\y|x)}{\pi_{\y}}\biggr)dP(x) \nonumber
\end{eqnarray}
for all $\left(S,\left\{\pi_{\y}\right\},\left\{\bias(\ell)\right\}\right)$. 
From the definition of $\widetilde{\mathcal{F}}_{0}(\epsilon)$, we have that 
\begin{eqnarray}
g\left(S,\left\{\pi_{\y}\right\},\left\{\bias(\ell)\right\}\right) > g\left(\widetilde{S}_{0}, \left\{\widetilde{\pi}_{\y}^{0}\right\},\left\{\widetilde{\bias}_{0}(\ell)\right\}\right) \nonumber
\end{eqnarray} 
for all $\left(S,\left\{\pi_{\y}\right\},\left\{\bias(\ell)\right\}\right) \in \widetilde{\mathcal{F}}_{0}(\epsilon)$ and $\left(\widetilde{S}_{0}, \left\{\widetilde{\pi}_{\y}^{0}\right\},\left\{\widetilde{\bias}_{0}(\ell)\right\}\right) \in \Gcal(\widetilde{\mathcal{F}}_{0})$. As $\widetilde{\mathcal{F}}_{0}(\epsilon)$ is a compact set, we further have that
\begin{eqnarray}
\inf \limits_{\left(S,\left\{\pi_{\y}\right\},\left\{\bias(\ell)\right\}\right) \in \widetilde{\mathcal{F}}_{0}(\epsilon)} g\left(S,\left\{\pi_{\y}\right\},\left\{\bias(\ell)\right\}\right) > g\left(\widetilde{S}_{0}, \left\{\widetilde{\pi}_{\y}^{0}\right\},\left\{\widetilde{\bias}_{0}(\ell)\right\} \right) \nonumber
\end{eqnarray} 
for all $\left(\widetilde{S}_{0}, \left\{\widetilde{\pi}_{\y}^{0}\right\},\left\{\widetilde{\bias}_{0}(\ell)\right\} \right) \in \Gcal(\widetilde{\mathcal{F}}_{0})$ and $\epsilon>0$. 

Now, according to the uniform laws of large numbers established in the proof of Theorem \ref{theorem:consistency_objective_LDAP_cross_entropy}, we have that $Y_{n} \to g\biggr(\widetilde{S}_{n},\left\{\widetilde{\pi}_{\y}\right\},\left\{\widetilde{\bias}(\ell)\right\}\biggr)$ almost surely as $n \to \infty$. According to the result of Theorem \ref{theorem:consistency_objective_LDAP_cross_entropy}, it implies that $g\biggr(\widetilde{S}_{n},\left\{\widetilde{\pi}_{\y}\right\},\left\{\widetilde{\bias}(\ell)\right\}\biggr) \to g\left(\widetilde{S}_{0}, \left\{\widetilde{\pi}_{\y}^{0}\right\},\left\{\widetilde{\bias}_{0}(\ell)\right\} \right)$ almost surely for all $\left(\widetilde{S}_{0}, \left\{\widetilde{\pi}_{\y}^{0}\right\},\left\{\widetilde{\bias}_{0}(\ell)\right\} \right) \in \Gcal(\widetilde{\mathcal{F}}_{0})$. Therefore, for each $\epsilon>0$ we can find sufficiently large $N$ such that we have
\begin{eqnarray}
\inf \limits_{\left(\widetilde{S}_{0}, \left\{\widetilde{\pi}_{\y}^{0}\right\},\left\{\widetilde{\bias}_{0}(\ell)\right\} \right) \in \Gcal(\widetilde{\mathcal{F}}_{0})} \biggr\{H(\widetilde{S}_{n},\widetilde{S}_{0}) + \sum \limits_{\y=1}^{K} |\widetilde{\pi}_{\y} - \widetilde{\pi}_{\y}^{0}| + \sum \limits_{\z \in \mathcal{L}} \|\widetilde{\bias}(\ell) - \widetilde{\bias}_{0}(\ell)\|\biggr\} < \epsilon \nonumber
\end{eqnarray}
almost surely for all $n \geq N$. As a consequence, we achieve the conclusion of the theorem.
\paragraph{PROOF OF THEOREM \ref{theorem:generalization_gap_classification_LDCE}}
The proof of the theorem is an application of Theorem 11 for generalization bound with margin from~\cite{Koltchinskii_2002} based on an evaluation of Rademacher complexity. In particular, we denote 
\begin{eqnarray}
\mathcal{J}_{n} & = & \biggr\{h_{\overline{\tau}_{n}}(\x,\y): \mathbb{R}^{D^{(0)}} \times \left\{1,\ldots,K\right\} \to \mathbb{R}| \nonumber \\
& &  h_{\overline{\tau}_{n}}(\x,\y) = \max \limits_{\z \in \mathcal{L}(\overline{\tau}_{n})} \biggr\{\h^{\top}(\y,\z;0)\x+ \tau(\theta_{\y,\z}) \biggr\} + \log \pi_{\y} \ \forall \ (\x,\y) \ \text{for some} \ |\mathcal{L}(\overline{\tau}_{n})| \leq \overline{\tau}_{n}|\mathcal{L}| \biggr\}. \nonumber
\end{eqnarray}
Now, we denote $\widetilde{\mathcal{J}}_{n} = \left\{h_{\overline{\tau}_{n}}(.,\y): \ \y \in \left\{1,\ldots,K\right\}, \ h_{\overline{\tau}_{n}} \in \mathcal{J}_{n} \right\}$. For any $\delta>0$, using the same argument as that of the proof of Theorem 11 in \cite{Koltchinskii_2002}, with probability at least $1-\delta$, we have
\begin{eqnarray}
R_{0}(f_{\overline{\tau}_{n}}) & \leq & \inf \limits_{\template \in (0,1]} \biggr\{R_{n,\template}(f_{\overline{\tau}_{n}})+\dfrac{8K(2K-1)}{\template}\Re_{n}(\widetilde{\mathcal{J}}_{n}) \nonumber \\
& & +\biggr(\dfrac{\log\log_{2}(2\template^{-1})}{n}\biggr)^{1/2} + \sqrt{\dfrac{\log(2\delta^{-1})}{2n}}\biggr\} \label{eqn:generalization_bound_classification_first}
\end{eqnarray}
where $\Re_{n}(\widetilde{\mathcal{J}}_{n})$ is Rademacher complexity of $\widetilde{\mathcal{J}}_{n}$, which in our case is defined as
\begin{eqnarray}
\Re_{n}(\widetilde{\mathcal{J}}_{n}) = \mathbb{E} \sup \limits_{\theta}\sup \limits_{|\mathcal{L}({\overline{\tau}_{n}})| \leq \overline{\tau}_{n}|\mathcal{L}|} \biggr|\dfrac{1}{n}\sum \limits_{i=1}^{n} \sigma_{i} \biggr(\max \limits_{g \in \mathcal{L}({\overline{\tau}_{n}})} \biggr\{\h^{\top}(\y,\z;0)\x_{i}+ \tau(\theta_{\y,\z}) \biggr\} + \log \pi_{\y}\biggr)\biggr| \nonumber
\end{eqnarray}
where $\sigma_{1},\ldots,\sigma_{n}$ are i.i.d. Rademacher random variables. Since $\overline{\gamma}$ is the lower bound of $\pi_{\y}$ for all $1 \leq \y \leq K$, we obtain that
\begin{eqnarray}
\Re_{n}(\widetilde{\mathcal{J}}) & \leq & \mathbb{E} \sup \limits_{\theta}\sup \limits_{|\mathcal{L}({\overline{\tau}_{n}})| \leq \overline{\tau}_{n}|\mathcal{L}|} \biggr|\dfrac{1}{n}\sum \limits_{i=1}^{n} \sigma_{i} \biggr(\max \limits_{g \in \mathcal{L}({\overline{\tau}_{n}})} \biggr\{\h^{\top}(\y,\z;0)\x_{i}+ \tau(\theta_{\y,\z}) \biggr\}\biggr)\biggr| \nonumber \\
& + &  \mathbb{E} \sup \limits_{\theta}\sup \limits_{|\mathcal{L}({\overline{\tau}_{n}})| \leq \overline{\tau}_{n}|\mathcal{L}|} \biggr|\dfrac{1}{n}\sum \limits_{i=1}^{n} \sigma_{i} \log \pi_{\y} \biggr| \nonumber \\
& \leq & \mathbb{E} \sup \limits_{\theta}\sup \limits_{|\mathcal{L}({\overline{\tau}_{n}})| \leq \overline{\tau}_{n}|\mathcal{L}|} \biggr|\dfrac{1}{n}\sum \limits_{i=1}^{n} \sigma_{i} \biggr(\max \limits_{g \in \mathcal{L}({\overline{\tau}_{n}})} \biggr\{\h^{\top}(\y,\z;0)\x_{i}+ \tau(\theta_{\y,\z}) \biggr\}\biggr)\biggr| + \dfrac{|\log \overline{\gamma}|}{\sqrt{n}} \nonumber
\end{eqnarray}
Furthermore, we have the following inequalities
\begin{eqnarray}
& & \hspace{ -8 em} \mathbb{E} \sup \limits_{\theta}\sup \limits_{|\mathcal{L}({\overline{\tau}_{n}})| \leq \overline{\tau}_{n}|\mathcal{L}|} \biggr|\dfrac{1}{n}\sum \limits_{i=1}^{n} \sigma_{i} \biggr(\max \limits_{g \in \mathcal{L}({\overline{\tau}_{n}})} \biggr\{\h^{\top}(\y,\z;0)\x_{i}+ \tau(\theta_{\y,\z}) \biggr\}\biggr)\biggr| \nonumber \\
& & \hspace{ 6 em} \leq \mathbb{E} \sup \limits_{|S'|  \leq  \overline{c}_{n}|\mathcal{L}|} \biggr|\dfrac{1}{n} \sum \limits_{i=1}^{n} \sigma_{i} \max \limits_{s \in S'} s^{\top}[\x_{i},1] \biggr| \nonumber \\
& & \hspace{ 6 em} \leq 2\overline{c}_{n}|\mathcal{L}| \mathbb{E} \sup \limits_{s \in \mathbb{B}(R)}\biggr|\dfrac{1}{n}\sum \limits_{i=1}^{n}\sigma_{i}s^{\top}[\x_{i},1]\biggr| \nonumber \\
& & \hspace{ 6 em} \leq 2\overline{c}_{n}|\mathcal{L}| R\mathbb{E}\biggr\|\dfrac{1}{n}\sum \limits_{i=1}^{n}\sigma_{i}[\x_{i},1]\biggr\| \nonumber \\
& & \hspace{ 6 em} \leq \dfrac{2\overline{c}_{n}|\mathcal{L}|(R^{2}+1)}{\sqrt{n}} \nonumber
\end{eqnarray}
where the final inequality is due to Cauchy-Schwartz's inequality. Combining the above results with \eqref{eqn:generalization_bound_classification_first}, we achieve the conclusion of the theorem.
\newpage
\section{Appendix D}
\label{Sec:appendix_D}
In this appendix, we provide further extensions of the main results in Appendix A and Appendix B.
\subsection{Additional discussion with DGM and cross entropy} \label{Section:more_connection_NRM_Cross_entropy}
Thus far, we have established the lower bound and the upper bound of maximizing the conditional log likelihood in terms of the cross entropy in Theorem \ref{theorem:cross_entropy_LD_DRM_full}. In the following full theorem, we will demonstrate that this cross entropy also the upper bound of maximizing the full posterior of DGM.
\begin{theorem} \label{theorem:full_posterior_cross_entropy}
Denote $\mathcal{A}_{\gamma} = \left\{\theta: \ \|\h(\y,\z;0)\| = \gamma \right\}$ for any $\gamma > 0$. For any $n \geq 1$ and $\sigma>0$, let $\x_{1},\ldots,\x_{n}$ be i.i.d. samples from the DGM. Then, the following holds
\begin{itemize}
\item[(a)] Upper bound:
\begin{eqnarray}
& & \hspace{ - 8 em} \max_{(\z_{i})_{i=1}^{n}, \theta \in \mathcal{A}_{\gamma}} \dfrac{1}{n} \sum \limits_{i=1}^{n} \log p(\y_{i},\z_{i}|\x_{i};\theta) \nonumber \\
& \leq & \max_{\theta \in \mathcal{A}_{\gamma}} \dfrac{1}{n} \sum_{i=1}^{n} \log \biggr(\text{Softmax}\biggr(\max_{\z_{i}}\biggr(\dfrac{\h^{\top}(\y_{i}, \z_{i}; 0)\x_{i} + \eta(\y_{i},\z_{i})}{\sigma^{2}}\biggr) + \bias_{\y_{i}} \biggr)\biggr) \nonumber \\
& = &  \max_{\theta \in \mathcal{A}_{\gamma}} \dfrac{1}{n} \sum_{i=1}^{n} \log q(\y_{i}|\x_{i}) = - \min_{\theta \in \mathcal{A}_{\gamma}} H_{p,q}(\y|\x) \nonumber
\end{eqnarray} 
where $\bias_{\y} = \log \pi_{\y}$ for all $1 \leq \y \leq K$, $q(\y|\x) = \text{Softmax} \biggr(\max \limits_{\z}\biggr(\h^{\top}(\y,\z;0)\x+\eta(\y,\z)\biggr)/\sigma^{2}+\bias_{\y}\biggr)$ for all $(\x,\y)$, and $H_{p,q}(\y|\x)$ is the cross-entropy between the estimated posterior $q(\y|\x)$ and the true posterior given by the ground-true labels $p(\y|\x)$. 
\item[(b)] Lower bound:
\begin{eqnarray}
\max_{(\z_{i})_{i=1}^{n}, \theta \in \mathcal{A}_{\gamma}} \dfrac{1}{n} \sum \limits_{i=1}^{n} \log p(\y_{i},\z_{i}|\x_{i};\theta) \geq -\log(|\mathcal{L}|) - \min_{\theta \in \mathcal{A}_{\gamma}} H_{p,q}(\y|\x)  \nonumber
\end{eqnarray}
where $|\mathcal{L}|$ denotes the total number of possible rendering paths with a given label.
\end{itemize}
\end{theorem}
\begin{remark}
Combining with the bounds in Theorem \ref{theorem:cross_entropy_LD_DRM_full}, we obtain that
\begin{eqnarray}
 \max_{(\z_{i})_{i=1}^{n}, \theta \in \mathcal{A}_{\gamma}} \sum_{i=1}^{n} \ln p(\y_{i}|\x_{i}, \z_{i};\theta) \geq - \min \limits_{\theta} H_{p,q}(\y|\x) \geq \max_{(\z_{i})_{i=1}^{n}, \theta \in \mathcal{A}_{\gamma}} \dfrac{1}{n} \sum \limits_{i=1}^{n} \log p(\y_{i},\z_{i}|\x_{i};\theta). \nonumber
\end{eqnarray}
Therefore, the cross entropy loss is the lower bound and the upper bound of the conditional log likelihood and the full posterior of DGM respectively.
\end{remark}
\subsection{Relaxation of constant norm with rendered images} 
\label{Section:relaxation_constant_norm}
So far, we have assumed constant norm of rendered image $\h(\y,\z;0)$ to derive the results in part (a) of Theorem~\ref{theorem:learning_lddrm} and Theorem \ref{theorem:full_posterior_cross_entropy}. We would like to emphasize that this assumption is just for the simplicity of the argument and the elegance of the statements in these results. Without this assumption, the results of these theorems may need to change as follows. We denote $\mathcal{B}_{M_{1},M_{2}} : = \left\{\theta: M_{1} \leq \|\h(\y,\z;0)\| \leq M_{2}\right\}$ for some given non-negative constants $M_{1} < M_{2}$. Here, $\theta \in \mathcal{B}_{M_{1},M_{2}}$ does not need to satisfy the non-negativity assumption. Then, we obtain that
\begin{theorem} \label{theorem:cross_entropy_LD_DRM_no_constraint} (Relaxation of Theorem \ref{theorem:cross_entropy_LD_DRM_full})
For any $n \geq 1$ and $\sigma>0$, let $\x_{1},\ldots,\x_{n}$ be i.i.d. samples from the DGM. Given the formulation of $\mathcal{B}_{M_{1},M_{2}}$, the following holds \\
(a) Lower bound:
\begin{eqnarray}
& & \hspace{- 4 em} \max_{(\z_{i})_{i=1}^{n}, \theta \in \mathcal{B}_{M_{1},M_{2}}}\dfrac{1}{n} \sum_{i=1}^{n} \ln p(\y_{i}|\x_{i}, \z_{i};\theta) \nonumber \\
& & \geq \max_{\theta \in \mathcal{B}_{M_{1},M_{2}}} \dfrac{1}{n} \sum_{i=1}^{n} \log \biggr(\text{Softmax}\biggr(g(\y_{i},\x_{i})\biggr) + \bias_{\y_{i}} \biggr)\biggr) - \dfrac{M_{2}^{2}-M_{1}^{2}}{2\sigma^{2}} \nonumber \\
& & = \max_{\theta \in \mathcal{B}_{M_{1},M_{2}}} \dfrac{1}{n} \sum_{i=1}^{n} \log \widetilde{q}(\y_{i}|\x_{i}) - \dfrac{M_{2}^{2}-M_{1}^{2}}{2\sigma^{2}} \nonumber \\
& & = - \min_{\theta \in \mathcal{B}_{M_{1},M_{2}}} H_{p,\widetilde{q}}(\y|\x) - \dfrac{M_{2}^{2}-M_{1}^{2}}{2\sigma^{2}} \nonumber
\end{eqnarray}
where $\bias_{\y}= \log \pi_{\y}$ as $1 \leq \y \leq K$, $\widetilde{q}(\y|\x) = \text{Softmax} \biggr(h(\y,\x)+\bias_{\y}\biggr)$ for all $(\x,\y)$, and 
\begin{eqnarray}
& & \hspace{ - 3 em} g(\y,\x) : = \dfrac{1}{\sigma^{2}}\h^{\top}(\y) \maxpool\biggr(\relu\left(\convo\biggr(\template(\ell),\cdots \maxpool\biggr(\relu\biggr(\convo\biggr(\template(1), \x \right)+\bias(1)\biggr)\biggr)\biggr) \nonumber \\
& & \hspace{ 30 em}\cdots +\bias(\ell) \biggr)\biggr) \nonumber
\end{eqnarray}
for all $(\x,\y)$. \\
(b) Upper bound: 
\begin{eqnarray}
& & \hspace {- 3 em} \max_{(\z_{i})_{i=1}^{n}, \theta \in \mathcal{B}_{M_{1},M_{2}}}\dfrac{1}{n} \sum_{i=1}^{n} \ln p(\y_{i}|\x_{i}, \z_{i};\theta) \nonumber \\
& & \hspace {- 3 em} \leq \max_{\theta \in \mathcal{B}_{M_{1},M_{2}}} \dfrac{1}{n} \sum_{i=1}^{n}\biggr\{\log \biggr(\text{Softmax}\biggr(g(\y_{i},\x_{i})\biggr) + \bias_{\y_{i}} \biggr)\biggr) + \biggr\{\underbrace{\max \limits_{\z_{i}}\biggr(\dfrac{\h^{\top}(\y_{i}, \z_{i}; 0)\x_{i} + \eta(\y_{i},\z_{i})}{\sigma^{2}}\biggr) - g(\y_{i},\x_{i})}_{\text{CNN loss}} \biggr\} \nonumber \\ 
& & + \max \limits_{\y} \biggr(g(\y,\x_{i}) - \dfrac{\h^{\top}(\y,\overline{\z}_{i}; 0)\x_{i} + \eta(\y,\overline{\z}_{i})}{\sigma^{2}} \biggr)\biggr\} + \log K + \dfrac{M_{2}^{2}-M_{1}^{2}}{2\sigma^{2}} \nonumber
\end{eqnarray}
where $\overline{\z}_{i} = \mathop{\arg \max} \limits_{\z_{i}} p(\y_{i}|\x_{i},\z_{i};\theta)$ for $1 \leq i \leq n$.
\end{theorem}
\begin{remark}
Comparing to the lower bound in Theorem \ref{theorem:cross_entropy_LD_DRM_full}, the cross entropy term $H_{p,\widetilde{q}}$ has a direct connection to CNN without relying on the non-negativity assumption. Additionally, we need to pay a price of $(M_{2}^2-M_{1}^2)/(2\sigma^{2})$ as we relax the constant norm assumption with rendering paths $\h(\y,\z;0)$. Last but not least, comparing to the upper bound in Theorem \ref{theorem:cross_entropy_LD_DRM_full}, apart from the inclusion of $(M_{2}^2-M_{1}^2)/(2\sigma^{2})$, we also need to pay an additional price of CNN loss as we directly connect the CNN structure to the cross entropy term $H_{p,\widetilde{q}}$ without hinging on the non-negativity assumption.
\end{remark}
Similar the the changes with results of Theorem \ref{theorem:cross_entropy_LD_DRM_no_constraint}, we also have the following modifications regarding the upper bound and lower bound of full posterior of DGM under the relaxation of constant norm with rendered images $\h(\y,\z;0)$.
\begin{theorem} \label{theorem:cross_entropy_full_posteterior_LD_DRM_no_constraint} (Relaxation of Theorem \ref{theorem:full_posterior_cross_entropy})
For any $n \geq 1$ and $\sigma>0$, let $\x_{1},\ldots,\x_{n}$ be i.i.d. samples from the DGM. Given the formulation of $\mathcal{B}_{M_{1},M_{2}}$, the following holds \\
(a) Upper bound:
\begin{eqnarray}
& & \hspace {- 3 em} \max_{(\z_{i})_{i=1}^{n}, \theta \in \mathcal{B}_{M_{1},M_{2}}} \dfrac{1}{n} \sum \limits_{i=1}^{n} \log p(\y_{i},\z_{i}|\x_{i};\theta) \nonumber \\
& & \hspace {- 3 em} \leq \max_{\theta \in \mathcal{B}_{M_{1},M_{2}}} \dfrac{1}{n} \sum_{i=1}^{n}\biggr\{\log \biggr(\text{Softmax}\biggr(g(\y_{i},\x_{i})\biggr) + \bias_{\y_{i}} \biggr)\biggr) + \biggr\{\underbrace{\max \limits_{\z_{i}}\biggr(\dfrac{\h^{\top}(\y_{i}, \z_{i}; 0)\x_{i} + \eta(\y_{i},\z_{i})}{\sigma^{2}}\biggr) - g(\y_{i},\x_{i})}_{\text{CNN loss}} \biggr\} \nonumber \\
& & \hspace {30 em} + \dfrac{M_{2}^{2}-M_{1}^{2}}{2\sigma^{2}} \nonumber \\
& & \hspace {- 3 em} =  \max_{\theta \in \mathcal{B}_{M_{1},M_{2}}} \dfrac{1}{n} \sum_{i=1}^{n} \log \widetilde{q}(\y_{i}|\x_{i}) + \biggr\{\max \limits_{\z_{i}}\biggr(\dfrac{\h^{\top}(\y_{i}, \z_{i}; 0)\x_{i} + \eta(\y,\z_{i})}{\sigma^{2}}\biggr) - g(\y_{i},\x_{i}) \biggr\} + \dfrac{M_{2}^{2}-M_{1}^{2}}{2\sigma^{2}} \nonumber \\
& & \hspace {- 3 em} = - \min_{\theta \in \mathcal{B}_{M_{1},M_{2}}} H_{p,\widetilde{q}}(\y|\x) + \biggr\{\max \limits_{\z_{i}}\biggr(\dfrac{\h^{\top}(\y_{i}, \z_{i}; 0)\x_{i} + \eta(\y,\z_{i})}{\sigma^{2}}\biggr) - g(\y_{i},\x_{i}) \biggr\} + \dfrac{M_{2}^{2}-M_{1}^{2}}{2\sigma^{2}} \nonumber
\end{eqnarray} 
where $\bias_{\y}, \widetilde{q}(\y|\x)$, and $g(\y,\x)$ are defined as in Theorem \ref{theorem:cross_entropy_LD_DRM_no_constraint}. \\
(b) Lower bound: 
\begin{eqnarray}
& & \hspace {- 5 em} \max_{(\z_{i})_{i=1}^{n}, \theta \in \mathcal{B}_{M_{1},M_{2}}} \dfrac{1}{n} \sum \limits_{i=1}^{n} \log p(\y_{i},\z_{i}|\x_{i};\theta) \nonumber \\
& & \geq \max_{\theta \in \mathcal{B}_{M_{1},M_{2}}} \dfrac{1}{n} \sum_{i=1}^{n}\biggr\{\log \biggr(\text{Softmax}\biggr(g(\y_{i},\x_{i})\biggr) + \bias_{\y_{i}} \biggr)\biggr) \nonumber \\
& & + \biggr(\max \limits_{\y} g(\y,\x_{i}) - \max \limits_{\y,\z}\biggr(\dfrac{\h^{\top}(\y,\z;0)\x_{i} + \eta(\y,\z)}{\sigma^{2}}\biggr)\biggr)\biggr\} -\log(|\mathcal{J}|) - \dfrac{M_{2}^{2}-M_{1}^{2}}{2\sigma^{2}} \nonumber
\end{eqnarray}
\end{theorem}
\begin{remark}
Comparing to the upper bound in Theorem \ref{theorem:full_posterior_cross_entropy}, we have the additional CNN loss and the term $(M_{2}^2- M_{1}^2)/(2\sigma^2)$ as we relax the constant norm with rendered image $\h(\y,\z;0)$ as well as the non-negativity assumption to CNN structure with the cross entropy term. Moreover, in addition to the price of CNN loss and of the term $(M_{2}^2- M_{1}^2)/(2\sigma^2)$, we also need to pay the price of $\log(|\mathcal{J}|)$ instead of $\log(|\mathcal{L}|)$ in Theorem \ref{theorem:full_posterior_cross_entropy}. 
\end{remark} 
\subsection{Statistical guarantees for unsupervised DGM without noise} 
\label{Section:unsupervised_LD_DRMM}
As being mentioned in Appendix B, under the non-negativity assumption of the intermediate rendered images $\h(\y,\z;\ell)$, the tractable relaxation of objective function of unsupervised DGM without noise can be formulated as
\begin{eqnarray}
\hspace{ -1 em} V_{n} := \min \limits_{\theta} \dfrac{1}{n} \sum \limits_{i=1}^{n} \sum \limits_{(\y',\z') \in \mathcal{J}}1_{\left\{(\y',\z') = \mathop{\arg \max}\limits_{(\y,\z; 0) \in \mathcal{J}} \biggr(\h^{\top}(\y,\z;0)\x_{i}+\tau(\theta_{\y,\z})\biggr) \right\}} \biggr( \dfrac{\| \x_{i} - \h(\y',\z'; 0)\|^{2}}{2} - \log(p_{\y',z'})\biggr) \label{eqn:objective_unsupervised_LDDRMM_no_noise}
\end{eqnarray}
where $\h(\y,\z; 0) = \Lambda(\z;1)\ldots \Lambda(\z;L)\muy(\y)$, $\Lambda(\z;l) = \sum_{\p\in \pp(\ell)}  \sparse(\ell, \p)\ttrans(t;\ell,\p) \pad(\ell, \p)\template(\ell, \p)$, and $p_{\y',\z'} = \exp\biggr(\tau(\theta_{\y',\z'})+\log \pi_{\y'}\biggr)/\biggr(\sum \limits_{\y=1}^{K}\exp\biggr(\max \limits_{\z \in \mathcal{L}} \tau(\theta_{\y,\z})+\log \pi_{\y}\biggr)\biggr)$. For each $\theta$, we denote $S = S(\theta) = \left\{\h(\y,\z; 0): \ (\y,\z) \in \mathcal{J} \right\}$ the set of all possible \textit{rendered images} associated with $\theta$. Since the rendered images in $S$ can be identical, it is clear that the number of elements of $S$, which is defined as $|S|$, is upper bounded by a fixed $|\mathcal{J}|$ total number of rendered images, which depends on the number of layers $L$ and the number of pixels in each layer. Similar to the semi-supervised setting in~\eqref{eqn:full_objective_semi_sup_SPLD_K_means}, a notable property of $S$ is that all of their rendered images share the parameters $\template(\ell)$ and $\muy(\y)$ for all $1 \leq  \ell \leq L$ and $1 \leq \y \leq K$. Therefore, we call the above objective function \textit{shared parameters latent dependence regularized K-means (SPLD regularized K-means)}.
\paragraph{Existence of optimal solutions} When there are no constraints with the parameter spaces $\Theta_{\ell}$, $\Omega$, and $\Xi_{l}$ of $\muy(\y)$, $\template(\ell)$, and $\bias(\ell)$ as $1 \leq \y \leq K$, $1 \leq  \ell \leq L$, and $g \in \mathcal{L}$, the existence of optimal solutions of objective function \eqref{eqn:objective_unsupervised_LDDRMM_no_noise} is not guaranteed. To ensure this existence, we will impose the compact contraints on these parameter spaces.
\begin{lemma} \label{lemma:existence_optimal_sets}
Assume that $\Theta_{\ell}$ is a compact subset of $\mathbb{R}^{F(\ell) \times D(\ell)}$, $\Omega$ is a compact subset of $\mathbb{R}^{D(\ell)}$, and $\Xi_{l}$ is a compact subset of $\mathbb{R}^{D(\ell)}$ for $1 \leq  \ell \leq L$ under Frobenius norm. Then, there exist optimal solutions $\widehat{\theta}=\biggr(\left\{\widehat{\muy}(y) \right\}_{\y=1}^{K},\left\{\widehat{\template}(\ell)\right\}_{\ell=1}^{L},\left\{\widehat{\pi}_{y}\right\}_{\y=1}^{K},\left\{\widehat{b}(\ell) \right\}_{\ell=1}^{L} \biggr)$ of SPLD regularized K-means.
\end{lemma}
To simplify the presentation later, we will assume throughout this section the following key assumptions:
\begin{itemize}
\item The parameter spaces $\Theta_{\ell}$, $\Omega$, $\Xi_{l}$ are compact subsets of $\mathbb{R}^{F(\ell) \times D(\ell)}$, $\mathbb{R}^{D(\ell)}$, and $\mathbb{R}^{D(\ell)}$ respectively for $1 \leq  \ell \leq L$.
\item The prior probability $\pi = \left\{\pi_{\y}\right\}_{\y=1}^{K}$ of labels in $\theta$ is such that $\pi_{\y}>\overline{\gamma}$ as $1 \leq \y \leq K$ for some given sufficiently small $\overline{\gamma}>0$.
\end{itemize} 
The assumption with prior probability $\pi$ is to guarantee that the value of $\log(p_{\y,\z})$ will not go to $-\infty$ as the value of $\pi_{\y}$ becomes close to 0.
\paragraph{Consistency of SPLD regularized K-means}
Now, we would like to establish the consistency of the objective function and optimal solutions of SPLD regularized K-means. We assume that $\x_{1},\ldots,\x_{n}$ are i.i.d. samples from a true but unknown distribution $P$ with finite second moment, i.e., $\int \|x\|^2dP(x) < \infty$. 
\begin{theorem} \label{theorem:consistency_SPLDAP_Kmeans} (Consistency of objective function of SPLD regularized K-means) 
Assume that $\mathbb{P}(\|I\| \leq R) = 1$ as $I \sim P$ for some given $R>0$. We denote the population version of SPLD regularized K-means as follows
\begin{eqnarray}
\overline{V} : = \min \limits_{\theta} \int \sum \limits_{(\y',\z') \in \mathcal{J}}1_{\left\{(\y',\z') = \mathop{\arg \max}\limits_{(\y,\z) \in \mathcal{J}} \biggr(\h^{\top}(\y,\z; 0)x+\tau(\theta_{\y,\z})\biggr)\right\}} \biggr( \dfrac{\| x - \h(\y',\z'; 0)\|^{2}}{2} - \log(p_{\y',\z'})\biggr)dP(x). \nonumber
\end{eqnarray}
Then, we obtain that $V_{n} \to \overline{V}$ almost surely as $n \to \infty$. 
\end{theorem}
So far, we have established the convergence of objective function of SPLD K-means to the objective function of its corresponding population version. Now, we would like to study the convergence of optimal rendered images and optimal solutions of SPLD regularized K-means to those of population SPLD regularized K-means. In particular, we denote $\mathcal{F}_{0}$ the set of all optimal solutions $\theta^{0} = \biggr(\left\{\muy^{0}(y)\right\}_{\y=1}^{K},\left\{\template_{0}(l)\right\}_{\ell=1}^{L},\left\{\pi_{\y}^{0}\right\}_{\y=1}^{K},\left\{\bias_{0}(\ell) \right\}_{\ell = 1}^{L} \biggr)$ of population SPLD regularized K-means. For each $\theta^{0}\in \mathcal{F}_{0}$, we define $S_{0}$ the set of optimal rendered images associated with $\theta^{0}$. We denote $\Gcal(\mathcal{F}_{0})$ the corresponding set of all optimal rendered images $S_{0}$, optimal prior probabilities $\left\{\pi_{\y}^{0}\right\}_{\y=1}^{K}$, and optimal biases $\left\{\bias_{0}(\ell) \right\}_{\ell = 1}^{L}$. 
\begin{theorem} \label{theorem:consistency_optimal_solutions_SPLDAP_Kmeans} (Consistency of optimal rendering paths and optimal solutions of SPLD regularized K-means) 
Assume that $\mathbb{P}(\|\x\| \leq R) = 1$ as $\x \sim P$ for some given $R>0$. Then, we obtain that 
\begin{eqnarray}
\inf \limits_{\left(S_{0}, \left\{\pi_{\y}^{0}\right\},\left\{\bias_{0}(\ell) \right\}\right) \in \Gcal(\mathcal{F}_{0})} \biggr\{H(\widehat{S}_{n},S_{0}) + \sum \limits_{\y=1}^{K} |\widehat{\pi}_{y} - \pi_{\y}^{0}| + \sum \limits_{\ell=1}^{L} \|\widehat{b}(\ell) - \bias_{0}(\ell)\|\biggr\}  \to 0 \nonumber
\end{eqnarray}
almost surely as $n \to \infty$. Here, $H(.,.)$ stands for the Hausdorff metric.  
\end{theorem}
\paragraph{Convergence rate with clustering risk of SPLD regularized K-means} 
Remind that we denote $\widehat{\theta} := \biggr(\left\{\widehat{\muy}(y) \right\}_{\y=1}^{K},\left\{\widehat{\template}(\ell)\right\}_{\ell=1}^{L},\left\{\widehat{\pi}_{y}\right\}_{\y=1}^{K},\left\{\widehat{b}(\ell) \right\}_{\ell=1}^{L} \biggr)$ the optimal solutions of SPLD regularized K-means. These optimal solutions lead to the corresponding set of optimal rendered images $\widehat{S}_{n} = \left\{\widehat \h(\y,\z; 0): \ (\y,\z) \in \mathcal{J} \right\}$ where 
\begin{align}
\widehat \h(\y,\z; 0) & : = \widehat \Lambda(\z;1)\ldots \widehat \Lambda(\z;L)\widehat \muy(\y), \nonumber \\
\widehat \Lambda(\z;\ell) & : = \sum_{\p\in \pp(\ell)}  \sparse(\ell, \p)\ttrans(t;\ell,\p) \pad(\ell, \p)\widehat \template(\ell, \p) \nonumber
\end{align}
for all $1 \leq  \ell \leq L$. In practice, only a small fraction of optimal rendered images in $\widehat{S}_{n}$ is indeed useful for clustering the data $\x_{1},\ldots,\x_{n}$. These optimal rendered images correspond to the optimal \textit{active rendering paths} of $\mathcal{J}$. In particular, we denote a smallest subset $\overline{\mathcal{J}}_{n}$ of $\mathcal{J}$ such that $|\overline{\mathcal{J}}_{n}| \leq \overline{c}_{n}|\mathcal{J}|$ and the following holds
\begin{eqnarray}
V_{n} =  \dfrac{1}{n} \sum \limits_{i=1}^{n} \sum \limits_{(\y',\z') \in \overline{\mathcal{J}}_{n}}1_{\left\{(\y',\z') = \mathop{\arg \max}\limits_{(\y,\z) \in \overline{\mathcal{J}}_{n}} \biggr(\widehat \h(\y,\z; 0)^{\top}\x_{i}+\tau(\widehat{\theta}_{\y,\z})\biggr) \right\}} \biggr(\dfrac{\| \x_{i} - \widehat \h(\y',\z'; 0) \|^{2}}{2} - \log(\widehat{p}(\y',\z'))\biggr). \nonumber
\end{eqnarray}
where $\widehat{p}(\y',\z') = \exp\biggr(\tau(\widehat{\theta}_{\y',\z'})+\log \widehat{\pi}_{y'}\biggr)/\biggr(\sum \limits_{\y=1}^{K}\exp\biggr(\max \limits_{\z \in \mathcal{L}} \tau(\widehat{\theta}_{\y,\z})+\log \widehat{\pi}_{\y}\biggr)\biggr)$. The set $\overline{\mathcal{J}}_{n}$ is called the set of optimal active paths of $\mathcal{J}$ while the positive number $\overline{c}_{n} \leq 1$ is called the ratio of active rendering paths. Throughout this section, we assume that $\overline{c}_{n}$ is independent of $\x_{1},\ldots,\x_{n}$ and only depends on sample size $n$. 

Inspired by the idea of generalization gap in classification setting, the measure that we utilize to evaluate the convergence rate in clustering is the difference between the objective function of SPLD regularized K-means and its corresponding clustering risk, which can be defined as follows
\begin{eqnarray}
\overline{V}_{n} = \int \biggr(\sum \limits_{(\y',\z') \in \overline{\mathcal{J}}_{n}}1_{\left\{(\y',\z') = \mathop{\arg \max}\limits_{(\y,\z) \in \overline{\mathcal{J}}_{n}} \biggr(\widehat \h(\y,\z; 0)^{\top}\x +\tau(\widehat{\theta}_{\y,\z})\biggr) \right\}} \biggr(\dfrac{\| \x - \widehat \h(\y,\z; 0) \|^{2}}{2} - \log(\widehat{p}(\y',\z'))\biggr)\biggr)dP(x). \nonumber 
\end{eqnarray}
for all $n \geq 1$. We have the following result establishing the generalization gap between $V_{n}$ and $\overline{V}_{n}$ based on the number of active rendering paths in SPLD regularized K-means.
\begin{theorem} \label{theorem:generalization_bound_SPLDAP K-means} (Convergence rate of SPLD regularized K-means) Assume that $\mathbb{P}(\|\x\| \leq R) = 1$ as $\x \sim P$ for some given $R>0$. Additionally, the parameter spaces $\Theta_{\ell}$ and $\Omega$ are chosen such that $\|\h(\y,\z; 0)\| \leq R$ for all $(\y,\z) \in \Jcal$. Then, the following holds
\begin{eqnarray}
\biggr|\mathbb{E} V_{n} - \mathbb{E}\overline{V}_{n}\biggr| \leq \dfrac{4 \overline{c}_{n}|\Jcal|R^{2}}{\sqrt{n}} + \left(2\log(1/\overline{\gamma})+R^{2}\right)\sqrt{\dfrac{2 (\log 8+(D^{(0)}+1)(\overline{c}_{n}|\Jcal|-1)\log n)}{n}}\nonumber
\end{eqnarray}
where $\overline{\gamma}$ is a given lower bound for the prior probability $\pi_{\y}$ as $1 \leq \y \leq K$. 
\end{theorem}
The proof of Theorem~\ref{theorem:generalization_bound_SPLDAP K-means} is deferred to Appendix~\ref{Sec:appendix_E}. 
\newpage
\section{Appendix E}
\label{Sec:appendix_E}
In this appendix, we provide detail proofs for the remaining results in Appendix~\ref{Sec:appendix_D}. 
\paragraph{PROOF OF THEOREM \ref{theorem:full_posterior_cross_entropy}}
(a) Regarding the lower bound, we obtain that
\begin{eqnarray}
& & \hspace{ - 8 em} \max_{\theta \in \mathcal{A}_{\gamma}} \dfrac{1}{n} \sum \limits_{i=1}^{n} \max \limits_{\z_{i}}\log p(\y_{i},\z_{i}|\x_{i};\theta) \nonumber \\
& & =  \max_{\theta \in \mathcal{A}_{\gamma}} \dfrac{1}{n} \sum \limits_{i=1}^{n} \max \limits_{\z_{i}}\log \dfrac{p(\x_{i}|\y_{i},\z_{i};\theta) p(\y_{i},\z_{i};\theta)}{\sum \limits_{\y,\z} p(\x_{i}|\y,\z;\theta) p(\y,\z;\theta)} \nonumber \\
& & \overset{(a)}{\leq} \max_{\theta \in \mathcal{A}_{\gamma}} \dfrac{1}{n} \sum \limits_{i=1}^{n} \max \limits_{\z_{i}} \log \dfrac{p(\x_{i}|\y_{i},\z_{i};\theta)p(\y_{i},\z_{i};\theta)}{\sum \limits_{\y=1}^{K} \max \limits_{\z} p(\x_{i}|\y,\z;\theta) p(\y,\z;\theta)} \nonumber \\
& & = \max_{\theta \in \mathcal{A}_{\gamma}} \dfrac{1}{n} \sum \limits_{i=1}^{n} \log\biggr(\text{Softmax} \biggr(\max \limits_{\z_{i}} \acts_{i}(\y_{i},\z_{i})\biggr)\biggr) \nonumber \\
& & = \max_{\theta \in \mathcal{A}_{\gamma}} \dfrac{1}{n} \sum_{i=1}^{n} \log q(\y_{i}|\x_{i}) = - \min_{\theta \in \mathcal{A}_{\gamma}} H_{p,q}(\y|\x) \nonumber
\end{eqnarray}
where the inequality in (a) is due to the fact that 
\begin{eqnarray}
\sum \limits_{\y,\z} p(\x_{i}|\y,\z;\theta) p(\y,\z;\theta) \geq \sum \limits_{\y=1}^{K} \max \limits_{\z} p(\x_{i}|\y,\z;\theta) p(\y,\z;\theta) \nonumber
\end{eqnarray} 
for all $1 \leq i \leq n$. Therefore, we achieve the conclusion of part (a) of the theorem.

(b) Regarding the upper bound, according to the above formulations, we also have
\begin{eqnarray}
& & \hspace{ - 6 em} \max_{\theta \in \mathcal{A}_{\gamma}} \dfrac{1}{n} \sum \limits_{i=1}^{n} \max \limits_{\z_{i}}\log p(\y_{i},\z_{i}|\x_{i};\theta) \nonumber \\
& & \geq \max_{\theta \in \mathcal{A}_{\gamma}} \dfrac{1}{n} \sum \limits_{i=1}^{n} \log \dfrac{\max \limits_{\z_{i}} p(\x_{i}|\y_{i},\z_{i};\theta)p(\y_{i},\z_{i};\theta)}{|\mathcal{L}| \sum \limits_{\y} \max \limits_{\z} p(\x_{i}|\y,\z;\theta) p(\y,\z;\theta)} \nonumber \\
& & = -\log(|\mathcal{L}|) - \min_{\theta \in \mathcal{A}_{\gamma}} H_{p,q}(\y|\x) \nonumber
\end{eqnarray}
where $|\mathcal{L}|$ denote the total number of possible paths with a given label. Note that, the above inequality is due to the simple upper bound
\begin{eqnarray}
\sum \limits_{\y,\z} p(\x_{i}|\y,\z;\theta) p(\y,\z;\theta) \leq |\mathcal{L}| \sum \limits_{\y} \max \limits_{\z} p(\x_{i}|\y,\z;\theta) p(\y,\z;\theta) \nonumber
\end{eqnarray}
for all $1 \leq i \leq n$.
\paragraph{PROOF OF THEOREM \ref{theorem:cross_entropy_LD_DRM_no_constraint}}
(a) Using the same argument as that in the proof of Theorem \ref{theorem:cross_entropy_LD_DRM_full}, we obtain the following equation
\begin{eqnarray}
& & \max_{(\z_{i})_{i=1}^{n}, \theta \in \mathcal{B}_{M_{1},M_{2}}}\dfrac{1}{n} \sum_{i=1}^{n} \log p(\y_{i}|\x_{i}, \z_{i};\theta) \nonumber \\
& & = \max_{\theta \in \mathcal{B}_{M_{1},M_{2}}} \dfrac{1}{n} \sum_{i=1}^{n} \max_{\z_{i}}\log \left(\dfrac{ p(\x_{i}|\y_{i}, \z_{i};\theta) \exp\biggr(\dfrac{\eta(\y_{i},\z_{i})}{\sigma^{2}}\biggr)\pi_{\y_{i}}}{\sum \limits_{\y=1}^{K} p(\x_{i}|\y, \z_{i};\theta)\exp\biggr(\dfrac{\eta(\y,\z_{i})}{\sigma^{2}} \biggr)\pi_{\y}}\right) \nonumber
\end{eqnarray}
From the assumption that $M_{1} \leq \|\h(\y,\z_{i};0)\| \leq M_{2}$, for any $(\y,\z_{i})$, it is clear that
\begin{eqnarray}
p(\x_{i}|\y, \z_{i};\theta)\exp\biggr(\dfrac{\eta(\y,\z_{i})}{\sigma^{2}} \biggr)\pi_{\y} \leq \dfrac{1}{(\sqrt{2\pi}\sigma)^{D^{(0)}}}\exp\biggr(-\dfrac{\|\x_{i}\|^{2}-2\h^{\top}(\y,\z_{i}; 0)\x_{i} + M_{1}^{2}-2\eta(\y,\z_{i})}{2\sigma^{2}}\biggr)\pi_{\y} \nonumber
\end{eqnarray}
and 
\begin{eqnarray}
p(\x_{i}|\y, \z_{i};\theta)\exp\biggr(\dfrac{\eta(\y,\z_{i})}{\sigma^{2}} \biggr)\pi_{\y} \geq \dfrac{1}{(\sqrt{2\pi}\sigma)^{D^{(0)}}}\exp\biggr(-\dfrac{\|\x_{i}\|^{2}-2\h^{\top}(\y,\z_{i}; 0)\x_{i}+M_{2}^{2} -2\eta(\y,\z_{i})}{2\sigma^{2}}\biggr)\pi_{\y}. \nonumber
\end{eqnarray}
Therefore, we achieve that
\begin{eqnarray}
& & \max_{(\z_{i})_{i=1}^{n}, \theta \in \mathcal{B}_{M_{1},M_{2}}}\dfrac{1}{n} \sum_{i=1}^{n} \log p(\y_{i}|\x_{i}, \z_{i};\theta) \nonumber \\
& & \geq \max_{\theta \in \mathcal{B}_{M_{1},M_{2}}} \dfrac{1}{n} \sum_{i=1}^{n} \max_{\z_{i}} \log \left(\dfrac{\exp\biggr(\dfrac{\h^{\top}(\y_{i}, \z_{i}; 0)\x_{i}+\eta(\y_{i},\z_{i})}{\sigma^{2}}\biggr)\pi_{\y_{i}}}{\sum \limits_{\y=1}^{K} \exp\biggr(\dfrac{\h^{\top}(\y,\z_{i};0)\x_{i}+\eta(\y,\z_{i})}{\sigma^{2}}\biggr)\pi_{\y}} \right) - \dfrac{M_{2}^{2}-M_{1}^{2}}{2\sigma^{2}} \nonumber \\
& & \geq \max_{\theta \in \mathcal{B}_{M_{1},M_{2}}} \dfrac{1}{n} \sum_{i=1}^{n} \log \biggr(\text{Softmax}\biggr(g(\y_{i},\x_{i})\biggr) + \bias_{\y_{i}} \biggr)\biggr) - \dfrac{M_{2}^{2}-M_{1}^{2}}{2\sigma^{2}} \nonumber \\
& & = \max_{\theta \in \mathcal{B}_{M_{1},M_{2}}} \dfrac{1}{n} \sum_{i=1}^{n} \log \widetilde{q}(\y_{i}|\x_{i}) - \dfrac{M_{2}^{2}-M_{1}^{2}}{2\sigma^{2}} = - \min_{\theta \in \mathcal{B}_{M_{1},M_{2}}} H_{p,\widetilde{q}}(\y|\x) - \dfrac{M_{2}^{2}-M_{1}^{2}}{2\sigma^{2}} \nonumber
\end{eqnarray}
As a consequence, we achieve the conclusion of part (a) of the theorem.

(b) Regarding the upper bound, we have
\begin{eqnarray}
& & \max_{(\z_{i})_{i=1}^{n}, \theta \in \mathcal{B}_{M_{1},M_{2}}}\dfrac{1}{n} \sum_{i=1}^{n} \log p(\y_{i}|\x_{i}, \z_{i};\theta) \nonumber \\
& & \leq \max_{\theta \in \mathcal{B}_{M_{1},M_{2}}} \dfrac{1}{n} \sum_{i=1}^{n} \log \left(\dfrac{\exp\biggr(\dfrac{\h^{\top}(\y_{i},\overline{\z}_{i}; 0)\x_{i}+\eta(\y_{i},\overline{\z}_{i})}{\sigma^{2}}\biggr)\pi_{\y_{i}}}{\sum \limits_{\y=1}^{K} \exp\biggr(\dfrac{\h^{\top}(\y,\overline{\z}_{i}; 0)\x_{i}+\eta(\y,\overline{\z}_{i})}{\sigma^{2}}\biggr)\pi_{\y}} \right) + \dfrac{M_{2}^{2}-M_{1}^{2}}{2\sigma^{2}} \nonumber \\
& & \leq \max_{\theta \in \mathcal{B}_{M_{1},M_{2}}} \dfrac{1}{n} \sum_{i=1}^{n} \log \left(\dfrac{\exp\biggr(\max \limits_{\z_{i}}\dfrac{\h^{\top}(\y_{i}, \z_{i}; 0)\x_{i}+\eta(\y_{i},\z_{i})}{\sigma^{2}}\biggr)\pi_{\y_{i}}}{\sum \limits_{\y=1}^{K} \exp\biggr(\dfrac{\h^{\top}(\y,\overline{\z}_{i}; 0)\x_{i}+\eta(\y,\overline{\z}_{i})}{\sigma^{2}}\biggr)\pi_{\y}} \right) + \dfrac{M_{2}^{2}-M_{1}^{2}}{2\sigma^{2}} \label{eqn:cross_entropy_DRM_no_constraint_first}
\end{eqnarray}
Additionally, denote $\widetilde{\z}_{i}$ the maximum value of $\z$ in the CNN structure of $h(\y,\x_{i})$ for all $1 \leq \y \leq K$ and $1 \leq i \leq n$. Then, the following inequality holds
\begin{eqnarray}
& & \hspace{ -2.5 em} \log \left(\dfrac{\exp\biggr(\max \limits_{\z_{i}}\dfrac{\h^{\top}(\y_{i}, \z_{i}; 0)\x_{i}+\eta(\y_{i},\z_{i})}{\sigma^{2}}\biggr)\pi_{\y_{i}}}{\sum \limits_{\y=1}^{K} \exp\biggr(\dfrac{\h^{\top}(\y,\overline{\z}_{i}; 0)\x_{i}+\eta(\y,\overline{\z}_{i})}{\sigma^{2}}\biggr)\pi_{\y}} \right) - \log \biggr(\text{Softmax}\biggr(g(\y_{i},\x_{i})\biggr) + \bias_{\y_{i}} \biggr)\biggr) \nonumber \\
& & \hspace{ -2.5 em}  = \max \limits_{\z_{i}}\biggr(\dfrac{\h^{\top}(\y_{i}, \z_{i}; 0)\x_{i} + \eta(\y_{i},\z_{i})}{\sigma^{2}}\biggr) - g(\y_{i},\x_{i}) + \log \left(\dfrac{\sum \limits_{\y=1}^{K} \exp\biggr(\dfrac{\h^{\top}(\y,\widetilde{\z}_{i}; 0)\x_{i}+\eta(\y,\widetilde{\z}_{i})}{\sigma^{2}}\biggr)\pi_{\y}}{\sum \limits_{\y=1}^{K} \exp\biggr(\dfrac{\h^{\top}(\y,\overline{\z}_{i}; 0)\x_{i}+\eta(\y,\overline{\z}_{i})}{\sigma^{2}}\biggr)\pi_{\y}} \right) \nonumber \\
& & \hspace{ -2.5 em}  \leq \max \limits_{\z_{i}}\biggr(\dfrac{\h^{\top}(\y_{i}, \z_{i}; 0)\x_{i} + \eta(\y_{i},\z_{i})}{\sigma^{2}}\biggr) - g(\y_{i},\x_{i}) + \max \limits_{\y} \biggr\{\biggr(\dfrac{\h^{\top}(\y,\widetilde{\z}_{i}; 0)\x_{i}+\eta(\y,\widetilde{\z}_{i})}{\sigma^{2}}\biggr)+\log \pi_{\y}\biggr\} \nonumber \\
& & \hspace{ 16 em} - \max \limits_{\y}\biggr\{\biggr(\dfrac{\h^{\top}(\y,\overline{\z}_{i}; 0)\x_{i}+\eta(\y,\overline{\z}_{i})}{\sigma^{2}}\biggr)+\log \pi_{\y}\biggr\} + \log K \nonumber \\
& & \hspace{ -2.5 em}  \leq \max \limits_{\z_{i}}\biggr(\dfrac{\h^{\top}(\y_{i}, \z_{i}; 0)\x_{i} + \eta(\y_{i},\z_{i})}{\sigma^{2}}\biggr) - g(\y_{i},\x_{i}) + \max \limits_{\y}\biggr\{ \biggr(g(\y,\x_{i}) - \dfrac{\h^{\top}(\y,\overline{\z}_{i}; 0)\x_{i} + \eta(\y,\overline{\z}_{i})}{\sigma^{2}} \biggr)\biggr\} + \log K\nonumber \\\label{eqn:cross_entropy_DRM_no_constraint_second}
\end{eqnarray}
Combining the results from \eqref{eqn:cross_entropy_DRM_no_constraint_first} and \eqref{eqn:cross_entropy_DRM_no_constraint_second}, we obtain the conclusion of part (b) of the theorem. 
\paragraph{PROOF OF THEOREM \ref{theorem:cross_entropy_full_posteterior_LD_DRM_no_constraint}}
To simplify the presentation, throughout this proof, we denote $A = \max_{(\z_{i})_{i=1}^{n}, \theta \in \mathcal{B}_{M_{1},M_{2}}} \dfrac{1}{n} \sum \limits_{i=1}^{n} \max \limits_{\z_{i}}\log p(\y_{i},\z_{i}|\x_{i};\theta)$.
(a) Regarding the lower bound, we have
\begin{eqnarray}
A & & =  \max_{\theta \in \mathcal{B}_{M_{1},M_{2}}}  \dfrac{1}{n} \sum \limits_{i=1}^{n} \max \limits_{\z_{i}}\log \dfrac{p(\x_{i}|\y_{i},\z_{i};\theta) p(\y_{i},\z_{i};\theta)}{\sum \limits_{\y,\z} p(\x_{i}|\y,\z;\theta) p(\y,\z;\theta)} \nonumber \\
& & \leq \max_{\theta \in \mathcal{B}_{M_{1},M_{2}}}  \dfrac{1}{n} \sum \limits_{i=1}^{n} \max \limits_{\z_{i}}\log \left(\dfrac{\exp\biggr(\dfrac{\h^{\top}(\y_{i}, \z_{i}; 0)\x_{i}+\eta(\y_{i},\z_{i})}{\sigma^{2}}\biggr)\pi_{\y_{i}}}{\sum \limits_{\y=1}^{K} \exp(g(\y,\x_{i}))\pi_{\y}}\right) + \dfrac{M_{2}^{2} - M_{1}^{2}}{2\sigma^{2}} \nonumber \\
& & = \max_{\theta \in \mathcal{B}_{M_{1},M_{2}}} \dfrac{1}{n} \sum_{i=1}^{n}\biggr\{\log \biggr(\text{Softmax}\biggr(g(\y_{i},\x_{i})\biggr) + \bias_{\y_{i}} \biggr)\biggr) \nonumber \\
& & \hspace{ 4 em} + \max \limits_{\z_{i}}\biggr(\dfrac{\h^{\top}(\y_{i}, \z_{i})\x_{i} + \eta(\y_{i},\z_{i})}{\sigma^{2}}\biggr) - g(\y_{i},\x_{i}) \biggr\} + \dfrac{M_{2}^{2}-M_{1}^{2}}{2\sigma^{2}} \nonumber
\end{eqnarray}
Therefore, we obtain the conclusion of part (a) of the theorem.

(b) Regarding the upper bound, we obtain that
\begin{eqnarray}
A & & \geq \max_{\theta \in \mathcal{B}_{M_{1},M_{2}}}\dfrac{1}{n} \sum \limits_{i=1}^{n} \log \left(\dfrac{\exp\biggr(\max \limits_{\z_{i}}\dfrac{\h^{\top}(\y_{i}, \z_{i})\x_{i}+\eta(\y_{i},\z_{i})}{\sigma^{2}}\biggr)\pi_{\y_{i}}}{\sum \limits_{\y,\z} \exp\biggr(\dfrac{\h^{\top}(\y,\z;0)\x_{i}+\eta(\y,\z)}{\sigma^{2}}\biggr)\pi_{\y}} \right) - \dfrac{M_{2}^{2} - M_{1}^{2}}{2\sigma^{2}} \nonumber \\
& & \geq \max_{\theta \in \mathcal{B}_{M_{1},M_{2}}} \dfrac{1}{n} \sum_{i=1}^{n}\biggr\{\log \biggr(\text{Softmax}\biggr(g(\y_{i},\x_{i})\biggr) + \bias_{\y_{i}} \biggr)\biggr) \nonumber \\
& & + \log \left(\dfrac{\sum \limits_{\y=1}^{K} \exp(g(\y,\x_{i}))\pi_{\y}}{\sum \limits_{\y,\z} \exp\biggr(\dfrac{\h^{\top}(\y,\z;0)\x_{i}+\eta(\y,\z)}{\sigma^{2}}\biggr)\pi_{\y}} \right) \biggr\} - \dfrac{M_{2}^{2}-M_{1}^{2}}{2\sigma^{2}} = B \nonumber \\
\end{eqnarray}
By means of direction computation, the following holds
\begin{eqnarray}
B & & \geq \max_{\theta \in \mathcal{B}_{M_{1},M_{2}}} \dfrac{1}{n} \sum_{i=1}^{n}\biggr\{\log \biggr(\text{Softmax}\biggr(g(\y_{i},\x_{i})\biggr) + \bias_{\y_{i}} \biggr)\biggr) \nonumber \\
& & + \biggr(\max \limits_{\y} g(\y,\x_{i}) - \max \limits_{\y,\z}\biggr(\dfrac{\h^{\top}(\y,\z;0)\x_{i} + \eta(\y,\z)}{\sigma^{2}}\biggr)\biggr)\biggr\} -\log(|\mathcal{J}|) - \dfrac{M_{2}^{2}-M_{1}^{2}}{2\sigma^{2}}. \nonumber
\end{eqnarray}
As a consequence, we achieve the conclusion of part (b) of the theorem. 
\paragraph{PROOF OF THEOREM \ref{theorem:consistency_SPLDAP_Kmeans}}
To achieve the conclusion of the theorem, we need the following uniform laws of large numbers results
\begin{eqnarray}
& & \hspace{-2 em} \sup \limits_{\theta} \biggr|\dfrac{1}{n}\sum \limits_{i=1}^{n}{\max \limits_{(\y,\z) \in \mathcal{J}} \biggr(\h^{\top}(\y,\z;0)\x_{i}+\tau(\theta_{\y,\z})\biggr)} - \int \max \limits_{(\y,\z) \in \mathcal{J}} \biggr(\h^{\top}(\y,\z;0)x+\tau(\theta_{\y,\z})\biggr) dP(x)\biggr| \to 0, \label{eqn:consistency_spldap_kmeans_first} \\
& & \hspace{-2 em} L_{n} = \sup \limits_{\theta} \biggr|\dfrac{1}{n}\sum \limits_{i=1}^{n} \sum \limits_{(\y',\z') \in \mathcal{J}}{1_{\left\{(\y',\z') = \mathop{\arg \max}\limits_{(\y,\z) \in \mathcal{J}} \biggr(\h^{\top}(\y,\z;0)\x_{i}+\tau(\theta_{\y,\z})\biggr) \right\}}\biggr(\dfrac{\|\h(\y',\z';0)\|^{2}}{2} - \log \pi_{\y'} \biggr)}\nonumber \\
& & - \hspace{-2.5 em} \int \biggr(\sum \limits_{(\y',\z') \in \mathcal{J}}{1_{\left\{(\y',\z') = \mathop{\arg \max}\limits_{(\y,\z) \in \mathcal{J}} \biggr(\h^{\top}(\y,\z;0)x +\tau(\theta_{\y,\z})\biggr) \right\}}\biggr(\dfrac{\|\h(\y',\z';0)\|^{2}}{2} - \log \pi_{\y'} \biggr)}\biggr)dP(x) \biggr| \to 0 \label{eqn:consistency_spldap_kmeans_second}
\end{eqnarray}
almost surely as $n \to \infty$. The proofs of these results are similar to those in Theorem~\ref{theorem:consistency_objective_LDAP_cross_entropy}; therefore, they are omitted.

Going back to the original problem, we denote $\theta^{0} = \biggr(\left\{\muy^{0}(y)\right\}_{\y=1}^{K},\left\{\template_{0}(l)\right\}_{\ell=1}^{L},\left\{\pi_{\y}^{0}\right\}_{\y=1}^{K},\left\{\bias_{0}(\ell) \right\}_{\ell = 1}^{L} \biggr)$ the optimal solutions of population SPLD regularized K-means (Note that, the existence of these optimal solutions is guaranteed due to the compact assumptions with the parameter spaces $\Theta_{l}$, $\Omega$, and $\Xi_{l}$ for all $1 \leq  l \leq L$). Then, according to the formulation of SPLD regularized K-means, we will have that
\begin{eqnarray}
V_{n} & \leq & \dfrac{1}{n} \sum \limits_{i=1}^{n} \sum \limits_{(\y',\z') \in \mathcal{J}}1_{\left\{(\y',\z') = \mathop{\arg \max}\limits_{(\y,\z) \in \mathcal{J}} \biggr(\x_{i}^{\top}\h^{0}(\y,\z;0)+\tau(\theta_{\y,\z}^{0})\biggr) \right\}} \biggr(\dfrac{\| \x_{i} - \h^{0}(\y',\z';0)\|^{2}}{2} - \log(\pi_{\y',\z'}^{0})\biggr) \nonumber \\
& = &  \dfrac{1}{2n} \sum \limits_{i=1}^{n} \|\x_{i}\|^{2} - \dfrac{1}{n} \sum \limits_{i=1}^{n} \max \limits_{(\y,\z) \in \mathcal{J}} \biggr(\x_{i}^{\top}\h^{0}(\y,\z;0)+\tau(\theta_{\y,\z}^{0})\biggr) \nonumber \\
& + & \dfrac{1}{n}\sum \limits_{(\y',\z') \in \mathcal{J}}1_{\left\{(\y',\z') = \mathop{\arg \max}\limits_{(\y,\z) \in \mathcal{J}} \biggr(\x_{i}^{\top}\h^{0}(\y,\z;0) +\tau(\theta_{\y,\z}^{0})\biggr)\right\}}\biggr(\dfrac{\|\h^{0}(\y',\z';0)\|^{2}}{2} - \log \pi_{\y'}^{0}\biggr) \nonumber \\
& + & \log \biggr(\sum \limits_{\y=1}^{K} \exp \biggr(\max \limits_{\z \in \mathcal{L}} \tau(\theta_{\y,\z}^{0})+\log \pi_{\y}^{0}\biggr)\biggr) \nonumber
\end{eqnarray}
where we define
\begin{align}
\h^{0}(\y,\z;0) & = \Lambda_{0}(z;1)\ldots\Lambda_{0}(z;1)\muy^{0}(\y), \nonumber \\
\Lambda_{0}(\z;\ell) & : = \sum_{\p\in \pp(\ell)}  \sparse(\ell, \p)\ttrans(t;\ell,\p) \pad(\ell, \p) \template_{0}(\ell, \p), \nonumber \\
p_{\y',\z'}^{0} & = \exp\biggr(\tau(\theta_{\y',\z'}^{0})+\log \pi_{\y'}^{0} \biggr)/\biggr(\sum \limits_{\y=1}^{K}\exp\biggr(\max \limits_{\z \in \mathcal{L}} \tau(\theta_{\y,\z}^{0})+\log \pi_{\y}^{0} \biggr)\biggr) \nonumber
\end{align}
for all $1 \leq \ell \leq L$. From the results with uniform laws of large numbers in \eqref{eqn:consistency_spldap_kmeans_first} and \eqref{eqn:consistency_spldap_kmeans_second}, we obtain that
\begin{eqnarray}
& & \dfrac{1}{n} \sum \limits_{i=1}^{n} \max \limits_{(\y,\z) \in \mathcal{J}} \biggr(\x_{i}^{\top}\h^{0}(\y,\z;0) +\tau(\theta_{\y,\z}^{0})\biggr) \to \int \max \limits_{(\y,\z) \in \mathcal{J}} \biggr(x^{\top}\h^{0}(\y,\z;0)+\tau(\theta_{\y,\z}^{0})\biggr) dP(x), \nonumber \\
& & \dfrac{1}{n}\sum \limits_{(\y',\z') \in \mathcal{J}}1_{\left\{(\y',\z') = \mathop{\arg \max}\limits_{(\y,\z) \in \mathcal{J}} \biggr(\x_{i}^{\top}\h^{0}(\y,\z;0) +\tau(\theta_{\y,\z}^{0})\biggr)\right\}}\biggr(\dfrac{\|\h^{0}(\y',\z';0) \|^{2}}{2} - \log \pi_{\y'}^{0} \biggr) \nonumber \\
& & \hspace{3 em} \to \int \biggr(\sum \limits_{(\y',\z') \in \mathcal{J}}1_{\left\{(\y',\z') = \mathop{\arg \max}\limits_{(\y,\z) \in \mathcal{J}} \biggr(x^{\top}\h^{0}(\y,\z;0) +\tau(\theta_{\y,\z}^{0})\biggr)\right\}}\biggr(\dfrac{\|\h^{0}(\y',\z';0) \|^{2}}{2} - \log \pi_{\y'}^{0} \biggr)\biggr)dP(x) \nonumber
\end{eqnarray}
almost surely as $n \to \infty$. The above results lead to
\begin{eqnarray}
\dfrac{1}{n} \sum \limits_{i=1}^{n} \sum \limits_{(\y',\z') \in \mathcal{J}}1_{\left\{(\y',\z') = \mathop{\arg \max}\limits_{(\y,\z) \in \mathcal{J}} \biggr(\x_{i}^{\top}\h^{0}(\y,\z;0) + \tau(\theta_{\y,\z}^{0})\biggr) \right\}} \biggr(\dfrac{\| \x_{i} - \h^{0}(\y',\z';0) \|^{2}}{2} - \log(\pi_{\y',\z'}^{0})\biggr)  \to \overline{V} \nonumber
\end{eqnarray}
almost surely as $n \to \infty$. Therefore, we have $\lim \limits_{n \to \infty}V_{n} \leq \overline{V}$ almost surely as $n \to \infty$. On the other hand, with the results of uniform laws of large numbers in \eqref{eqn:consistency_spldap_kmeans_first} and \eqref{eqn:consistency_spldap_kmeans_second}, we also have that
\begin{eqnarray}
& & \dfrac{1}{n} \sum \limits_{i=1}^{n} \max \limits_{(\y,\z) \in \mathcal{J}} \biggr(\x_{i}^{\top}\widehat{\h}(\y,\z;0)+\tau(\widehat{\theta}_{\y',\z'})\biggr) \to \int \max \limits_{(\y,\z) \in \mathcal{J}} \biggr(x^{\top}\widehat{\h}(\y,\z;0)+\tau(\widehat{\theta}_{\y',\z'})\biggr) dP(x), \nonumber \\
& & \dfrac{1}{n}\sum \limits_{(\y',\z') \in \mathcal{J}}1_{\left\{(\y',\z') = \mathop{\arg \max}\limits_{(\y,\z) \in \mathcal{J}} \biggr(\x_{i}^{\top}\widehat{\h}(\y,\z;0)+\tau(\widehat{\theta}_{\y',\z'})\biggr) \right\}}\biggr(\dfrac{\|\widehat{\h}(\y',\z';0)\|^{2}}{2} - \log \widehat{\pi}_{\y'} \biggr) \nonumber \\
& & \hspace{3 em} \to \int \biggr(\sum \limits_{(\y',\z') \in \mathcal{J}}1_{\left\{(\y',\z') = \mathop{\arg \max}\limits_{(\y,\z) \in \mathcal{J}} \biggr(\x_{i}^{\top}\widehat{\h}(\y,\z;0)+\tau(\widehat{\theta}_{\y',\z'})\biggr) \right\}}\biggr(\dfrac{\|\widehat{\h}(\y',\z';0)\|^{2}}{2} - \log \widehat{\pi}_{\y'} \biggr)\biggr)dP(x) \nonumber
\end{eqnarray}
almost surely as $n \to \infty$ where $\widehat{\theta}=\biggr(\left\{\widehat{\muy}(\y)\right\}_{\y=1}^{K},\left\{\widehat{\template}(\ell)\right\}_{\ell=1}^{L},\left\{\widehat{\pi}_{\y}\right\}_{\y=1}^{K},\left\{\widehat{\bias}(\ell)\right\}_{\ell=1}^{L}\biggr)$ is the optimal solution of SPLD regularized K-means and
\begin{align}
\widehat{\h}(\y,\z;0) & = \widehat{\Lambda}(\z;1)\ldots\widehat{\Lambda}(\z;L)\widehat{\muy}(\y), \nonumber \\
\widehat{\Lambda}(\z;\ell) & = \sum_{\p\in \pp(\ell)}  \sparse(\ell, \p)\ttrans(t;\ell,\p) \pad(\ell, \p)\template(\ell, \p), \nonumber \\
\widehat{p}(\y',\z') & = \exp\biggr(\tau(\widehat{\theta}_{\y',\z'})+\log \widehat{\pi}_{\y'} \biggr)/\biggr(\sum \limits_{\y=1}^{K}\exp\biggr(\max \limits_{\z \in \mathcal{L}} \tau(\widehat{\theta}_{\y',\z'})+\log \widehat{\pi}_{\y} \biggr)\biggr) \nonumber
\end{align} 
for all $1 \leq  \ell \leq L$.
Hence, 
\begin{eqnarray}
V_{n} \to \int \sum \limits_{(\y',\z') \in \mathcal{J}} 1_{\left\{(\y',\z') = \mathop{\arg \max}\limits_{(\y,\z) \in \mathcal{J}} \biggr(\x_{i}^{\top}\widehat{\h}(\y,\z;0)+\tau(\widehat{\theta}_{\y',\z'})\biggr) \right\}} \biggr(\dfrac{\| x - \widehat{\h}(\y',\z';0)\|^{2}}{2} - \log(\widehat{p}(\y',\z'))\biggr) dP(x) \geq \overline{V} \nonumber
\end{eqnarray}
almost surely as $n \to \infty$. As a consequence, we achieve $V_{n} \to \overline{V}$ almost surely as $n \to \infty$. We reach the conclusion of the theorem. 
\paragraph{PROOF OF THEOREM \ref{theorem:consistency_optimal_solutions_SPLDAP_Kmeans}}
The proof of the theorem is a straightforward application of the results with uniform laws of large numbers established in the proof of Theorem \ref{theorem:consistency_SPLDAP_Kmeans}. In particular, we denote the following set
\begin{eqnarray}
& & \mathcal{F}_{0}(\epsilon) = \biggr\{\left(S,\left\{\pi_{\y}\right\},\left\{\bias(\ell)\right\}\right): S=\biggr\{\h(\y,\z;0): \ (\y,\z) \in \Jcal \biggr\}, \nonumber \\
& &  \text{and} \inf \limits_{\left(S_{0}, \left\{\pi_{\y}^{0}\right\},\left\{\bias_{0}(\ell)\right\}\right) \in \Gcal(\mathcal{F}_{0})} \biggr\{H(\widehat{S}_{n},S_{0}') + \sum \limits_{\y=1}^{K} |\pi_{\y} - \pi_{\y}^{0}| + \sum \limits_{\z \in \mathcal{L}} \|\bias(\ell) - \bias_{0}(\ell)\|\biggr\} \geq \epsilon \biggr\} \nonumber
\end{eqnarray}
for any $\epsilon>0$. Since the parameter spaces of $\theta$ are compact sets, the set $\mathcal{F}_{0}(\epsilon)$ is also a compact set for all $\epsilon>0$. Denote
\begin{eqnarray}
g\left(S,\left\{\pi_{\y}\right\},\left\{\bias(\ell)\right\}\right) & = & \int \sum \limits_{(\y',\z') \in \mathcal{J}}1_{\left\{(\y',\z') = \mathop{\arg \max}\limits_{(\y,\z) \in \mathcal{J}} \biggr(\h^{\top}(\y,\z;0)x +\tau(\theta_{\y,\z})\biggr)\right\}} \nonumber \\
& & \hspace{6 em} \times  \biggr(\dfrac{\| x - \h(\y',\z';0)\|^{2}}{2} - \log(\pi_{\y',\z'})\biggr)dP(x)  \nonumber
\end{eqnarray}
for all $\left(S,\left\{\pi_{\y}\right\},\left\{\bias(\ell)\right\}\right)$. 
From the definition of $\mathcal{F}_{0}(\epsilon)$, we have that 
\begin{eqnarray}
g\left(S,\left\{\pi_{\y}\right\},\left\{\bias(\ell)\right\}\right) > g\left(S_{0}, \left\{\pi_{\y}^{0}\right\},\left\{\bias_{0}(\ell)\right\}\right) \nonumber
\end{eqnarray} 
for all $\left(S,\left\{\pi_{\y}\right\},\left\{\bias(\ell)\right\}\right) \in \mathcal{F}_{0}(\epsilon)$ and $\left(S_{0}, \left\{\pi_{\y}^{0}\right\},\left\{\bias_{0}(\ell)\right\}\right) \in \Gcal(\mathcal{F}_{0})$. As $\mathcal{F}_{0}(\epsilon)$ is a compact set, we further have that
\begin{eqnarray}
\inf \limits_{\left(S,\left\{\pi_{\y}\right\},\left\{\bias(\ell)\right\}\right) \in \mathcal{F}_{0}(\epsilon)} g\left(S,\left\{\pi_{\y}\right\},\left\{\bias(\ell)\right\}\right) > g\left(S_{0}, \left\{\pi_{\y}^{0}\right\},\left\{\bias_{0}(\ell)\right\}\right) \nonumber
\end{eqnarray} 
for all $\left(S_{0}, \left\{\pi_{\y}^{0}\right\},\left\{\bias_{0}(\ell)\right\}\right) \in \Gcal(\mathcal{F}_{0})$ and $\epsilon>0$. 

Now, according to the uniform laws of large numbers established in \eqref{eqn:consistency_spldap_kmeans_first} and \eqref{eqn:consistency_spldap_kmeans_second}, we have that $V_{n} \to g(\widehat{S}_{n},\left\{\widehat{\pi}_{\y}\right\},\left\{\widehat{\bias}(\ell)\right\})$ almost surely as $n \to \infty$. According to the result of Theorem \ref{theorem:consistency_SPLDAP_Kmeans}, it implies that $g(\widehat{S}_{n},\left\{\widehat{\pi}_{\y}\right\},\left\{\widehat{\bias}(\ell)\right\}) \to g\left(S_{0}, \left\{\pi_{\y}^{0}\right\},\left\{\bias_{0}(\ell)\right\}\right)$ almost surely for all $\left(S_{0}, \left\{\pi_{\y}^{0}\right\},\left\{\bias_{0}(\ell)\right\}\right) \in \Gcal(\mathcal{F}_{0})$. Therefore, for each $\epsilon>0$ we can find sufficiently large $N$ such that we have
\begin{eqnarray}
\inf \limits_{\left(S_{0}, \left\{\pi_{\y}^{0}\right\},\left\{\bias_{0}(\ell)\right\}\right) \in \Gcal(\mathcal{F}_{0})} \biggr\{H(\widehat{S}_{n},S_{0}') + \sum \limits_{\y=1}^{K} |\pi_{\y} - \pi_{\y}^{0}| + \sum \limits_{\z \in \mathcal{L}} \|\bias(\ell) - \bias_{0}(\ell)\|\biggr\} < \epsilon \nonumber
\end{eqnarray}
almost surely for all $n \geq N$. As a consequence, we achieve the conclusion of the theorem.
\paragraph{PROOF OF THEOREM \ref{theorem:generalization_bound_SPLDAP K-means}}
The proof of this result relies on the evaluation of Rademacher complexity being established in the proof of Theorem \ref{theorem:consistency_objective_LDAP_cross_entropy}. In particular, from the definitions of $V_{n}$ and $\overline{V}_{n}$, the conclusion of the theorem is equivalent to demonstrate that
\begin{eqnarray}
& & A = \biggr|\mathbb{E} \biggr\{\dfrac{1}{n}\sum \limits_{i=1}^{n} \sum \limits_{(\y',\z') \in \overline{\mathcal{J}}_{n}}1_{\left\{(\y',\z') = \mathop{\arg \max}\limits_{(\y,\z) \in \overline{\mathcal{J}}_{n}} \biggr(\widehat{\h}(\y,\z;0)^{\top}\x_{i}+\tau(\widehat{\theta}_{\y,\z})\biggr)\right\}} \biggr(\dfrac{\| \x_{i} - \widehat{\h}(\y',\z';0)\|^{2}}{2} - \log(\widehat{p}(\y',\z'))\biggr)\biggr\} \nonumber \\
& & \hspace{-1 em} - \mathbb{E} \biggr\{\int \biggr(\sum \limits_{(\y',\z') \in \overline{\mathcal{J}}_{n}}1_{\left\{(\y',\z') = \mathop{\arg \max}\limits_{(\y,\z) \in \overline{\mathcal{J}}_{n}} \biggr(\widehat{\h}(\y,\z;0)^{\top}x+\tau(\widehat{\theta}_{\y,\z})\biggr) \right\}} \biggr(\dfrac{\| \x_{i} - \widehat{\h}(\y',\z';0)\|^{2}}{2} - \log(\widehat{p}(\y',\z'))\biggr)\biggr)dP(x)\biggr\}\biggr| \nonumber \\
& & \hspace{5 em} \leq  \dfrac{4 \overline{c}_{n}|\Jcal|R^{2}}{\sqrt{n}} + \left(2\log(1/\overline{\gamma})+R^{2}\right)\sqrt{\dfrac{2 (\log 8+(D^{(0)}+1)(\overline{c}_{n}|\Jcal|-1)\log n)}{n}}. \nonumber
\end{eqnarray}
In fact, we can rewrite $A$ as 
\begin{eqnarray}
& & \hspace{- 1 em} A = \biggr|\mathbb{E} \biggr\{\dfrac{1}{n}\sum \limits_{i=1}^{n} \sum \limits_{(\y',\z') \in \overline{\mathcal{J}}_{n}}1_{\left\{(\y',\z') = \mathop{\arg \max}\limits_{(\y,\z) \in \overline{\mathcal{J}}_{n}} \biggr(\widehat{\h}(\y,\z;0)^{\top}\x_{i}+\tau(\widehat{\theta}_{\y,\z})\biggr)\right\}}  \biggr(l_{\widehat{\h}(\y',\z';0)}(\x_{i}) - \tau(\widehat{\theta}_{\y',\z'}) - \log \widehat{\pi}_{\y'}\biggr)\biggr\} \nonumber \\
& & \hspace{- 1 em} - \mathbb{E} \biggr\{\int \biggr(\sum \limits_{(\y',\z') \in \overline{\mathcal{J}}_{n}}1_{\left\{(\y',\z') = \mathop{\arg \max}\limits_{(\y,\z) \in \overline{\mathcal{J}}_{n}} \biggr(\widehat{\h}(\y,\z;0)^{\top}x+\tau(\widehat{\theta}_{\y,\z})\biggr) \right\}} \biggr(l_{\widehat{\h}(\y',\z';0)}(x) - \tau(\widehat{\theta}_{\y',\z'}) - \log \widehat{\pi}_{\y'}\biggr)\biggr)dP(x)\biggr\}\biggr| \nonumber
\end{eqnarray}
where $l_{y}(x)= -x^{\top}y+\|y\|^{2}/2$ for all $(x,y)$. By means of the standard symmetrization argument from empirical processes theory~\cite{vanderVaart-Wellner-96}, we obtain that
\begin{eqnarray}
& & \hspace{- 2 em} A \leq \mathbb{E} \biggr| \sum \limits_{i=1}^{n} \sum \limits_{(\y',\z') \in \overline{\mathcal{J}}_{n}}1_{\left\{(\y',\z') = \mathop{\arg \max}\limits_{(\y,\z) \in \overline{\mathcal{J}}_{n}} \biggr(\widehat{\h}(\y,\z;0)^{\top}\x_{i}+\tau(\widehat{\theta}_{\y,\z})\biggr)\right\}} \biggr(l_{\widehat{\h}(\y',\z';0)}(\x_{i}) - \tau(\widehat{\theta}_{\y',\z'}) - \log \widehat{\pi}_{\y'} \biggr) \nonumber \\
& & \hspace{- 2 em} - \int \biggr(\sum \limits_{(\y',\z') \in \overline{\mathcal{J}}_{n}}1_{\left\{(\y',\z') = \mathop{\arg \max}\limits_{(\y,\z) \in \overline{\mathcal{J}}_{n}} \biggr(\widehat{\h}(\y,\z;0)^{\top}x+\tau(\widehat{\theta}_{\y,\z})\biggr) \right\}} \biggr(l_{\widehat{\h}(\y',\z';0)}(x) - \tau(\widehat{\theta}_{\y',\z'}) - \log \widehat{\pi}_{\y'} \biggr)\biggr)dP(x)\biggr| \nonumber \\
& & \hspace{- 2 em} \leq \mathbb{E} \sup \limits_{\theta} \sup \limits_{|\Jcal'| \leq \overline{c}_{n}|\Jcal|} \biggr| \dfrac{1}{n} \sum \limits_{i=1}^{n} \sum \limits_{(\y',\z') \in \Jcal'}1_{\left\{(\y',\z') = \mathop{\arg \max}\limits_{(\y,\z) \in \Jcal'} \biggr(\h^{\top}(\y,\z;0)\x_{i}+\tau(\theta_{\y,\z})\biggr) \right\}} \nonumber \\
& & \hspace{20 em} \times \biggr(l_{\h(\y',\z';0)}(\x_{i}) - \tau(\theta_{\y',\z'})-\log \pi_{\y'} \biggr) \nonumber \\
& & \hspace{- 2 em} - \int \biggr(\sum \limits_{(\y',\z') \in \Jcal'}1_{\left\{(\y',\z') = \mathop{\arg \max}\limits_{(\y,\z) \in \Jcal'} \biggr(\h^{\top}(\y,\z;0)x+\tau(\theta_{\y,\z})\biggr)\right\}} \biggr(l_{\h(\y',\z';0)}(x) - \tau(\theta_{\y',\z'}) - \log \pi_{\y'} \biggr)\biggr)dP(x)\biggr| \nonumber \\
& & \hspace{- 2 em} \leq 2 \mathbb{E} \sup \limits_{\theta} \sup \limits_{|\Jcal'| \leq \overline{c}_{n}|\Jcal|} \biggr| \dfrac{1}{n} \sum \limits_{i=1}^{n} \sigma_{i}\sum \limits_{(\y',\z') \in \Jcal'}1_{\left\{(\y',\z') = \mathop{\arg \max}\limits_{(\y,\z) \in \Jcal'} \biggr(\h^{\top}(\y,\z;0)\x_{i}+\tau(\theta_{\y,\z})\biggr) \right\}} \nonumber \\
& & \hspace{20 em} \times \biggr(l_{\h(\y',\z';0)}(\x_{i}) - \tau(\theta_{\y',\z'})-\log \pi_{\y'} \biggr)\biggr| \label{eqn:generalization_bound_SPLDAP_Kmeans_first}
\end{eqnarray}
where $\sigma_{1},\ldots,\sigma_{n}$ are i.i.d. Rademacher random variables, i.e., $\mathbb{P}(\sigma_{i}=-1) = \mathbb{P}(\sigma_{i}=1) = 1/2$ for $1 \leq i \leq n$. By means of the triangle inequality and the additive property of supremum, we have the following inequality
\begin{eqnarray}
& & \hspace{- 2 em} \mathbb{E} \sup \limits_{\theta} \sup \limits_{|\Jcal'| \leq \overline{c}_{n}|\Jcal|} \biggr| \dfrac{1}{n} \sum \limits_{i=1}^{n} \sigma_{i}\sum \limits_{(\y',\z') \in \Jcal'}1_{\left\{(\y',\z') = \mathop{\arg \max}\limits_{(\y,\z) \in \Jcal'} \biggr(\h^{\top}(\y,\z;0)\x_{i}+\tau(\theta_{\y,\z})\biggr) \right\}} \nonumber \\
& & \hspace{20 em} \times  \biggr(l_{\h(\y',\z';0)}(\x_{i}) - \tau(\theta_{\y',\z'})-\log \pi_{\y'} \biggr) \biggr| \nonumber \\
& & \hspace{- 2 em} \leq \mathbb{E} \sup \limits_{\theta} \sup \limits_{|\Jcal'| \leq \overline{c}_{n}|\Jcal|} \biggr| \dfrac{1}{n} \sum \limits_{i=1}^{n} \sigma_{i}\sum \limits_{(\y',\z') \in \Jcal'}1_{\left\{(\y',\z') = \mathop{\arg \max}\limits_{(\y,\z) \in \Jcal'} \biggr(\h^{\top}(\y,\z;0)\x_{i}+\tau(\theta_{\y,\z})\biggr) \right\}}\biggr(\dfrac{\|\h(\y',\z';0)\|^{2}}{2}-\log \pi_{\y'} \biggr) \biggr| \nonumber \\
& & \hspace{- 2 em} + \mathbb{E} \sup \limits_{\theta} \sup \limits_{|\Jcal'| \leq \overline{c}_{n}|\Jcal|} \biggr| \dfrac{1}{n} \sum \limits_{i=1}^{n} \sigma_{i} \max \limits_{(\y,\z) \in \Jcal'} \biggr(\h^{\top}(\y,\z;0)\x_{i}+\tau(\theta_{\y,\z})\biggr)\biggr|. \label{eqn:generalization_bound_SPLDAP_Kmeans_second}
\end{eqnarray}
According to the property of Rademacher complexity and the Cauchy-Schwarz's inequality, we get
\begin{eqnarray}
& & \hspace{ -8 em} \mathbb{E} \sup \limits_{\theta} \sup \limits_{|\Jcal'| \leq \overline{c}_{n}|\Jcal|} \biggr| \dfrac{1}{n} \sum \limits_{i=1}^{n} \sigma_{i} \max \limits_{(\y,\z) \in \Jcal'} \biggr(\h^{\top}(\y,\z;0)\x_{i}+\tau(\theta_{\y,\z})\biggr)\biggr| \nonumber \\
& \leq & \mathbb{E} \sup \limits_{|S'|  \leq  \overline{c}_{n}|\Jcal|} \biggr|\dfrac{1}{n} \sum \limits_{i=1}^{n} \sigma_{i} \max \limits_{s \in S'} s^{\top}[\x_{i},1] \biggr| \nonumber \\
& \leq & 2\overline{c}_{n}|\Jcal| \mathbb{E} \sup \limits_{s \in \mathbb{B}(R)}\biggr|\dfrac{1}{n}\sum \limits_{i=1}^{n}\sigma_{i}s^{\top}[\x_{i},1]\biggr| \nonumber \\
& \leq & 2\overline{c}_{n}|\Jcal| R\mathbb{E}\biggr\|\dfrac{1}{n}\sum \limits_{i=1}^{n}\sigma_{i}[\x_{i},1]\biggr\| \leq \dfrac{2\overline{c}_{n}|\Jcal|(R^{2}+1)}{\sqrt{n}}. \label{eqn:generalization_bound_SPLDAP_Kmeans_third}
\end{eqnarray}
Additionally, using the results with VC dimension of Voronoi partitions in the proof of Theorem \ref{theorem:consistency_SPLDAP_Kmeans}, we achieve that
\begin{eqnarray}
& & \hspace{- 3 em} \mathbb{E} \sup \limits_{\theta} \sup \limits_{|\Jcal'| \leq \overline{c}_{n}|\Jcal|} \biggr| \dfrac{1}{n} \sum \limits_{i=1}^{n} \sigma_{i}\sum \limits_{(\y',\z') \in \Jcal'}1_{\left\{(\y',\z') = \mathop{\arg \max}\limits_{(\y,\z) \in \Jcal'} \biggr(\h^{\top}(\y,\z;0)\x_{i}+\tau(\theta_{\y,\z})\biggr) \right\}}\biggr(\dfrac{\|\h(\y',\z';0)\|^{2}}{2}-\log \pi_{\y'} \biggr) \biggr| \nonumber \\
& = & \mathbb{E}\biggr(\mathbb{E}_{\sigma} \sup \limits_{\theta} \sup \limits_{|\Jcal'| \leq \overline{c}_{n}|\Jcal|} \biggr| \dfrac{1}{n} \sum \limits_{i=1}^{n} \sigma_{i}\sum \limits_{(\y',\z') \in \Jcal'}1_{\left\{(\y',\z') = \mathop{\arg \max}\limits_{(\y,\z) \in \Jcal'} \biggr(\h^{\top}(\y,\z;0)\x_{i}+\tau(\theta_{\y,\z})\biggr) \right\}} \nonumber \\
& & \hspace{16 em} \times \biggr(\dfrac{\|\h(\y',\z';0)\|^{2}}{2}-\log \pi_{\y'} \biggr) \biggr||\x_{1},\ldots,\x_{n}\biggr) \nonumber \\
& \leq & \biggr(\log(1/\overline{\gamma})+R^{2}/2\biggr) \mathbb{E}\biggr(\mathbb{E}_{\sigma} \sup \limits_{|S'| \leq \overline{c}_{n}|\Jcal|} \sum \limits_{j=1}^{|S'|}\biggr|\dfrac{1}{n}\sum \limits_{i=1}^{n}\sigma_{i}1_{\left\{j = \mathop{\arg \min}\limits_{1 \leq l \leq |S'|} [\x_{i},1]^{\top}s_{l}\right\}}\biggr| | \x_{1},\ldots,\x_{n}\biggr) \nonumber \\
& \leq &\biggr(\log(1/\overline{\gamma})+R^{2}/2\biggr)\sqrt{\dfrac{2 (\log 8+(D^{(0)}+1)(\overline{c}_{n}|\Jcal|-1)\log n)}{n}}. \label{eqn:generalization_bound_SPLDAP_Kmeans_fourth}
\end{eqnarray}
where the supremum in the above inequality is taken with $S'=\left\{s_{1},\ldots,s_{|S'|}\right\}$ for all $|\Jcal'| \leq \overline{c}_{n}|\Jcal|$. By combining the results from \eqref{eqn:generalization_bound_SPLDAP_Kmeans_first} \eqref{eqn:generalization_bound_SPLDAP_Kmeans_second}, \eqref{eqn:generalization_bound_SPLDAP_Kmeans_third}, and \eqref{eqn:generalization_bound_SPLDAP_Kmeans_fourth}, we achieve the conclusion of the theorem.
\newpage
\section{Appendix F}
\label{Sec:appendix_F}
In this appendix, we provide detail descriptions for several simulation studies in the main text. 

\subsection{Architecture of the Network Used in Our Semi-Supervised Experiments}
\begin{table}[h]
\centering
\caption{\label{tbl:network}%
The network architecture used in all of semi-supervised experiments on CIFA10, CIFAR100 and SVHN.
}
\vspace*{\baselineskip}
\begin{tabular}{|l|l|}
\hline
\textsc{Name} & \textsc{Description} \\
\hline
\raisebox{0mm}[3mm]{}%
input  & $32\times32$ RGB image \\
conv1a & $128$ filters, $3\times3$, pad = 'same', LReLU ($\alpha=0.1$)\\
conv1b & $128$ filters, $3\times3$, pad = 'same', LReLU ($\alpha=0.1$) \\
conv1c & $128$ filters, $3\times3$, pad = 'same', LReLU ($\alpha=0.1$) \\
pool1   & Maxpool $2\times2$ pixels \\
drop1   & Dropout, $p=0.5$ \\
conv2a & $256$ filters, $3\times3$, pad = 'same', LReLU ($\alpha=0.1$) \\
conv2b & $256$ filters, $3\times3$, pad = 'same', LReLU ($\alpha=0.1$) \\
conv2c & $256$ filters, $3\times3$, pad = 'same', LReLU ($\alpha=0.1$) \\
pool2   & Maxpool $2\times2$ pixels \\
drop2   & Dropout, $p=0.5$ \\
conv3a & $512$ filters, $3\times3$, pad = 'valid', LReLU ($\alpha=0.1$) \\
conv3b & $256$ filters, $1\times1$, LReLU ($\alpha=0.1$) \\
conv3c & $128$ filters, $1\times1$, LReLU ($\alpha=0.1$) \\
pool3   & Global average pool ($6\times6 \to 1\times$1 pixels) \\
dense & Fully connected $128 \to 10$\\
output & Softmax \\\hline
\end{tabular}
\end{table}

\subsection{Training Details}
\subsubsection{Semi-Supervised Learning Experiments on CIFAR10, CIFAR100, and SVHN}
\label{sec:semi-sup-training-details}
The training losses are discussed in Section \ref{sec:learning-in-lddrm}. In addition to the cross-entropy loss, the recontruction loss, and the RPN regularization, in order to further improve the performance of DGM, we introduce two new losses for training the model. Those two new losses are from our derivation of batch normalization using the DGM framework and from applying variational inference on the DGM. More details on these new training losses can be found in Appendix A\ref{sec:inference-NRM_semi_sup}. We compare DGM with state-of-the-art methods on semi-supervised object classification tasks which use consistency regularization, such as the $\Pi$ model \cite{laine2016temporal}, the Temporal Ensembling \cite{laine2016temporal}, the Mean Teacher \cite{tarvainen2017mean}, the Virtual Adversarial Training (VAT), and the Ladder Network \cite{rasmus2015semi}. We also compare DGM with methods that do not use consistency regularization including the improved GAN \cite{salimans2016improved} and the Adversarially Learned Inference (ALI) \cite{dumoulin2016adversarially}.

All networks were trained using Adam with learning rate of 0.001 for the first 20 epochs. Adam momentum parameters were set to beta1 = 0.9 and beta2 = 0.999. Then we used SGD with decayed learning rate to train the networks for another 380 epochs. The starting learning rate for SGD is 0.15 and the end learning rate at epoch 400 is 0.0001. We use batch size 128. Let the weights for the cross-entropy loss, the reconsruction loss, the KL divergence loss, the moment matching loss, and the RPN regularization be $\alpha_{CE}$, $\alpha_{RC}$, $\alpha_{KL}$, $\alpha_{MM}$, and $\alpha_{PN}$, respectively. In our training, $\alpha_{CE}=1.0$, $\alpha_{RC}=0.5$, $\alpha_{KL}=0.5$, $\alpha_{MM}=0.5$, and $\alpha_{PN}=1.0$. For Max-Min cross-entropy, we use $\alpha^{max}=\alpha^{min}=0.5$.

\subsubsection{Supervised Learning Experiments with Max-Min Cross Entropy}
{\bf Training  on CIFAR10}
We use the 26 2 x 96d "Shake-Shake-Image" ResNet in \cite{gastaldi2017shake} with the Cutout data augmentation suggested in \cite{devries2017improved} as our baseline. We implement the Max-Min cross-entropy on top of this baseline and turn it into a Max-Min network. In addition to Cutout data augmentation, standard translation and flipping data augmentation is applied on the 32 x 32 x 3 input image. Training procedure are the same as in ~\cite{gastaldi2017shake}. In particular, the models were trained for 1800 epochs. The learning rate is initialized at 0.2 and
is annealed using a Cosine function without restart (see~\cite{loshchilov2016sgdr}). We train our models on 1 GPU with a mini-batch size of 128. 

{\bf Training  on CIFAR10}
We use the Squeeze-and-Excitation ResNeXt-50 as in \cite{hu2017squeeze} as our baseline. As with CIFAR10, we implement the Max-Min cross-entropy for the baseline and turn it into a Max-Min network. During
training, we follow standard practice and
perform data augmentation with random-size cropping \cite{szegedy2015going}
to 224 x 224 x 3 pixels. We train the network with the Nesterov accelerated SGD for 125 epochs. The intial learning rate is 0.1 with momentum 0.9. We divide the learning rate by 10 at epoch 30, 60, 90, 95, 110, 120. Our network is trained on 8 GPUs with batch size of 32.

\subsubsection{Semi-Supervised Training on MNIST with 50K Labeled to Get the Trained Model for Generating Reconstructed Image in Figure \ref{fig:reconst-imgs}}
The architecture of the baseline CNN we use is given in the Table \ref{tbl:mnistnetwork}. We use the same training procedure as in Section \ref{sec:semi-sup-training-details}

\begin{table}[h]
\centering
\caption{\label{tbl:mnistnetwork}%
The network architecture used in our MNIST semi-supervised training.
}
\vspace*{\baselineskip}
\begin{tabular}{|l|l|}
\hline
\textsc{Name} & \textsc{Description} \\
\hline
\raisebox{0mm}[3mm]{}%
input  & $28\times28$ image \\
conv1 & $32$ filters, $5\times5$, pad = 'full', ReLU \\
pool1   & Maxpool $2\times2$ pixels \\
conv2a & $64$ filters, $3\times3$, pad = 'valid', ReLU \\
conv2b & $64$ filters, $3\times3$, pad = 'full', ReLU \\
pool2   & Maxpool $2\times2$ pixels \\
conv3 & $128$ filters, $3\times3$, pad = 'valid', ReLU \\
pool3   & Global average pool ($6\times6 \to 1\times$1 pixels) \\
dense & Fully connected $128 \to 10$\\
output & Softmax \\\hline
\end{tabular}
\end{table}

\end{document}